\documentclass[10pt,journal,compsoc]{IEEEtran}

\ifCLASSOPTIONcompsoc
  \usepackage[nocompress]{cite}
\else
  \usepackage{cite}
\fi
\usepackage[utf8]{inputenc} 
\usepackage[T1]{fontenc}    
\usepackage{booktabs}       
\usepackage{amsfonts}       
\usepackage{nicefrac}       
\usepackage{microtype}      
\usepackage{amssymb}
\usepackage{graphicx}
\usepackage{tabularx}
\usepackage{multirow}
\usepackage{subfigure}
\usepackage{amsmath,bm}
\usepackage{xcolor}
\definecolor{mydarkred}{rgb}{0.6,0,0}
\definecolor{mydarkgreen}{rgb}{0,0.6,0}
\definecolor{mydarkblue}{rgb}{0,0,0}
\usepackage[colorlinks,
linkcolor=mydarkred,
citecolor=mydarkgreen]{hyperref}
\usepackage{url}
\usepackage{algorithm}
\usepackage{algorithmic}
\usepackage[algo2e]{algorithm2e}
\usepackage{amsfonts}
\usepackage{amsthm}
\usepackage{mathtools}
\usepackage{soul}
\newcommand{\httpsurl}[1]{\href{https://#1}{\nolinkurl{#1}}}

\newenvironment{proofskt}{%
  \proof}{\endproof}

\begin{document}

\title{Butterfly: One-step Approach towards \\ Wildly Unsupervised Domain Adaptation}

\author{Feng~Liu,~\IEEEmembership{Member,~IEEE,} 
        Jie~Lu,~\IEEEmembership{Fellow,~IEEE}, Bo~Han,~Gang~Niu,~Guangquan~Zhang~and~Masashi~Sugiyama
\IEEEcompsocitemizethanks{\IEEEcompsocthanksitem Feng Liu is with the Centre for Artificial Intelligence (CAI), Faculty of Engineering and Information Technology, University of Technology Sydney, Sydney, NSW, Australia, and with Center for Advanced Intelligence Project, RIKEN, Japan. Jie Lu and Guangquan Zhang are with the Centre for Artificial Intelligence (CAI), Faculty of Engineering and Information Technology, University of Technology Sydney, Sydney, NSW, Australia. Bo Han is with Center for Advanced Intelligence Project, RIKEN, Japan, and with Department of Computer Science, Hong Kong Baptist University, Hong Kong SAR. Gang Niu is with Center for Advanced Intelligence Project, RIKEN, Japan. Masashi~Sugiyama is with Center for Advanced Intelligence Project, RIKEN, Japan, and with Graduate School of Frontier Sciences, University of Tokyo, Japan. \protect\\

E-mail: \{Feng.liu, jie.lu\}@uts.edu.au, \{bo.han, gang.niu\}@riken.jp, guangquan.zhang@uts.edu.au and sugi@k.u-tokyo.ac.jp
}
\thanks{Manuscript received January 28, 2020.}
}

\markboth{IEEE Transactions on Pattern Analysis and Machine Intelligence,~Vol.~xx, No.~xx, January~2020}%
{Liu \MakeLowercase{\textit{et al.}}: Butterfly: One-step Approach towards Wildly Unsupervised Domain Adaptation}

\IEEEtitleabstractindextext{%
\begin{abstract}
In \emph{unsupervised domain adaptation}~(UDA), classifiers for the \emph{target domain}~(TD) are trained with \emph{clean} labeled data from the \emph{source domain}~(SD) and unlabeled data from TD. However, in the wild, it is difficult to acquire a large amount of perfectly clean labeled data in SD given limited budget. Hence, we consider a new, more realistic and more challenging problem setting, where classifiers have to be trained with \emph{noisy} labeled data from SD and unlabeled data from TD---we name it \emph{wildly UDA}~(WUDA). We show that WUDA ruins all UDA methods if taking no care of label noise in SD, and to this end, we propose a \emph{Butterfly} framework, a powerful and efficient solution to WUDA. Butterfly maintains four deep networks simultaneously, where two take care of all adaptations (i.e., noisy-to-clean, labeled-to-unlabeled, and SD-to-TD-distributional) and then the other two can focus on classification in TD. As a consequence, Butterfly possesses all the conceptually necessary components for solving WUDA. Experiments demonstrate that, under WUDA, Butterfly significantly outperforms existing baseline methods. The code of Butterfly can be found at \httpsurl{github.com/fengliu90/Butterfly}.
\end{abstract}

\begin{IEEEkeywords}
machine learning, weakly-supervised learning, transfer learning
\end{IEEEkeywords}}

\maketitle

\IEEEdisplaynontitleabstractindextext

\IEEEpeerreviewmaketitle

\newtheorem{mythm}{Theorem}
\newtheorem{mylem}{Lemma}
\newtheorem{mycor}{Corollary}
\newtheorem{myrem}{Remark}
\newtheorem{mydef}{Definition}
\newtheorem{mypro}{Problem}

\IEEEraisesectionheading{\section{Introduction}\label{sec:introduction}}

\IEEEPARstart{D}{omain} adaptation (DA) aims to learn a discriminative classifier in the presence of a shift between training data in source domain and test data in target domain \cite{Ben_DA_bounds,GaninL15,XiaoG15,Kun_zhang_multi_source,Kun_zhang_TCS,stojanov2019data}. Currently, DA can be divided into three categories: \emph{supervised DA} \cite{Kate_SDA}, \emph{semi-supervised DA} \cite{ICML_SSDA,Li2014,Duan2012b,Xiao2015,yan2017learning} and \emph{unsupervised DA} (UDA) \cite{KSaito_ICML17,Gong2016,Long_DAN_journal,Gopalan2014,zhang2019self,liu2018TFS,ghifary2016deep,deng2019cluster}. When the number of labeled data is few in target domain, supervised DA is also known as \emph{few-shot DA} \cite{FSDA17}. Since unlabeled data in target domain can be easily obtained, UDA exhibits the greatest potential in the real world \cite{li2017visual,DANN_JMLR,GFK_CVPR,Saito:MCD,DIRT-T,Kang_cvpr19,GCAN_cvpr19,Ziser_ACL19,Xu2014,gong2019dlow,li2018unsupervised}.

UDA methods train with clean labeled data in a source domain (i.e., clean source data) and unlabeled data in a target domain (i.e., unlabeled target data) to obtain classifiers for the \emph{target domain} (TD), which mainly consist of three orthogonal techniques: \emph{integral probability metrics} (IPM) \cite{Ghifary2017,Gong2016,Gretton2012,Wasserstein_distances,Long_DAN,Long_JAN,Xiyu_TL_noise}, \emph{adversarial training} \cite{DANN_JMLR,Mingming_CGDAB,Judy_adversarial,Kun_zhang_CIAN,Saito:MCD,Judy_ADDA,zhang2018collaborative} and \emph{pseudo labeling} \cite{KSaito_ICML17}. Compared to IPM- and adversarial-training-based methods, the pseudo-labeling-based method (i.e., \emph{asymmetric tri-training domain adaptation} (ATDA) \cite{KSaito_ICML17}) can construct a high-quality target-specific representation, providing a better classification performance. 

\begin{figure}[!tp]
    \centering
    \includegraphics[width=0.4\textwidth]{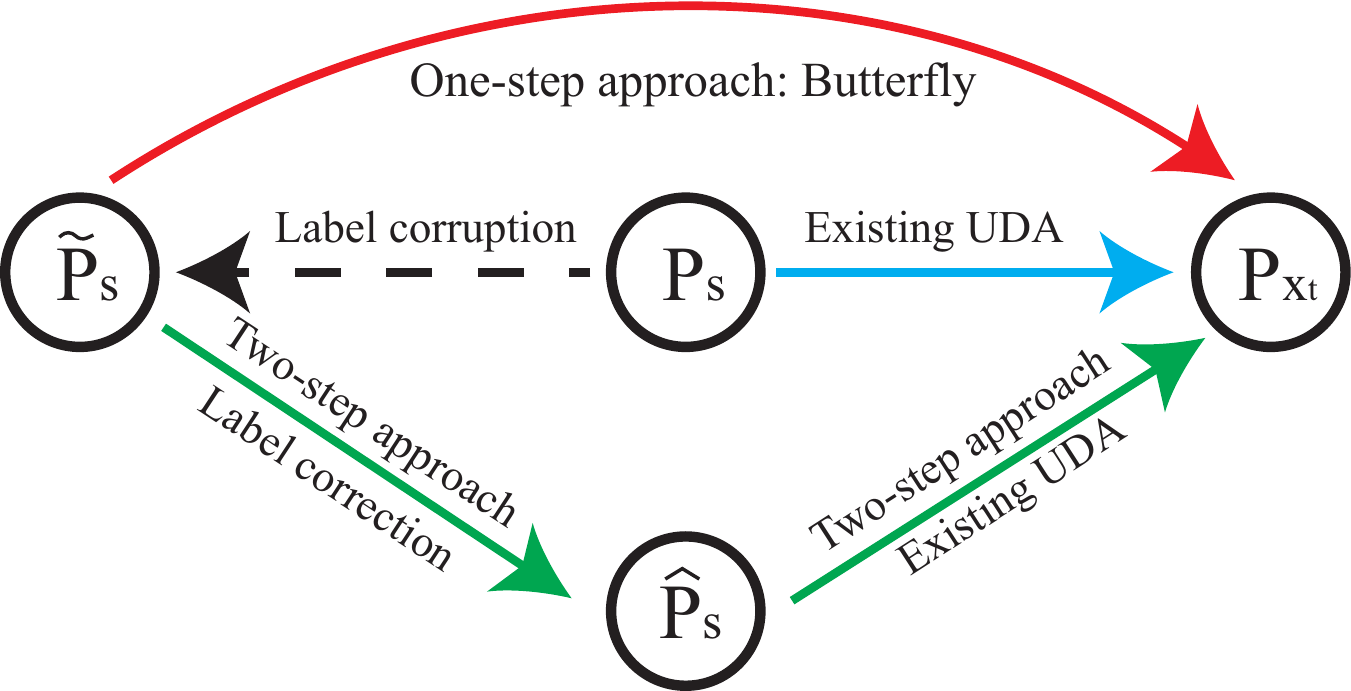}
	\caption{Wildly unsupervised domain adaptation (WUDA). The blue line denotes that UDA transfers knowledge from clean source data ($P_s$) to unlabeled target data ($P_{x_t}$). However, perfectly clean data is hard to acquire. This brings \emph{wildly unsupervised domain adaptation} (WUDA), namely transferring knowledge from noisy source data ($\widetilde{P}_s$) to unlabeled target data ($P_{x_t}$). Note that label corruption process (black dash line) is unknown in practice. To handle WUDA, a compromise solution is a two-step approach (green line), which sequentially combines label-noise algorithms ($\widetilde{P}_s \rightarrow \hat{P}_s$, label correction) and existing UDA ($\hat{P}_s \rightarrow P_{x_t}$). This paper proposes a robust one-step approach called Butterfly (red line, $\widetilde{P}_s \rightarrow P_{x_t}$ directly), which eliminates noise effects from $\tilde{P}_s$.}
	\label{fig: our_solution}
	\vspace{-1em}
\end{figure}

\begin{figure*}[tp]
	\begin{center}
		\subfigure[Symmetry-flip noise: \emph{S}$\rightarrow$\emph{M} (left), \emph{M}$\rightarrow$\emph{S} (right)]
		{\includegraphics[width=0.24\textwidth]{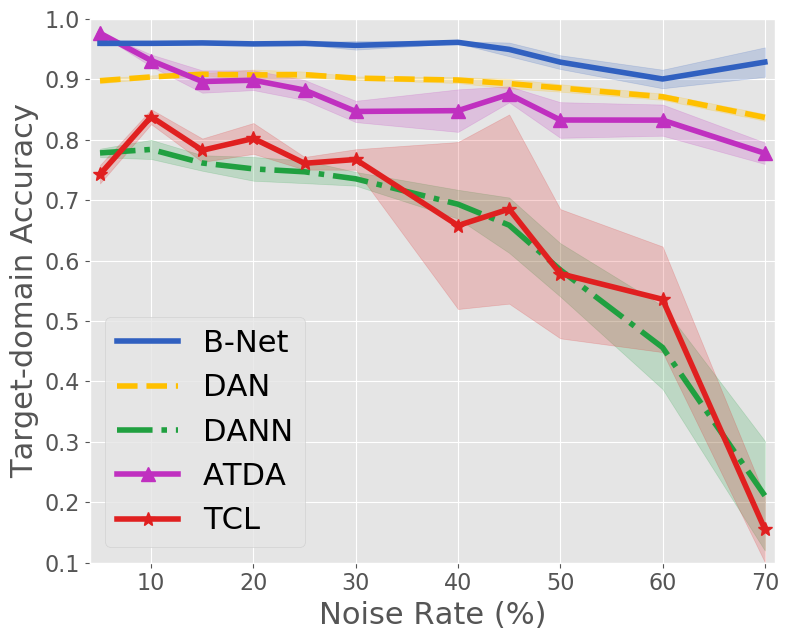} \includegraphics[width=0.24\textwidth]{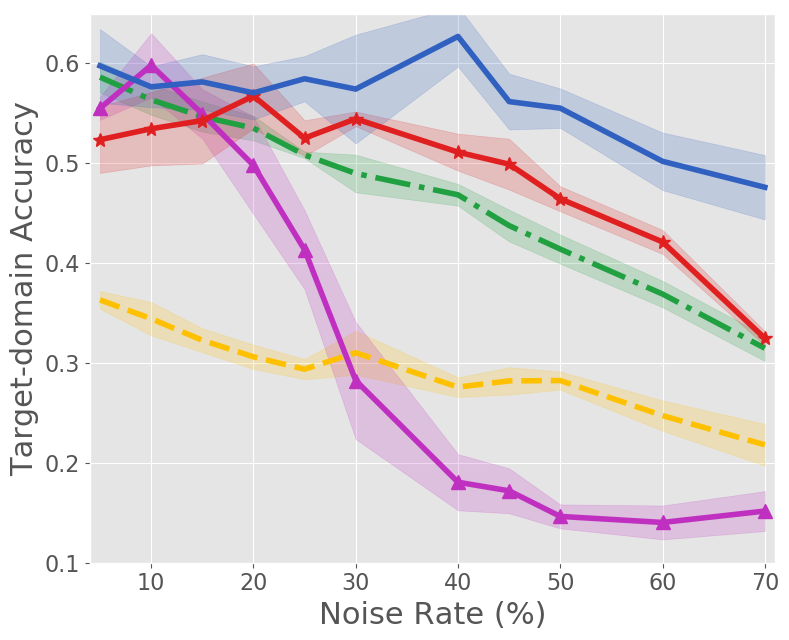}}
		\subfigure[Pair-flip noise: \emph{S}$\rightarrow$\emph{M} (left), \emph{M}$\rightarrow$\emph{S} (right)]
		{\includegraphics[width=0.24\textwidth]{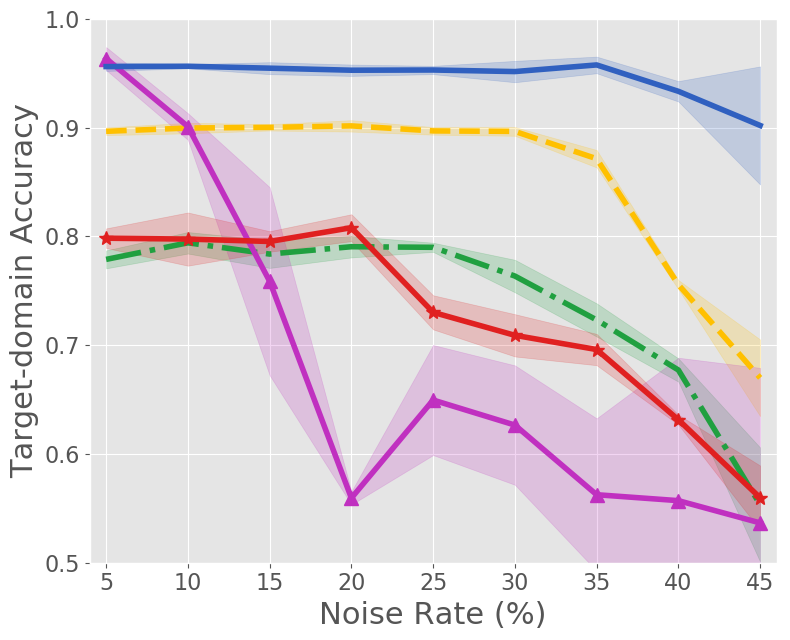} \includegraphics[width=0.24\textwidth]{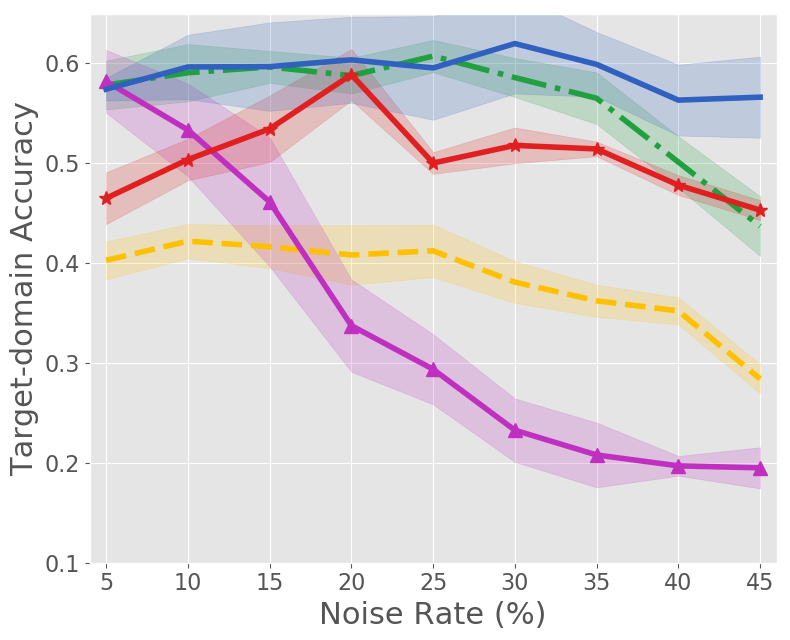}}
		\end{center}
		
		\vspace{-1em}
		\caption{{WUDA ruins representative UDA methods.} Representative UDA methods includes \emph{deep adaptation network} (DAN, an IPM based method \mbox{\cite{Long_DAN}}), \emph{domain-adversarial neural network} (DANN, an adversarial training based method \mbox{\cite{DANN_JMLR}}), \emph{asymmetric tri-training domain adaptation} (ATDA, a pseudo-label based method \mbox{\cite{KSaito_ICML17}}) and \emph{transferable curriculum learning} (TCL, a robust UDA method \mbox{\cite{TCL_long}}). B-Net is our proposed WUDA method. We report target-domain accuracy of all methods when the noise rate of source domain changes (a) from $5\%$ to $70\%$ (symmetry-flip noise) and (b) from $5\%$ to $45\%$ (pair-flip noise). Clearly, as the noise rate of source domain increases, the target-domain accuracy of representative UDA methods drops quickly while that of B-Net keeps stable consistently.}
		\label{fig:UDA_failure}
        \vspace{-1em}
\end{figure*}

However, in the wild, the data volume of the source domain tends to be large \cite{tan2014towards}. To avoid the expensive labeling cost, labeled data in the source domain normally come from amateur annotators or the Internet \cite{cleanNet,Data_web,test_bed_dataset}. This brings us a new, more realistic and more challenging problem, \emph{wildy unsupervised domain adaptation} (abbreviated as WUDA, Figure~\ref{fig: our_solution}). This adaptation aims to transfer knowledge from noisy labeled data in the source domain ($\widetilde{P}_s$, i.e., noisy source data) to unlabeled target data ($P_{x_t}$). Unfortunately, existing UDA methods share an implicit assumption that \textit{there are no noisy source data} \cite{yu2017transfer,TCL_long}. Namely, these methods focus on transferring knowledge from clean source data ($P_s$) to unlabeled target data ($P_{x_t}$). Therefore, these methods cannot well handle WUDA (Figure~\ref{fig:UDA_failure}).

To validate this fact, we empirically reveal the deficiency of existing UDA methods (Figure~\ref{fig:UDA_failure}, e.g., \emph{deep adaptation network} (DAN) \cite{Long_DAN} and \emph{domain-adversarial neural network} (DANN) \cite{DANN_JMLR}). To improve these methods, a straightforward solution is a two-step approach. In Figure~\ref{fig: our_solution}, we can first use label-noise algorithms to train a classifier on noisy source data, then leverage this trained classifier to assign pseudo labels for noisy source data. Via UDA methods, we can transfer knowledge from pseudo-labeled source data  ($\hat{P}_s$) to unlabeled target data ($P_{x_t}$). Nonetheless, pseudo-labeled source data are still noisy, and such two-step approach may not eliminate noise effects.

To circumvent the issue of two-step approach, we present a robust one-step approach called \textit{Butterfly}. In high level, Butterfly directly transfers knowledge from $\widetilde{P}_s$ to $P_{x_t}$, and uses the transferred knowledge to construct target-specific representations. In low level, Butterfly maintains four networks dividing two branches (Figure~\ref{fig: sketch_fig}): Two networks in Branch-I are jointly trained on noisy source data and pseudo-labeled target data (data in \emph{mixture domain} (MD)); while two networks in Branch-II are trained on pseudo-labeled target data. Our ablation study (see Section~\ref{sec:abl_study}) confirms the network design of Butterfly (see Section~\ref{sec:Butterfly_net}) is the optimal.

The reason why Butterfly can be robust takes root in the \textit{dual-checking principle} (DCP): Butterfly checks high-correctness data out, from not only the data in MD but also the pseudo-labeled target data. After cross-propagating these high-correctness data, Butterfly can obtain high-quality \emph{domain-invariant representations} (DIR) and \emph{target-specific representations} (TSR) simultaneously in an iterative manner. If we only check data in MD (i.e., B-Net-M in Section~\ref{sec:abl_study}), the error existed in pseudo-labeled target data will accumulate, leading to the low-quality DIR and TSR.

We conduct experiments on simulated WUDA tasks, including $4$ \emph{MNIST-to-SYND} tasks, $4$ \emph{SYND-to-MNIST} tasks and $24$ human-sentiment tasks. Besides, we conduct experiments on $3$ real-world WUDA tasks. Empirical results demonstrate that Butterfly can robustly transfer knowledge from noisy source data to unlabeled target data. Meanwhile, Butterfly performs much better than existing UDA methods when \emph{source domain} (SD) suffers the extreme (e.g., $45\%$) noise.

\section{{Literature Review}}

{This section reviews the existing UDA methods in detail. UDA methods train with clean source data and unlabeled target data to classify target-domain data, which mainly consist of three orthogonal techniques: \textit{integral probability metrics} (IPM) \mbox{\cite{Ghifary2017,Gong2016,Gretton2012,Wasserstein_distances,Long_DAN}}, \textit{adversarial training} \mbox{\cite{DANN_JMLR,Mingming_CGDAB,Judy_adversarial,Kun_zhang_CIAN,Saito:MCD,Judy_ADDA}} and \textit{pseudo labeling} \mbox{\cite{KSaito_ICML17}}.} 

{IPMs (such as maximum mean discrepancy \mbox{\cite{Gretton2012,liu2020learning}} and Wasserstein distance \mbox{\cite{Wasserstein_distances}}) are used to measure the discrepancy between distributions of two domains. By minimizing the IPM between two domains, models trained with clean source data can classify unlabeled target data accurately \mbox{\cite{Gong2016,Ghifary2017,Long_DAN}}. In this line, representative methods include conditional transferable components \mbox{\cite{Gong2016}}, scatter component analysis \mbox{\cite{Ghifary2017}} and DAN \mbox{\cite{Long_DAN}}.}

{Another technique is the adversarial training method inspired by the theory of domain adaptation \mbox{\cite{Ben_DA_bounds}}. This theory suggests that predictions must be based on features, and these features cannot be used to discriminate source and target domains \mbox{\cite{DANN_JMLR,Judy_adversarial,Judy_ADDA}}. For example, DANN considers two deep networks: one is used to construct new features that predict labels in the TD; while the other is to make two domains non-distinguishing based on these new features \mbox{\cite{DANN_JMLR}}. DANN simultaneously trains two deep networks to find domain-invariant representations between two domains.}

{The last technique is the pseudo-label method, which regards pseudo labels given by a classifier as true labels \mbox{\cite{KSaito_ICML17,Long_JDA}}. The \emph{joint domain adaptation} (JDA) matches joint distributions of two domains using these pseudo labels \mbox{\cite{Long_JDA}}. The \emph{asymmetric tri-training domain adaptation} (ATDA) leverages three networks asymmetrically \mbox{\cite{KSaito_ICML17}}. Specifically, two networks are used to annotate unlabeled target data, namely generating pseudo labels. The other network can obtain target-specific representations based on the pseudo-labeled data. Since pseudo-label UDA methods can effectively reduce the upper bound of expected risk in the TD \mbox{\cite{SourceDataFreeUDA20,KSaito_ICML17}}, we also consider using the pseudo-label technique to help address the WUDA problem (like ATDA \mbox{\cite{KSaito_ICML17}}).}

\section{Preliminary}

\label{Asec: noisy generation}
In this section, we introduce notations used in this paper and two common label-noise generation processes \cite{Tongliang_TPAMI,jiang2017mentornet}.

\subsection{Notations}
\label{sec:notations}
The following notations are used to demonstrate theoretical results of this paper. 
\begin{itemize}
    \item a space $\mathcal{X}\subset \mathbb{R}^d$ and $\mathcal{Y}=\{1,2,\dots,K\}$ as a label set;
    \item $f_t(x_t)$ and $\tilde{f}_t(x_t)$ represent the ground-truth and pseudo labeling function of the target domain, where $f_t,\tilde{f}_t: \mathcal{X} \rightarrow \mathcal{Y}$;
    \item $A = \{x:\tilde{f}_t(x)\neq f_t(x)\}$ and $B = \mathcal{X}/A$ represent the area where $\tilde{f}_t(x)\neq f_t(x)$ (the set $A$) and the area where $\tilde{f}_t(x) = f_t(x)$ (the set $B$);
    \item $\tilde{p}_s(x_s,\tilde{y}_s)$, $p_s(x_s,y_s)$ and $q_s(x_s,y_s)$ represent probability densities of noisy, correct and incorrect \emph{multivariate random variable} (m.r.v.) defined on $\mathcal{X}\times \mathcal{Y}$, respectively, and $\tilde{p}_{x_s}(x_s)$, $p_{x_s}(x_s)$ and $q_{x_s}(x_s)$ are their marginal densities on $\mathcal{X}$;
    \item $p_{x_t}(x_t)$ represents the probability density of m.r.v.~$X_t$ defined on $\mathcal{X}$;
    \item $q_{x_t}(x) = p_{x_t}(x)1_A(x)/P_{x_t}(A)$ represents the probability density of $X_t$ restricted in $A$;
    \item $p_{x_t}^{\prime}(x_t)=p_{x_t}(x_t)1_B(x_t)/P_{x_t}(B)$ represents the probability density of $X_t$ restricted in $B$;
    \item $\mathcal{H}$ is the class of arbitrary \emph{decision functions} $h: \mathcal{X} \rightarrow \mathbb{R}^K$;
    \item $\ell: \mathbb{R}^K\times \mathcal{Y} \rightarrow \mathbb{R}_{+}$ is the loss function. $\ell(t,y)$ means the loss incurred by predicting an output $t$ (e.g., $h(x)$) when the ground truth is $y$;
    \item $\mathbb{L}_\mathcal{H}=\{\ell(h(x),y)|h\in \mathcal{H},x\in \mathcal{X},y \in \mathcal{Y}\}$ is the class of loss functions associated with $\mathcal{H}$;
    \item {expected risks on the noisy m.r.v. and correct m.r.v.:}
    \begin{align*}
        \tilde{R}_s(h)&=\mathbb{E}_{\tilde{p}_s(x_s,\tilde{y}_s)}[\ell(h(x_s),\tilde{y}_s)], \\
        {R}_s(h)&=\mathbb{E}_{ {p}_s(x_s,{y}_s)}[\ell(h(x_s),{y}_s)];
    \end{align*}
    \item expected discrepancy (associated with $\ell$) between an arbitrary {decision function} $h$ and a ground-truth or pseudo labeling function $f$ ($f$ could be $f_t$ or $\tilde{f}_t$) under different marginal densities:
    \begin{align*}
        \tilde{R}_s(h,f)&=\mathbb{E}_{\tilde{p}_{x_s}(x_s)}[\ell(h(x_s),f(x_s))], \\
        R_s(h,f)&=\mathbb{E}_{p_{x_s}(x_s)}[\ell(h(x_s),f(x_s))], \\
        R_t(h,f)&=\mathbb{E}_{p_{x_t}(x_t)}[\ell(h(x_t),f(x_t))].
    \end{align*}
\end{itemize}

\subsection{Generating label-noise via the transition matrix}
\label{Asec: noisy generation_Q}
{We assume that there are clean source data denoted by a m.r.v. ($X_s,Y_s$) defined on $\mathcal{X}\times \mathcal{Y}$ with the probability density $p_s(x_s,y_s)$. However, samples of ($X_s,Y_s$) cannot be directly obtained and we can only observe noisy source data (denoted by m.r.v. ($X_s,\tilde{Y}_s$)) with the probability density $\tilde{p}_s(x_s,\tilde{y}_s)$ \mbox{\cite{Tongliang_TPAMI}}. $\tilde{p}_s(x_s,\tilde{y}_s)$ is generated from ${p}_s(x_s,{y}_s)$ and a transition matrix $Q_{ij}=\textnormal{Pr}(\tilde{Y}_s=j|Y_s = i)$. Each element in $Q$, $\textnormal{Pr}(\tilde{Y}_s=j|Y_s = i)$, is a transition probability, i.e., the flip rate from a correct label $i$ to a noisy label $j$.} 

\subsection{Generating label-noise via the sample selection}
\label{Asec: noisy generation_S}
{The transition matrix $Q$ is easily estimated in certain situations \mbox{\cite{Tongliang_TPAMI}}. However, in more complex situations, such as clothing1M dataset \mbox{\cite{Clothing1M_xiaodong}}, noisy source data is directly generated by selecting data from a pool, which mixes correct data (data with correct labels) and incorrect data (data with incorrect labels). Namely, how the correct label $i$ is corrupted to $j$ ($i\neq j$) is unclear.

Let $(X_s,Y_s,V_s)$ be a m.r.v. defined on $\mathcal{X}\times \mathcal{Y} \times \mathcal{V}$ with the probability density $p_s^{\text{po}}(x_s,y_s,v_s)$, where $\mathcal{V}=\{0,1\}$ is the \textit{perfect-selection random variable}. $V_s=1$ means ``correct'' and $V_s=0$ means ``incorrect''. Nonetheless, samples of $(X_s,Y_s,V_s)$ cannot be obtained and we can only observe $(X_s,\tilde{Y}_s)$ from a distribution with the following density.}
\begin{align}
\label{eq: noise_observ}
    \tilde{p}_s(x_s,\tilde{y}_s) = \sum_{v_s=0}^1p_{X_s,Y_s|V_s}^{\text{po}}(x_s,y_s|v_s)p_{V_s}^{\text{po}}(v_s),
\end{align}
{where $p_{V_s}^{\text{po}}(v_s) = \int_{\mathcal{X}}\sum_{y_s=1}^Kp_s^{\text{po}}(x_s,y_s,v_s)dx_s$. Eq.~({\ref{eq: noise_observ}}) means that we lose the information regarding $V_s$. If we uniformly draw samples from $\tilde{p}_s(x_s,\tilde{y}_s)$, the noise rate of these samples is $p_{V_s}^{\text{po}}(0)$. It is clear that the m.r.v. $(X_s,Y_s|V_s=1)$ is the m.r.v. $(X_s,Y_s)$ mentioned in Section~{\ref{Asec: noisy generation_Q}}. Then, $q_s(x_s,y_s)$ is used to describe the density of incorrect m.r.v. $(X_s,Y_s|V_s=0)$.
Using $p_s(x_s,y_s)$ and $q_s(x_s,y_s)$, $\tilde{p}_s(x_s,\tilde{y}_s)$ is expressed by}
\begin{align}
\label{eq: noise_observ_simple}
    \tilde{p}_s(x_s,\tilde{y}_s) = (1-\rho)p_s(x_s,y_s)+\rho q_s(x_s,y_s),
\end{align}
{where $\rho=p_{V_s}^{\text{po}}(0)$. To reduce noise effects from incorrect data, researchers aim to recover the information of $V_s$, i.e., to select the correct data \mbox{\cite{Co-teaching,jiang2017mentornet,DeCoupling}}.}

\section{Wildly Unsupervised Domain Adaptation}

In this section, we first define a new, more realistic and more challenging problem setting called \textit{wildly unsupervised domain adaptation (WUDA)}, and explain the nature of WUDA. Then, we empirically show that representative UDA methods cannot handle WUDA well, which motivates us to propose a novel method to address the WUDA problem (Section~\ref{sec:Butterfly_net}).

\begin{mypro}[{Wildly Unsupervised Domain Adaptation}]
\label{def-1}
{Let $X_t$ be a m.r.v. defined on the space $\mathcal{X}$ with respect to the probability density $p_{x_t}$, $(X_s,\widetilde{Y}_s)$ be a m.r.v. defined on the space $\mathcal{X}\times \mathcal{Y}$ with respect to the probability density $\tilde{p}_{s}$, where $\tilde{p}_{s}$ is the probability density regarding noisy source data (generated in Section~{\ref{Asec: noisy generation_Q}} or {\ref{Asec: noisy generation_S}}), and $\mathcal{Y}=\{1,\dots, K\}$ is the label set. Let $p_{x_s}$ be the marginal density of $\tilde{p}_{s}$. Given i.i.d. data $\tilde{D}_s=\{(x_{si},\tilde{y}_{si})\}_{i=1}^{n_s}$ and $D_t=\{x_{ti}\}_{i=1}^{n_t}$ drawn from $\tilde{P_s}$ and $P_{x_t}$ separately, in wildly unsupervised domain adaptation, we  aim to train with noisy source data $\tilde{D}_s$ and target data $D_t$ to accurately annotate data drawn from $P_{x_t}$, where $p_{x_s}\neq p_{x_t}$.}
\end{mypro}
\begin{myrem}\upshape
In Problem~\ref{def-1}, $\tilde{D}_s$ is noisy source data, $D_t$ is unlabeled target data, and $\tilde{P_s}$ and $P_{x_t}$ are two probability measures corresponding to densities $\tilde{p}_s(x_s,\tilde{y}_s)$ and $p_{x_t}(x_t)$. 
\end{myrem}

\begin{figure*}[!tp]
	\begin{center}
		{\includegraphics[width=0.7\textwidth]{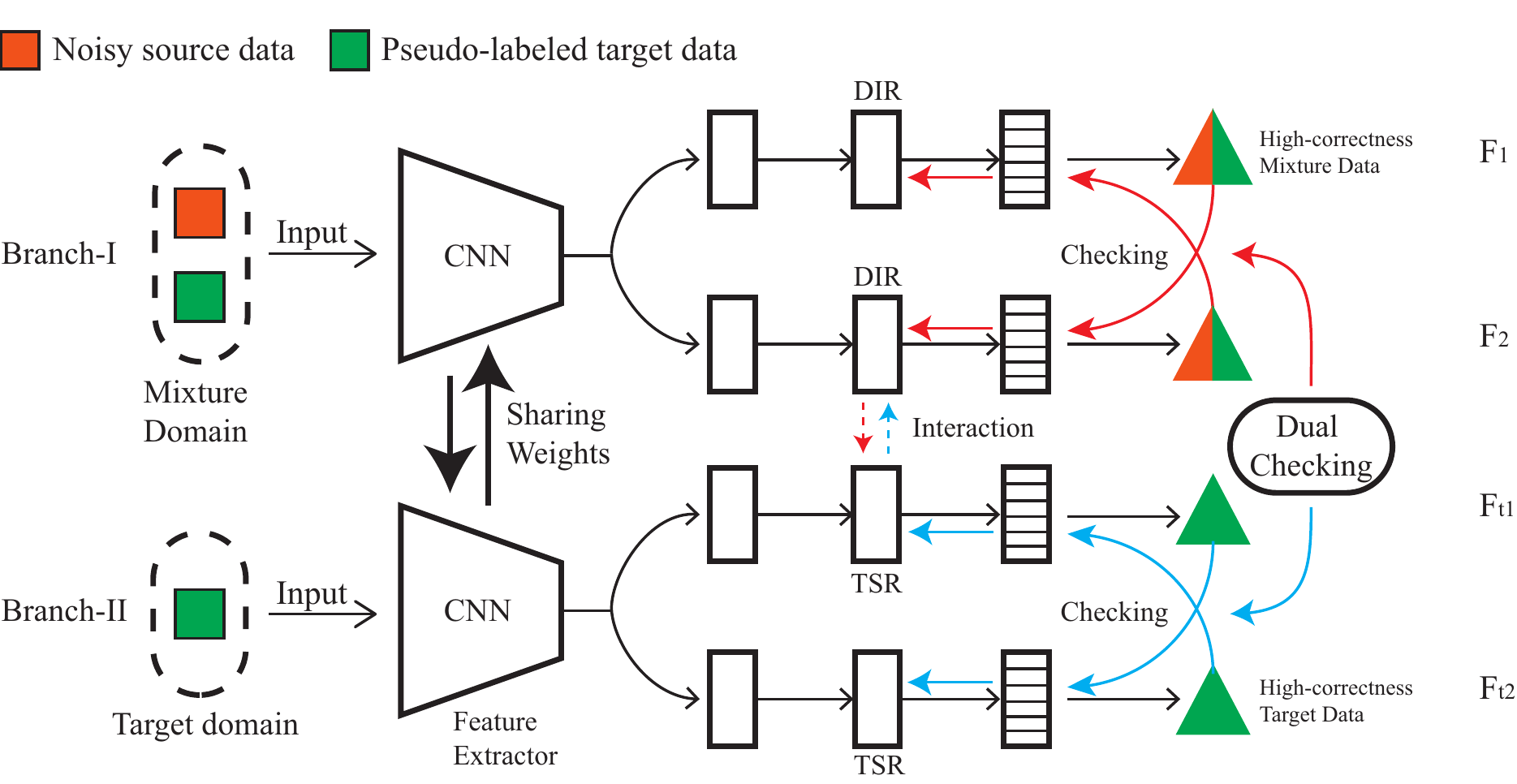}}
		\end{center}
		
		\vspace{-1em}
		\caption{Butterfly Framework. Two networks ($F_1$ and $F_2$) in Branch-I are jointly trained on noisy source data and pseudo-labeled target data (mixture domain). Two networks in Branch-II ($F_{t1}$ and $F_{t2}$) are trained on pseudo-labeled target data. By using the \emph{dual-checking principle} (DCP), Butterfly checks high-correctness data out from both mixture and pseudo-labeled target data. After cross-propagating checked data, Butterfly can obtain high-quality \emph{domain-invariant representations} (DIR) and \emph{target-specific representations} (TSR) simultaneously in an iterative manner (Algorithms {\ref{alg: Cross_update}} and {\ref{alg: ButterNET}}). {Note that DIR interacts with TSR via the shared CNN. Besides, in the first training epoch, since we do not have any pseudo-labeled target data, we use noisy source data as the pseudo-labeled target data, which follows \mbox{\cite{KSaito_ICML17}}.}}
		\label{fig: sketch_fig}
\end{figure*}

\subsection{Nature of WUDA} 
Specifically, there are five distributions involved in WUDA setting: 1) a marginal distribution on source data, i.e., $p_{x_s}$ in Problem~\ref{def-1}; 2) a marginal distribution on target data, i.e., $p_{x_t}$ in Problem~\ref{def-1}; 3) an incorrect conditional distribution of label given $x_s$, $q(y_s|x_s)$; 4) a correct conditional distribution of label given $x_s$, $p(y_s|x_s)$ and 5) a correct conditional distribution of label given $x_t$, $p(y_t|x_t)$.

Based on Problem~\ref{def-1} and Section \ref{Asec: noisy generation_S}, noisy source data $\tilde{D}_s$ are drawn from  $\tilde{p}_s(x_s,{y}_s)=p_{x_s}(x_s)((1-\rho)p(y_s|x_s)+\rho q(y_s|x_s))$, where $\rho$ is the noise rate in source data. Namely, source data $\tilde{D}_s$ are mixture of correct source data from $p_{x_s}(x_s)p(y_s|x_s)$ and incorrect data from $p_{x_s}(x_s)q(y_s|x_s)$. Target data ${D}_t$ are drawn from $p_{x_t}$. In WUDA setting, we aim to train a classifier with $\tilde{D}_s$ and ${D}_t$. This classifier is expected to accurately annotate data from $p_{x_t}$, i.e., to accurately simulate distribution 5).

{This paper considers WUDA under the common assumption used in the label-noise field, i.e., the $i^{th}$ element in the diagonal of the noise transition matrix is greater than other elements in the $i^{th}$ row or $i^{th}$ column of the noise transition matrix, where $i=1,2,\dots,K$ \mbox{\cite{Tongliang_TPAMI}}. Therefore, the proposed approach is able to solve any WUDA problem under the above assumption in principle. 
}

\subsection{WUDA ruins UDA methods} 

We take a simple example to illustrate the phenomenon that WUDA ruins representative UDA methods. In Section~\ref{Asec:theory:WUDA_ruins_UDA}, we theoretically analyze the reason of this phenomenon.

We corrupt source data using symmetry flipping \cite{Patrini_CVPR2017} and pair flipping \cite{Co-teaching} that are two representative ways to corrupt true labels. {Precise definitions of symmetry flipping ($Q_S$) and pair flipping ($Q_P$) are presented below, where $\rho$ is the noise rate and $K$ is the number of labels. }
\begin{align}
\label{eq:QS}
Q_S =
\begin{bmatrix}
    1-\rho & \frac{\rho}{K-1} & \dots  & \frac{\rho}{K-1} & \frac{\rho}{K-1}\\
    \frac{\rho}{K-1} & 1-\rho & \frac{\rho}{K-1} & \dots & \frac{\rho}{K-1}\\
    \vdots &  & \ddots &  & \vdots\\
    \frac{\rho}{K-1} & \dots & \frac{\rho}{K-1} & 1-\rho & \frac{\rho}{K-1}\\
    \frac{\rho}{K-1} & \frac{\rho}{K-1} & \dots  & \frac{\rho}{K-1} & 1-\rho
\end{bmatrix}_{K\times K},
\end{align}
\begin{align}
\label{eq:QP}
Q_P =
\begin{bmatrix}
    1-\rho & \rho & 0 & \dots  & 0\\
    0 & 1-\rho & \rho & & 0 \\
    \vdots &  & \ddots & \ddots & \vdots \\
    0 & & & 1- \rho & \rho \\
    \rho & 0 & \dots & 0 & 1-\rho \\
\end{bmatrix}_{K\times K}.
\end{align}
{For example, if $\rho=0.4$ and $K=11$, for the symmetry flipping, the probability that label ``0'' is corrupted to label ``1'' is $(1-\rho)/(K-1)=0.04$. For the pair flipping, the probability that label ``0'' is corrupted to label ``1'' is $\rho=0.4$.} To instantiate noisy source data and target data, we leverage \emph{MNIST} and \emph{SYND} (see Figure~\ref{fig:V_dig}), respectively (i.e., $K=10$).


We first construct two WUDA tasks with symmetry-flip noise: corrupted \emph{SYND}$\rightarrow$\emph{MNIST} (\emph{S}$\rightarrow$\emph{M}) and corrupted \emph{MNIST}$\rightarrow$\emph{SYND} (\emph{M}$\rightarrow$\emph{S}).
In Figure~\ref{fig:UDA_failure}-(a), we report accuracy of representative UDA methods on unlabeled target data, when the noise rate $\rho$ of SD changes from $5\%$ to $70\%$. It is clear that target-domain accuracy of these representative UDA methods drops quickly when $\rho$ increases. This means that WUDA ruins representative UDA methods. Then, we construct another two WUDA tasks with pair-flip noise. In Figure~\ref{fig:UDA_failure}-(b), we report target-domain accuracy, when the noise rate $\rho$ of SD changes from $5\%$ to $45\%$. Again, WUDA still ruins representative UDA methods. Note that, in practice,  pair-flip noise is much harder than symmetry-flip noise, the noise rate of pair-flip noise cannot be over $50\%$ \cite{Co-teaching}.
However, the proposed Butterfly network (abbreviated as B-Net, Figure~\ref{fig: sketch_fig}) performs robustly when $\rho$ increases (blue lines in Figure~\ref{fig:UDA_failure}). 

In Section~\ref{sec:ana_WUDA}, we will analyze the WUDA problem in theory and show why WUDA provably ruins all UDA methods and why the two-step approach is a compromise solution. Then, Section~\ref{sec:ana_address_WUDA} presents how to address the WUDA problem in principle. 

\section{Analysis of WUDA problem}
\label{sec:ana_WUDA}

{In this section, we analyze the WUDA problem from a theoretical view and show the difficulty of the WUDA problem. Completed proofs of lemmas and theorems are demonstrated in the Appendix. In the main content, we provide the main ideas of proving these theoretical results (i.e., \emph{Proof (sketch))}.} 

\subsection{WUDA provably ruins UDA methods}
\label{Asec:theory:WUDA_ruins_UDA}

{Theoretically, we show that existing UDA methods cannot directly transfer useful knowledge from $\tilde{D}_s$ to $D_t$. We first present the relation between $R_s(h)$ and $\tilde{R}_s(h)$.}

\begin{mythm}
\label{thm: risks relation}
If $\tilde{p}_s(x_s,\tilde{y}_s)$ is generated by a transition matrix $Q$ as demonstrated in Section~\ref{Asec: noisy generation_Q}, we have
\begin{equation}
\label{eq: risks relation_Q_TM}
    \tilde{R}_s(h) = R_s(h) + \mathbb{E}_{p_{x_s}(x_s)}[\bm{\eta}^T(x_s)(Q-I)\bm{\ell}(h(x_s))],
\end{equation}
where $\bm{\ell}(h(x_s)) = [\ell(h(x_s),1),..., \ell(h(x_s),K)]^T$ and $\bm{\eta}(x_s)=[p_{Y_s|X_s}(1|x_s),..., p_{Y_s|X_s}(K|x_s)]^T$. If $\tilde{p}_s(x_s,\tilde{y}_s)$ is generated by sample selection as described in in Section~\ref{Asec: noisy generation_S}, we have
\begin{equation}
\label{eq: risks relation}
    \tilde{R}_s(h) = (1-\rho)R_s(h) + \rho\mathbb{E}_{q_{x_s}(x_s)}[\bm{\eta_q}^T(x_s)\bm{\ell}(h(x_s))],
\end{equation}
where $\bm{\eta_q}(x_s)=[q_{Y_s|X_s}(1|x_s),..., q_{Y_s|X_s}(K|x_s)]^T$. 
\end{mythm}

\begin{proofskt}
For Eq.~(\ref{eq: risks relation_Q_TM}), we can prove it using the definition of the transition matrix defined in Section~\ref{Asec: noisy generation_Q} and the fact $\tilde{p}_s(x_s,\tilde{y}_s) = \tilde{p}_{\tilde{Y}_s|X_s}(\tilde{y}_s|x_s)p_{x_s}(x_s)$. For Eq.~\eqref{eq: risks relation}, we can prove it using Eq.~\eqref{eq: noise_observ_simple} and the definition of $R_s(h)$.
\end{proofskt}

\begin{myrem}\upshape
\label{rem: tremondous noise assumption}
In Eq.~(\ref{eq: risks relation}), $\mathbb{E}_{q_{x_s}(x_s)}[\bm{\eta_q}^T(x_s)\bm{\ell}(h(x_s))]$ represents the expected risk on the incorrect m.r.v.. To ensure to obtain useful knowledge from $\tilde{P}_s$, we need to avoid $\tilde{R}_s(h)\approx\mathbb{E}_{q_{x_s}(x_s)}$ $[\bm{\eta_q}^T(x_s)\bm{\ell}(h(x_s))]$. 
Specifically, we assume: there is a constant $0<M_s<\infty$ such that $\mathbb{E}_{q_{x_s}(x_s)}[\bm{\eta_q}^T(x_s)\bm{\ell}(h(x_s))]\leq R_s(h)+M_s$.
\end{myrem}

{Theorem {\ref{thm: risks relation}} shows that $\tilde{R}_s(h)$ equals ${R}_s(h)$ if only two cases happen: 1) $Q=I$ and $\rho=0$, or 2) some special combinations (e.g., special $p_{x_s}$, $q_{x_s}$, $Q$, $\mathbf{\eta}$ and $\ell$) make the second term in Eq.~({\ref{eq: risks relation_Q_TM}}) equal zero or make the second term in Eq.~({\ref{eq: risks relation}}) equal $\rho R_s(h)$. Case 1) means that source data are clean, which is not real in the wild. Case 2) rarely happens, since it is difficult to find such special combinations when $p_{x_s}$, $q_{x_s}$, $Q$ and $\mathbf{\eta}$ are unknown. As a result, $\tilde{R}_s(h)$ has an essential difference with $R_s(h)$. Then, following the proof skills in \mbox{\cite{Ben_DA_bounds}}, we derive the upper bound of $R_t(h)$ as below.}

\begin{mythm}
\label{thm:upper_bound_target}
For any $h\in\mathcal{H}$, we have
\begin{align}
\label{eq:risk bound population}
    R_t(h,f_t) &\leq \underbrace{\tilde{R}_s(h)}_{(i)~\textbf {noisy-data risk}} +~~~~ \underbrace{|R_t(h,\tilde{f}_t) - \tilde{R}_s(h,\tilde{f}_t)|}_{(ii)~\textbf {discrepancy~between~distributions}}\nonumber \\
    &~~~~+~ \underbrace{|R_s(h,\tilde{f}_t) - R_s(h)|}_{(iii)~\textbf {domain dissimilarity}} \nonumber \\
    &~~~~+~\underbrace{|\tilde{R}_s(h)-R_s(h)|+|\tilde{R}_s(h,\tilde{f}_t) - R_s(h,\tilde{f}_t)|}_{(iv)~\textbf {noise~ effects~from~source~$\Delta_s$}}~\nonumber \\
    &~~~~+~ \underbrace{|R_t(h,f_t) -  R_t(h,\tilde{f}_t)|}_{(v)~\textbf {noise~ effects~from~target~$\Delta_t$}}.
\end{align}
\end{mythm}
\begin{proofskt}
For any $h\in\mathcal{H}$, we have
\begin{align*}
    &~~~~~R_t(h,f_t) \nonumber \\
    &= R_t(h,f_t) + \tilde{R}_s(h) - \tilde{R}_s(h) + R_s(h,f_t) - R_s(h,f_t) \nonumber \\
    &=\tilde{R}_s(h) + R_t(h,f_t) - \tilde{R}_s(h,f_t) + R_s(h,f_t) - R_s(h)  \nonumber \\
    &~~~~~+R_s(h)- \tilde{R}_s(h)+ \tilde{R}_s(h,f_t) - R_s(h,f_t).
\end{align*}
Since we do not know $f_t$, we substitute the following equations into the above equation,
\begin{align*}
    R_t(h,f_t) = R_t(h,\tilde{f}_t) + R_t(h,f_t) - R_t(h,\tilde{f}_t), \\
    \tilde{R}_s(h,f_t) = \tilde{R}_s(h,\tilde{f}_t) + \tilde{R}_s(h,f_t) - \tilde{R}_s(h,\tilde{f}_t), \\
    R_s(h,f_t) = R_s(h,\tilde{f}_t) + R_s(h,f_t) - R_s(h,\tilde{f}_t),
\end{align*}
which proves this theorem.
\end{proofskt}
\begin{myrem}\upshape
\label{rem: tremondous_noise_assumption_f}
To ensure that we can gain useful knowledge from  $\tilde{f}_t(x_t)$, we assume: there is a constant $0<M_t<\infty$ such that $\mathbb{E}_{q_{x_s}(x)}[\ell(h(x),\tilde{f}_t(x))]\leq R_s(h,\tilde{f}_t)+M_t$ and  $\mathbb{E}_{q_{x_t}(x)}[\ell(h(x),\tilde{f}_t(x))]\leq R_t(h,f_t)+M_t$. {Since we do not have labels in the target domain, we also assume that there exists an $h\in\mathcal{H}$ such that $R_t(h,f_t)+{R}_s(h)$ is a small value. This assumption follows common assumption of UDA problem \mbox{\cite{Ben_DA_bounds}} and ensures that the adaptation is possible.} 
\end{myrem}


It is clear that the upper bound of $R_t(h,f_t)$, shown in Eq.~(\ref{eq:risk bound population}), has $5$ terms.
However, existing UDA methods only focus on minimizing $(i)$ + $(ii)$ \cite{DANN_JMLR,Ghifary2017,Long_DAN} or $(i)$  + $(ii)$ + $(iii)$ \cite{KSaito_ICML17}, which ignores terms $(iv)$ and $(v)$ (i.e., $\Delta = \Delta_s + \Delta_t$). Thus, existing UDA methods cannot handle WUDA well.

\subsection{Two-step approach is a compromise solution} \label{sec:two-step}
To reduce noise effects, a straightforward solution is two-step approach. For example, in the first step, we can train a classifier with noisy source data using co-teaching \cite{Co-teaching} and use this classifier to annotate pseudo labels for source data. In the second step, we use ATDA \cite{KSaito_ICML17} to train a target-domain classifier with pseudo-labeled-source data and pseudo-labeled target data.

Nonetheless, the pseudo-labeled source data are still noisy. Let labels of noisy source data $\tilde{y}_s$ be replaced with pseudo labels $\tilde{y}^{\prime}_s$ after using co-teaching. Noise effects $\Delta$ will become pseudo-label effects $\Delta_p$ as follows.
\begin{equation}
\setlength{\abovedisplayskip}{5pt}
\setlength{\belowdisplayskip}{5pt}
\label{eq: pseudo-label effects}
    \Delta_p = \underbrace{|\tilde{R}'_s(h)-R_s(h)|+|\tilde{R}'_s(h,\tilde{f}_t) - R_s(h,\tilde{f}_t)|}_{\textbf {pseudo-labeled-source~effects~$\Delta_s^{\prime}$}}+\Delta_t,
\end{equation}
where $\tilde{R}'_s(h)$ and $\tilde{R}'_s(h,\tilde{f}_t)$ correspond to $\tilde{R}_s(h)$ and $\tilde{R}_s(h,\tilde{f}_t)$ in $\Delta_s$. It is clear that the difference between $\Delta_p$ and $\Delta$ is $\Delta_s^{\prime}-\Delta_s$. {The left term in $\Delta_s^{\prime}$ may be less than that in $\Delta_s$ due to a label-noise algorithm (e.g., co-teaching \mbox{\cite{Co-teaching}}), but the right term in $\Delta_s^{\prime}$ may be higher than that in $\Delta_s$ since a label-noise algorithm does not consider minimizing it.} Thus, it is hard to say whether $\Delta_s^{\prime}<\Delta_s$ (i.e., $\Delta_p<\Delta$). This means that two-step approach may not really reduce noise effects.

\section{How to address WUDA in principle}
\label{sec:ana_address_WUDA}

\label{sec:butter_eliminate_noise}
To eliminate noise effects $\Delta$, we aim to select correct data simultaneously from noisy source data and pseudo-labeled target data.
In theory, we prove that noise effects will be eliminated if we can select correct data with a high probability. Let $\rho_{01}^s$ represent the probability that incorrect data is selected from noisy source data, and $\rho_{01}^t$ represent the probability that incorrect data is selected from pseudo-labeled target data. Theorem \ref{thm: epsilon_effects} shows that $\Delta\rightarrow0$ if $\rho^s_{01}\rightarrow0$ and $\rho^t_{01}\rightarrow0$ and presents a new upper bound of $R_t(h,f_t)$. 
Before stating Theorem~\ref{thm: epsilon_effects}, we first present two m.r.v.s below.
\begin{itemize}
    \item $(X_s,Y_s,V_s)$ defined on $\mathcal{X}\times \mathcal{Y} \times \mathcal{V}$ with the probability density $p_s^{\text{po}}(x_s,y_s,v_s)$, where $\mathcal{V}=\{0,1\}$;
    \item $(X_t,V_t)$ defined on $\mathcal{X} \times \mathcal{V}$ with the probability density ${p}_t^{\text{po}}(x_t,v_t)$, where $\mathcal{V}=\{0,1\}$. $p_{V_t}^{\text{po}}(v_t)$ is the marginal density of ${p}_t^{\text{po}}(x_t,v_t)$.
\end{itemize}

The $V_s$ has been introduced in Section~\ref{Asec: noisy generation_S}. Similar with $V_s$, $V_t$ is also a \emph{perfect-selection random variable}. Data drawn from the distribution of $(X_t,V_t)$ can be regarded as a pool that mixes the correct ($v_t=1$) and incorrect ($v_t=0$) pseudo-labeled target data. 
Namely, $V_t=1$ means $f_t(x_t)=\tilde{f}_t(x_t)$ and $V_t=0$ means $f_t(x_t)\neq \tilde{f}_t(x_t)$. It is clear that, higher value of $p_{V_t}^{\text{po}}(V_t=1)$ means that $\tilde{f}_t$ is more like $f_t$. In following, we use $\rho_{v_t}$ to represent $p_{V_t}^{\text{po}}(v_t=0)$. Note that both perfect-selection random variables $V_s$ and $V_t$ cannot be observed and we can only observe following m.r.v.s.

\begin{itemize}
    \item $(X_s,Y_s,U_s)$ defined on $\mathcal{X}\times \mathcal{Y} \times \mathcal{V}$ with the probability density $\tilde{p}_s^{\text{po}}(x_s,y_s,u_s)$;
    \item $(X_t,U_t)$ defined on $\mathcal{X} \times \mathcal{V}$ with the probability density $\tilde{p}_t^{\text{po}}(x_t,u_t)$. $\tilde{p}_{U_t}^{\text{po}}(u_t)$ is the marginal density of $\tilde{p}_t^{\text{po}}(x_t,u_t)$.
\end{itemize}

The $U_s$ and $U_t$ are \textit{algorithm-selection random variables}. Data drawn from the distribution of $(X_s,Y_s,U_s)$ can be regarded as a pool that mixes the selected ($u_s=1$) and unselected ($u_s=0$) noisy source data. Data drawn from the distribution of $(X_t,U_t)$ can be regarded as a pool that mixes the selected ($u_t=1$) and unselected ($u_t=0$) pseudo-labeled target data. 
We can obtain observations of $(X_s,Y_s,U_s)$ and $(X_t,U_t)$ using an algorithm that is used to select correct data. After executing the algorithm, we can obtain observations $\{x_{si},\tilde{y}_{si},u_{si}\}_{i=1}^{n_s}$ and $\{x_{ti},u_{ti}\}_{i=1}^{n_t}$. Based on $(X_s,Y_s,U_s)$ and $(X_t,U_t)$, we can define the following expected risks. 
\begin{align*}
    &\tilde{R}^{\text{po}}_s(h,u_s)=(1-\rho_{u_s})^{-1}\mathbb{E}_{\tilde{p}_s^{\text{po}}(x_s,y_s,u_s)}[u_s\ell(h(x_s),y_s)], \\
    &\tilde{R}^{\text{po}}_t(h,\tilde{f}_t,u_t)=(1-\rho_{u_t})^{-1}\mathbb{E}_{\tilde{p}_t^{\text{po}}(x_t,u_t)}[u_t\ell(h(x_t),\tilde{f}_t(x_t))], \\
    &\tilde{R}^{\text{po}}_s(h,\tilde{f}_t,u_s)=(1-\rho_{u_s})^{-1}\mathbb{E}_{\tilde{p}_s^{\text{po}}(x_s,y_s,u_s)}[u_s\ell(h(x_s),\tilde{f}_t(x_s))].
\end{align*}
where $\rho_{u_s}=\tilde{p}_{U_s}^{\text{po}}(u_s=0)$ and $\rho_{u_t}=\tilde{p}_{U_t}^{\text{po}}(u_t=0)$. Since we can observe $(X_s,Y_s,U_s)$ and $(X_t,U_t)$, the empirical estimators of these three risks can be easily computed. 
Then, we define following probabilities to describe the relation between perfect-selection random variables and algorithm-selection random variables, where $i,j=0,1$.
\begin{itemize}
    \item $\rho_{ji}^s=\textnormal{Pr}(V_s=j|U_s=i)$ represents the probability of the event: $V_s=j$ given $U_s=i$,
    \item $\rho_{ji}^t=\textnormal{Pr}(V_t=j|U_t=i)$ represents the probability of the event: $V_t=j$ given $U_t=i$.
\end{itemize}

\begin{myrem}\upshape
Based on above definitions, we know that 1) $\rho_{01}^s$ is the probability that incorrect data is selected from noisy source data, and 2) $\rho_{01}^t$ is the probability that incorrect data is selected from pseudo-labeled target data. 
\end{myrem} 
Using $\rho_{ji}^s$ and $\rho_{ji}^t$, we can show the relation between probability densities of $(X_s,Y_s|V_s)$ and $(X_s,Y_s|U_s)$, and the relation between probability densities of $(X_t|V_t)$ as follows. 
\begin{align*}
    &\tilde{p}_{X_s,Y_s|U_s}^{\text{po}}(x_s,y_s|i) =~ \rho_{0i}^s{p}_{X_s,Y_s|V_s}^{\text{po}}(x_s,y_s|0) \\
    &~~~~~~~~~~~~~~~~~~~~~~~~~~~~~~~~+ \rho_{1i}^s{p}_{X_s,Y_s|V_s}^{\text{po}}(x_s,y_s|1), \\
    &\tilde{p}_{X_t|U_t}^{\text{po}}(x_t|i) =~\rho_{0i}^t{p}_{X_t|V_t}^{\text{po}}(x_t|0)+ \rho_{1i}^t{p}_{X_t|V_t}^{\text{po}}(x_t|1).
\end{align*}
Since 
\begin{align*}
    & {p}_{X_s,Y_s|V_s}^{\text{po}}(x_s,y_s|1) = p_s(x_s,y_s), \\
    & {p}_{X_s,Y_s|V_s}^{\text{po}}(x_s,y_s|0) = q_s(x_s,y_s), \\
    &{p}_{X_t|V_t}^{\text{po}}(x_t|0) =p_{x_t}(x_t)1_A(x_t)/P_{x_t}(A) = q_{x_t}(x_t), \\
    &{p}_{X_t|V_t}^{\text{po}}(x_t|1) =p_{x_t}(x_t)1_B(x_t)/P_{x_t}(B) = p_{x_t}^{\prime}(x_t),
\end{align*}
we have
\begin{align}
\label{eq: transit_u_s}
    \tilde{p}_{X_s,Y_s|U_s}^{\text{po}}(x_s,y_s|i) = \rho_{0i}^sq_s(x_s,y_s)+ \rho_{1i}^sp_s(x_s,y_s), 
\end{align}
\begin{align}
\label{eq: transit_u_t}
    \tilde{p}_{X_t|U_t}^{\text{po}}(x_t|i) =  \rho_{0i}^tq_{x_t}(x_t)+ \rho_{1i}^tp_{x_t}^{\prime}(x_t).
\end{align}
\begin{myrem}\upshape
Eq.~\eqref{eq: transit_u_s} and Eq.~\eqref{eq: transit_u_t} show that, if $\rho_{01}^s\rightarrow 0$ and $\rho_{01}^t\rightarrow 0$, we have 1) $\tilde{p}_{X_s,Y_s|U_s}^{\text{po}}(x_s,y_s|1)\rightarrow p_s(x_s,y_s)$ and 2)  $\tilde{p}_{X_t|U_t}^{\text{po}}(x_t|1)\rightarrow p_{x_t}^{\prime}(x_t)$. However, we cannot prove the main theorem (Theorem~\ref{thm: epsilon_effects}) using 1) and 2), since we only take care of risks instead of densities (like 1) and 2)).
\end{myrem}

Next, we present a lemma to show the relation between $\tilde{R}_s^{\text{po}}(h,u_s)$ and $R_s(h)$.
\begin{mylem}
\label{lem: selection discrepancy}
Given the m.r.v. $(X_s,Y_s,U_s)$ with the probability density $\tilde{p}^{\text{po}}_s(x_s,y_s,u_s)$ and Eq.~(\ref{eq: transit_u_s}), we have
\begin{align}
\label{eq: selection_risk}
    &|\tilde{R}_s^{\text{po}}(h,u_s)-R_s(h)|\nonumber \\
    \leq &\rho_{01}^s\max \{\mathbb{E}_{q_s(x_s,y_s)}[\ell(h(x_s),y_s)], R_s(h)\}.
\end{align}
\end{mylem}
\begin{proofskt}
Based on definition of $\tilde{R}_s^{\text{po}}(h,u_s)$ and the fact $\tilde{p}_s^{\text{po}}(x_s,y_s,u_s) = \tilde{p}_{X_s,Y_s|U_s}^{\text{po}}(x_s,y_s|1)\tilde{p}_{U_s}^{\text{po}}(1)$, $\tilde{R}_s^{\text{po}}(h,u_s)$ equals
\begin{align*}
    \frac{\int_\mathcal{X}\sum_{y_s=1}^K\ell(h(x_s),y_s)\tilde{p}_{X_s,Y_s|U_s}^{\text{po}}(x_s,y_s|1)\tilde{p}_{U_s}^{\text{po}}(1)dx_s}{1-\rho_{u_s}} 
\end{align*}
Then, we can use the definition of $\rho_{u_s}$ and the Eq.~\eqref{eq: transit_u_s} to prove this lemma.
\end{proofskt}
Similar with Lemma \ref{lem: selection discrepancy}, we can obtain
\begin{align}
\label{eq: selection_risk_ts}
    &|\tilde{R}_s^{\text{po}}(h,\tilde{f}_t,u_s)-R_s(h,\tilde{f}_t)|\nonumber \\
    \leq &\rho^s_{01}\max \{\mathbb{E}_{q_{x_s}(x_s)}[\ell(h(x_s),\tilde{f}_t(x_s))], R_s(h,\tilde{f}_t)\}.
\end{align}
Then, we give another lemma to show relation between $\tilde{R}_t^{\text{po}}(h,\tilde{f}_t,u_t)$ and $R_t(h,\tilde{f}_t)$.

\begin{mylem}
\label{lem: selection discrepancy_t}
Given the m.r.v. $(X_t,U_t)$ with the probability density $\tilde{p}^{\text{po}}_s(x_t,u_t)$ and Eq.~(\ref{eq: transit_u_t}), if $\mathbb{E}_{p_{x_t}^{\prime}(x_t)}[\ell(h(x_t),f_t(x_t))]\leq R_t(h,f_t)+\rho_{01}^sM_t$, then we have
\begin{align}
\label{eq: selection_risk_t}
    &|\tilde{R}_t^{\text{po}}(h,\tilde{f}_t,u_t)-R_t(h,{f}_t)|\nonumber \\
    \leq &\rho^t_{01} \max\{\mathbb{E}_{q_{x_t}(x_t)}[\ell(h(x_t),\tilde{f}_t(x_t))],R_t(h,f_t)\}+\rho_{11}^t\rho_{01}^sM_t.
\end{align}
\end{mylem}

\begin{proofskt}
According to definition of $\tilde{R}_t^{\text{po}}(h,\tilde{f}_t,u_t)$, we can unfold it to be
\begin{align*}
     &~~~~~\tilde{R}_t^{\text{po}}(h,\tilde{f}_t,u_t)\nonumber \\
    & = (1-\rho_{u_t})^{-1}\int_\mathcal{X}\ell(h(x_t),\tilde{f}_t(x_t))\tilde{p}_{X_t|U_t}^{\text{po}}(x_t|1)\tilde{p}_{U_t}^{\text{po}}(1)dx_t.
\end{align*}
Then, using the definition of $\rho_{u_s}$, Eq.~(\ref{eq: transit_u_s}), the definition of $V_t$  ($f_t(x_t)=\tilde{f}_t(x_t)$ when $V_t=1$) and the assumption that $\mathbb{E}_{p_{x_t}^{\prime}(x_t)}[\ell(h(x_t),f_t(x_t))]\leq R_t(h,f_t)+\rho_{01}^sM_t$, we have
\begin{align*}
    &~~~~\tilde{R}_t^{\text{po}}(h,\tilde{f}_t,u_t)\nonumber \\
    &\leq \rho_{01}^t\mathbb{E}_{q_{x_t}(x_t)}[\ell(h(x_t),\tilde{f}_t(x_t))] + \rho_{11}^t(R_t(h,f_t)+\rho_{01}^sM_t).
\end{align*}
Finally, we can upper bound $|\tilde{R}_t^{\text{po}}(h,\tilde{f}_t,u_t)-R_t(h,{f}_t)|$ using the above inequality, which proves this lemma.
\end{proofskt}

\begin{myrem}\upshape
\label{rem:last_assumption}
In Lemma \ref{lem: selection discrepancy_t}, $\mathbb{E}_{p_{x_t}^{\prime}(x_t)}[\ell(h(x_t),f_t(x_t))]\leq R_t(h,f_t)+\rho_{01}^sM_t$ means that the expected risk restricted in $B$ (i.e., $\mathbb{E}_{p_{x_t}^{\prime}(x_t)}[\ell(h(x_t),f_t(x_t))]$) can represent the true risk $R_t(h,f_t)$ when $\rho_{01}^s$ is small. 
If this assumption fails, we cannot gain useful knowledge from $\tilde{f}_t$ even when we can select correct data from pseudo-labeled target data ($\rho_{01}^t=0$).
\end{myrem}

Inequalities (\ref{eq: selection_risk}), (\ref{eq: selection_risk_ts}) and (\ref{eq: selection_risk_t}) show that if we can perfectly avoid annotating incorrect data as ``correct'' (i.e., $\rho^s_{01}=0$ and $\rho^t_{01}=0$), we have $\tilde{R}_s^{\text{po}}(h,u_s) = R_s(h)$, $\tilde{R}_s^{\text{po}}(h,\tilde{f}_t,u_t)=R_s(h,\tilde{f}_t)$ and $\tilde{R}_t^{\text{po}}(h,\tilde{f}_t,u_t)=R_t(h,{f}_t)$. Nonetheless, $\rho^s_{01}$ and $\rho^t_{01}$ never equal zero, and $\mathbb{E}_{q_s(x_s,y_s)}[\ell(h(x),y)]$, $\mathbb{E}_{q_{x_s}(x_s)}[\ell(h(x_s),\tilde{f}_t(x_s))]$ and $\mathbb{E}_{q_{x_t}(x_t)}[\ell(h(x_t),\tilde{f}_t(x_t))]$ may equal $+\infty$ for some $h\in \mathcal{H}$. Namely, even when $\rho^s_{01}$ and $\rho^t_{01}$ are very small, $\tilde{R}_s^{\text{po}}(h,u_s)$ is probably far away from $R_s(h)$. Thus, without proper assumptions,it is useless to use $(X_s,Y_s,U_s)$ to represent $(X_s,Y_s|V_s=1)$.

In Theorem \ref{thm: epsilon_effects}, we prove that, under assumptions in Remarks \ref{rem: tremondous noise assumption}, \ref{rem: tremondous_noise_assumption_f} and Lemma~\ref{lem: selection discrepancy_t}, $\tilde{R}_s^{\text{po}}(h,u_s) \rightarrow R_s(h)$, $\tilde{R}_s^{\text{po}}(h,\tilde{f}_t,u_t)\rightarrow R_s(h,\tilde{f}_t)$ and $\tilde{R}_t^{\text{po}}(h,\tilde{f}_t,u_t)\rightarrow R_t(h,{f}_t)$ if $\rho^s_{01} \rightarrow 0$ and $\rho^t_{01} \rightarrow 0$. Moreover, we give a new upper bound of $R_t(h,f_t)$. In the new upper bound, we show that: $\Delta\rightarrow0$ if $\rho^s_{01}\rightarrow0$ and $\rho^t_{01}\rightarrow0$.

\begin{mythm}
\label{thm: epsilon_effects}
Given two m.r.v.s $(X_s,Y_s,U_s)$ defined on $\mathcal{X} \times \mathcal{Y} \times \mathcal{V}$ and $(X_t,U_t)$ defined on $\mathcal{X} \times \mathcal{V}$, under the assumptions in Remark \ref{rem: tremondous noise assumption}, Remark \ref{rem: tremondous_noise_assumption_f} and Lemma~\ref{lem: selection discrepancy_t}, $\forall \epsilon \in (0,1)$, there are $\delta_s$ and $\delta_t$, if $\rho^s_{01}<\delta_s$ and $\rho^t_{01}<\delta_t$, for any $h\in\mathcal{H}$, we will have
\begin{equation}
\setlength{\abovedisplayskip}{4pt}
\setlength{\belowdisplayskip}{4pt}
    \label{eq:3epsilon}
    |\tilde{R}^{\text{po}}_s(h,\tilde{f}_t,u_s) -R_s(h,\tilde{f}_t)| + |\tilde{R}^{\text{po}}_s(h,u_s)-R_s(h)|< 2\epsilon.
\end{equation}
Moreover, we will have
\begin{align}
\label{eq:risk bound NEW}
    R_t(h,f_t) &\leq \underbrace{\tilde{R}^{\text{po}}_s(h,u_s)}_{(i)~\textbf {noisy-data risk}} +~\underbrace{|\tilde{R}^{\text{po}}_t(h,\tilde{f}_t,u_t) - \tilde{R}^{\text{po}}_s(h,\tilde{f}_t,u_s)|}_{(ii)~\textbf {discrepancy~between~distributions}} \nonumber \\
    &~~~~+~ \underbrace{|R_s(h,\tilde{f}_t) - R_s(h)|}_{(iii)~\textbf {domain dissimilarity}}+~\underbrace{2\epsilon}_{(iv)~\textbf {noise~effects ~$\Delta_s$}} \nonumber \\
    &~~~~+~\underbrace{2\epsilon}_{(iv)~\textbf {noise~effects ~$\Delta_t$}}.
\end{align}
\end{mythm}

\begin{proof}
We first prove upper bounds of $|\tilde{R}_s^{\text{po}}(h,u_s)-R_s(h)|$, $|\tilde{R}_s^{\text{po}}(h,\tilde{f}_t,u_t)- R_s(h,\tilde{f}_t)|$ and $|\tilde{R}_t^{\text{po}}(h,\tilde{f}_t,u_t)-R_t(h,{f}_t)|$ under assumptions in Theorem \ref{thm: epsilon_effects}.
Based on Lemma \ref{lem: selection discrepancy},
\begin{align}
\label{eq:first2_epsilon}
    &~~~~|\tilde{R}_s^{\text{po}}(h,u_s) - R_s(h)|\nonumber \\
    & = |\rho_{01}^s\mathbb{E}_{q_s(x_s,y_s)}[\ell(h(x_s),y_s)] - (1-\rho_{11}^s)R_s(h)|\nonumber \\
    & \leq |\rho_{01}^s(R_s(h)+M_s) - \rho_{01}^sR_s(h)|\nonumber \\
    & = \rho_{01}^sM_s.
\end{align}
Similar, we have
\begin{align}
\label{eq:second2_epsilon}
    |\tilde{R}_s^{\text{po}}(h,\tilde{f}_t,u_s)- R_s(h,\tilde{f}_t)| \leq \rho_{01}^tM_t,
\end{align}
\begin{align}
\label{eq:third2_epsilon}
    |\tilde{R}_t^{\text{po}}(h,\tilde{f}_t,u_t)-R_t(h,{f}_t)| \leq \rho_{01}^tM_t + \rho_{11}^t\rho_{01}^sM_t.
\end{align}
Since $M_s$ and $M_t$ are positive constants, it is clear that $\tilde{R}_s^{\text{po}}(h,u_s) \rightarrow R_s(h)$, $\tilde{R}_s^{\text{po}}(h,\tilde{f}_t,u_s)\rightarrow R_s(h,\tilde{f}_t)$ and $\tilde{R}_t^{\text{po}}(h,\tilde{f}_t,u_t)\rightarrow R_t(h,{f}_t)$ when $\rho^s_{01} \rightarrow 0$ and $\rho^t_{01} \rightarrow 0$. 

Specifically, $\forall \epsilon \in (0,1)$, let $\delta_t=\epsilon/M_t$ and $\delta_s=\epsilon/\max\{M_s, \rho_{11}^tM_t\}$. When $\rho^s_{01}<\delta_s$ and $\rho^t_{01}<\delta_t$, we have
\begin{align}
\label{eq:first_epsilon}
    |\tilde{R}_s^{\text{po}}(h,u_s) - R_s(h)|+|\tilde{R}_s^{\text{po}}(h,\tilde{f}_t,u_s)- R_s(h,\tilde{f}_t)| < 2\epsilon
\end{align}
\begin{align}
\label{eq:second_epsilon}
    |\tilde{R}_t^{\text{po}}(h,\tilde{f}_t,u_t)-R_t(h,{f}_t)| < 2\epsilon.
\end{align}
Hence, we prove the Eq.~(\ref{eq:3epsilon}). In following, we give a new upper bound of $R_t(h,f_t)$. Recall Theorem \ref{thm:upper_bound_target}, we replace 1) $\tilde{R}_s(h)$ with $\tilde{R}_s^{\text{po}}(h,u_s)$, 2) $\tilde{R}_s(h,\tilde{f}_t)$ with $\tilde{R}_s^{\text{po}}(h,\tilde{f}_t,u_s)$, 3) $R_t(h,\tilde{f}_t)$ with $\tilde{R}_t^{\text{po}}(h,\tilde{f}_t,u_t)$. Then, we have
\begin{align}
\label{eq:risk_bound_NEW:basic}
    R_t(h,f_t) &\leq {\tilde{R}_s^{\text{po}}(h,u_s)} +{|\tilde{R}_t^{\text{po}}(h,\tilde{f}_t,u_t) - \tilde{R}_s^{\text{po}}(h,\tilde{f}_t,u_t))|} \nonumber \\
    &~~~~+ {|R_s(h,\tilde{f}_t) - R_s(h)|+|\tilde{R}_s^{\text{po}}(h,u_s)-R_s(h)|} \nonumber \\
    &~~~~+~{|\tilde{R}_s^{\text{po}}(h,\tilde{f}_t,u_s) - R_s(h,\tilde{f}_t)|}\nonumber \\
    &~~~~+{|R_t(h,f_t) -  \tilde{R}_t^{\text{po}}(h,\tilde{f}_t,u_t)|}.
\end{align}
Let $\rho^s_{01}\leq\delta_s$ and $\rho^t_{01}\leq\delta_t$, based on Eqs.~\eqref{eq:first_epsilon} and \eqref{eq:second_epsilon}, we have
\begin{align*}
    R_t(h,f_t) &\leq \underbrace{\tilde{R}^{\text{po}}_s(h,u_s)}_{(i)~\textbf {noisy-data risk}} +~\underbrace{|\tilde{R}^{\text{po}}_t(h,\tilde{f}_t,u_t) - \tilde{R}^{\text{po}}_s(h,\tilde{f}_t,u_s)|}_{(ii)~\textbf {discrepancy~between~distributions}} \nonumber \\
    &~~~~+~ \underbrace{|R_s(h,\tilde{f}_t) - R_s(h)|}_{(iii)~\textbf {domain dissimilarity}}+~\underbrace{2\epsilon}_{(iv)~\textbf {noise~effects ~$\Delta_s$}} \nonumber \\
    &~~~~+~\underbrace{2\epsilon}_{(iv)~\textbf {noise~effects ~$\Delta_t$}}.
\end{align*}
Hence, we prove this theorem.
\end{proof}


Theorem \ref{thm: epsilon_effects} shows that if selected data have a high probability to be correct ones ($\rho^s_{01}\rightarrow0$ and $\rho^t_{01}\rightarrow0$), then $\Delta_s$ and $\Delta_t$ approach zero, meaning that noise effects are eliminated. This motivates us to find a reliable way to select correct data from noisy source data and pseudo-labeled target data and propose the butterfly to WUDA problem.
\begin{myrem}\upshape
{Note that, since Theorems~{\ref{thm:upper_bound_target}} and {\ref{thm: epsilon_effects}} hold for any hypothesis and any data distributions, the bounds in both theorems are loose and pessimistic. However, both theorems are proposed to show which factors we should take care of in the WUDA problem and both theorems point out the major difference between WUDA and UDA. From this perspective, both theorems are very important for positioning and understanding the WUDA problem.}
\end{myrem}

\vspace{-1.5em}
\section{Butterfly: Towards robust one-step approach}\label{sec:Butterfly_net}

This section presents Butterfly to solve the WUDA problem.
\vspace{-0.5em}
\subsection{What is the Principle-guided Solution?} 
\label{sec:principle_butterfly}
\label{sec:principle_rule}

Guided by Theorem \ref{thm: epsilon_effects}, a robust approach should check high-correctness data out (meaning $\rho^s_{01}\rightarrow0$ and $\rho^t_{01}\rightarrow0$). This checking process will make $(iv)$ and $(v)$, $2\epsilon+2\epsilon$, become $0$. Then, we can obtain gradients of $\tilde{R}^{\text{po}}_s(h,u_s)$,  $\tilde{R}^{\text{po}}_s(h,\tilde{f}_t,u_s)$ and $\tilde{R}^{\text{po}}_t(h,\tilde{f}_t,u_t)$ w.r.t. parameters of $h$ and use these gradients to minimize them, which minimizes $(i)$ and $(ii)$ as $(i)+(ii)\leq \tilde{R}^{\text{po}}_s(h,u_s)+\tilde{R}^{\text{po}}_s(h,\tilde{f}_t,u_s)+\tilde{R}^{\text{po}}_t(h,\tilde{f}_t,u_t)$. Note that $(iii)$ cannot be directly minimized since we cannot pinpoint clean source data. However, following \cite{KSaito_ICML17}, we can indirectly minimize $(iii)$ via minimizing $\tilde{R}^{\text{po}}_s(h,u_s) + \tilde{R}^{\text{po}}_s(h,\tilde{f}_t,u_s)$, as $(iii)\leq R_s(h,\tilde{f}_t) + R_s(h) \leq \tilde{R}^{\text{po}}_s(h,u_s)+\tilde{R}^{\text{po}}_s(h,\tilde{f}_t,u_s)+2 \epsilon$, where the last inequality follows Eq.~(\ref{eq:3epsilon}). This means that a robust approach guided by Theorem \ref{thm: epsilon_effects} can minimize all terms in the right side of inequality in Eq.~(\ref{eq:risk bound NEW}). 



\subsection{Dual-checking principle}
\label{Sec:dualCheckingP}
{\textbf{Memorization effects of deep networks.} Recently, an interesting observation for deep networks is that they can memorize easy samples first, and gradually adapt to hard samples as increasing training epochs \mbox{\cite{arpit2017closer}}. Namely, although deep networks can fit everything (e.g., mislabeled data) in the end, they \emph{learn patterns first} \mbox{\cite{arpit2017closer}}: this suggests deep networks can gradually memorize the data, moving from regular data to irregular data such as outliers. To utilize this memorization effects, previous studies have shown that we can regard small-loss data as correct ones (also known as the \emph{small-loss trick}). Then we can obtain a good classifier that is trained with the small-loss data \mbox{\cite{jiang2017mentornet}}.} 
\vspace{-0.5em}
\newline
\newline
{\textbf{Co-teaching learning paradigm.} However, if we only use small-loss trick to select correct data (like \mbox{\cite{jiang2017mentornet}}), we will get accumulated errors caused by sample-selection bias \mbox{\cite{Co-teaching}}. Therefore, researchers also consider a new deep learning paradigm called \emph{co-teaching}, where we train two deep networks simultaneously, and let them \emph{teach each other} \mbox{\cite{Co-teaching}}. Based on this novel learning paradigm, we can effectively reduce the negative effects from the accumulated errors caused by sample-selection bias.}
\vspace{-0.5em}
\newline
\newline
{\textbf{Dual-checking principle.} Motivated by Section~{\ref{sec:principle_rule}}, we propose the \emph{dual-checking principle} (DCP): we need to check high-correctness data out in the source and target domains simultaneously. According to the memorization effects of deep networks, we realize DCP based on deep networks, small-loss trick and the co-teaching learning paradigm (i.e., the Butterfly introduced below).} 

\subsection{Principle-guided Butterfly}

To realize the robust approach for addressing the WUDA problem, we propose a Butterfly framework, which trains four networks dividing into two branches (Figure \ref{fig: sketch_fig}). By using DCP, Branch-I checks which data is correct in the mixture domain; while Branch-II checks which pseudo-labeled target data is correct. To ensure these checked data highly-correct, we apply the small-loss trick based on memorization effects of deep learning \cite{arpit2017closer}. After cross-propagating these checked data \cite{bengio2014evolving}, Butterfly can obtain high-quality DIR and TSR simultaneously in an iterative manner. Theoretically, Branch-I minimizes $(i)+(ii)+(iii)+(iv)$; while Branch-II minimizes $(ii)+(v)$. This means that Butterfly can minimize all terms in the right side of inequality in Eq.~(\ref{eq:risk bound NEW}). 
\begin{algorithm*}[t]
\small

{\bfseries 1: Input} networks $F_1$, $F_2$, mini-batch $D$, learning rate $\eta$, remember rate $\alpha$;

{\bfseries 2: Obtain} ${{\bm{u}}}_1 = \arg\min_{{{\bm{u}}}^{\prime}_1:\bm{1}{{\bm{u}}}^{\prime}_1>\alpha|D|}\mathcal{L}(\theta_1,{{\bm{u}}}^{\prime}_1;F_1, D)$;  \hfill // Check high-correctness data

{\bfseries 3: Obtain} ${{\bm{u}}}_2 = \arg\min_{{{\bm{u}}}^{\prime}_2:\bm{1}{{\bm{u}}}^{\prime}_2>\alpha|D|}\mathcal{L}(\theta_2,{{\bm{u}}}^{\prime}_2;F_2, D)$; \hfill // Check high-correctness data

{\bfseries 4: Update} $\theta_1 = \theta_1 - \eta\nabla \mathcal{L}(\theta_1,{{\bm{u}}}_2;F_1, D)$; \hfill // Update $\theta_1$

{\bfseries 5: Update} $\theta_2 = \theta_2 - \eta\nabla \mathcal{L}(\theta_2,{{\bm{u}}}_1;F_2, D)$; \hfill // Update $\theta_2$

{\bfseries 6: Output $F_{1}$ and $F_{2}$}
\caption{Checking($F_1$, $F_2$, $D$, $\eta$, $\alpha$) }\label{alg: Cross_update}
\end{algorithm*}
\begin{algorithm*}[!tp]
\small
{\bfseries 1: Input} $\tilde{D}_s$, $D_t$, learning rate $\eta$, fixed $\tau$, fixed $\tau_t$, epoch $T_k$ and $T_{max}$, iteration $N_{max}$, \# of pseudo-labeled target data $n_{init}$, max of $n_{init}$ $n_{t,max}^l$;

{\bfseries 2: Initial} $F_1$, $F_2$, $F_{t1}$, $F_{t2}$, $\tilde{D}_t^l=\tilde{D}_s$, $\tilde{D}=\tilde{D}_s$,  $n_t^l = n_{init}$;

\For{$T = 1,2,\dots,T_{max}$}{

{\bfseries 3: Shuffle} training set $\tilde{D}$; \hfill // Noisy dataset

\For{$N = 1,\dots,N_{max}$}{

{\bfseries 4: Fetch} mini-batch $\check{D}$ from $\tilde{D}$;

{\bfseries 5: Update} Branch-I: $F_1,F_2$ = Checking($F_1,F_2,\check{D},\eta,R(T)$); \hfill // Check data in MD using Algorithm~\ref{alg: Cross_update}

{\bfseries 6: Fetch} mini-batch $\check{D}_t$ from $\tilde{D}_t^l$;

{\bfseries 7: Update} Branch-II: $F_{t1},F_{t2}$ = Checking($F_{t1},F_{t2},\check{D}_t,\eta,R_t(T)$); \hfill // Check data in TD using Algorithm~\ref{alg: Cross_update}

}

{\bfseries 8: Obtain} $\tilde{D}_t^l$ = Labeling$(F_1,F_2,D_t,n_t^l)$; \hfill // Label $D_t$, following \cite{KSaito_ICML17}

{\bfseries 9: Obtain} $\tilde{D}=\tilde{D}_s \cup \tilde{D}_t^l$; \hfill // Update MD

{\bfseries 10: Update} $n_t^l = \min\{T/20*n_t,n_{t,max}^l\}$;

{\bfseries 11: Update} $R(T) = 1 - \min\{\frac{T}{T_k} \tau,\tau\}$, $R_t(T) = 1 - \min\{\frac{T}{T_k} \tau_t,\tau_t\}$;

}
{\bfseries 12: Output} $F_{t1}$ and $F_{t2}$
\caption{Butterfly Framework: quadruple training for WUDA problem}\label{alg: ButterNET}
\end{algorithm*}

\subsection{Loss function in Butterfly} 
According to $\tilde{R}^{\text{po}}_s(h,u_s)$, $\tilde{R}^{\text{po}}_t(h,\tilde{f}_t,u_t)$ and $\tilde{R}^{\text{po}}_s(h,\tilde{f}_t,u_s)$ defined in Section~\ref{sec:ana_address_WUDA}, four networks trained by Butterfly share the same loss function but with different inputs.
\begin{equation}
\setlength{\abovedisplayskip}{5pt}
\setlength{\belowdisplayskip}{5pt}
\label{eq: loss_butterfly}
    \mathcal{L}(\theta,\bm{u};F,D) =\frac{1}{\sum_{i=1}^n{u}_{i}}\sum_{i=1}^nu_{i}\ell(F(x_i),\check{y}_i),
\end{equation}
where $n$ is the batch size (i.e., $n=|D|$), and $F$ represents a network (e.g., $F_1,F_2,F_{t1}$ and $F_{t2}$). $D=\{(x_i,\check{y}_i)\}_{i=1}^n$ is a mini-batch for training a network, where $\{x_i,\check{y}_i\}_{i=1}^n$ could be data in MD or TD (Figure~\ref{fig: sketch_fig}), and $\theta$ represents parameters of $F$ and $\bm{u}=[u_1, ..., u_n]^T$ is an $n$-by-$1$ vector whose elements equal $0$ or $1$. 
For two networks in Branch-I, following \cite{KSaito_ICML17}, we also add a regularizer $|\theta_{f11}^T \theta_{f21}|$ in their loss functions, where $\theta_{f11}$ and $\theta_{f21}$ are weights of the first fully-connect layer of $F_1$ and $F_2$. With this regularizer, $F_1$ and $F_2$ will learn from different features.
\vspace{-0.5em}
\newline
\newline
\textbf{Nature of the loss $\mathcal{L}$.} 
{In the loss function $\mathcal{L}$, we have $n$ samples: $\{(x_i, \check{y}_i)\}_{i=1}^n$. For the $i^{th}$ sample, we will compute its cross-entropy loss (i.e., $\ell(F(x_i),\check{y}_i)$), and we will denote this sample as ``selected'' if $u_i = 1$. Thus, the nature of $\mathcal{L}$ is actually the average value of cross-entropy loss of these ``selected'' samples. Note that, we need to set a constrain to prevent $\sum_{i=1}^nu_{i}=0$ in $\mathcal{L}$, which means that we should select at least one sample to compute $\mathcal{L}$.}

\subsection{Training procedures of Butterfly}
\label{sec:butterfly_net_train}
{This subsection will first present the checking process in Butterfly (Algorithm~{\ref{alg: Cross_update}}). Then, the full training procedure of Butterfly (Algorithm~{\ref{alg: ButterNET}}) will be introduced in detail.} 
\subsubsection{Checking process in Butterfly (Algorithm~\ref{alg: Cross_update})}
We first obtain four inputs: 1) networks $F_1$ and $F_2$, and 2) a mini-batch $D$, and 3) learning rate $\eta$ and 4) remember rate $\alpha$ (line 1).
Then, we will obtain the best $\bm{u}_1$ by solving a minimization problem (line 2). $\mathcal{L}$ represents the loss function defined in Eq.~(\ref{eq: loss_butterfly}). $\theta_1$ represents the parameters of the network $F_1$. Similarly
we will obtain the best $\bm{u}_2$ (line 3). $\theta_2$ represents the parameters of the network $F_2$. Next, $\theta_1$ and $\theta_2$ are updated using gradient descent, where the gradients are computed using a given optimizer (lines 4-5).
Finally, we substitute the updated $\theta_1$ into $F_1$ and the updated $\theta_2$ into $F_2$ and output $F_1$ and $F_2$ (line 6). Note that, lines 2-3 correspond to the small-loss trick mentioned in Section~\ref{Sec:dualCheckingP}, and lines 4-5 corresponds to the co-teaching paradigm in Section~\ref{Sec:dualCheckingP}.
\begin{myrem}\upshape
{In line $2$ or $3$ in Algorithm~\ref{alg: Cross_update}, we need to solve a minimization problem: $\min_{{{\bm{u}}}^{\prime}:\bm{1}{{\bm{u}}}^{\prime}>\alpha|D|}\mathcal{L}(\theta,{{\bm{u}}}^{\prime};F, D)$ and return the best $\bm{u}^{\prime}$ as $\bm{u}$ ($\bm{u}_1$ in line $2$ and $\bm{u}_2$ in line $3$). In this paragraph, we will show how to quickly solve this problem using a sorting algorithm. Recall the nature of the loss $\mathcal{L}$, we know $\mathcal{L}$ is the average value of cross-entropy losses of ``selected'' samples, and  $\bm{1}\bm{u}^{\prime}$ is the number of these ``selected'' samples. Therefore, this minimization problem is equivalent to ``given a fixed $F$ ($F_1$ or $F_2$) and $n$ samples in $D$, how to select at least $k$ samples such that $\mathcal{L}$ is minimized'', where $k=\lceil \alpha|D| \rceil$. To solve this problem, we first use a sorting algorithm (top\_k function in TensorFlow) to sort these $n$ samples according to their cross-entropy losses $\ell(F_1(x_i),\check{y}_i)$. Then, we select $k$ samples with the smallest cross-entropy losses. Finally, let $u_i$ of these $k$ samples be $1$ and $u_i$ of the other samples be $0$, and we can get the best $\bm{u}=[u_1, \dots, u_n]$. The average value of cross-entropy losses of these $k$ samples is the minimized value of $\mathcal{L}(\theta,\bm{u}^{\prime};F, D)$ under the constrain $\bm{1}\bm{u}^{\prime}>\alpha|D|$. It is clear that this solving process is equivalent to finding small-loss samples.}
\end{myrem}

\subsubsection{Training procedures of Butterfly (Algorithm~\ref{alg: ButterNET})}

\textbf{Update parameters of networks.}
First, we initialize training data for two branches ($\tilde{D}$ for Branch-I and $\tilde{D}^l_t$ for Branch-II), four networks ($F_1,F_2,F_{t1}$ and $F_{t2}$) and the number of pseudo labels (line $2$).
{In the first epoch ($T=1$), following \mbox{\cite{KSaito_ICML17}}, $\tilde{D}^l_t$ is the same with $\tilde{D}_s$ (i.e., we use noisy source data as pseudo-labeled target data), since we cannot annotate pseudo labels for target data when $T=1$.}
After mini-batch $\check{D}$ is fetched from $\tilde{D}$ (line $4$), $F_1$ and $F_2$ check high-correctness data out and update their parameters (lines $5$) using Algorithm \ref{alg: Cross_update}. Using similar procedures, $F_{t1}$ and $F_{t2}$ also update their parameters using Algorithm \ref{alg: Cross_update} (lines $6$-$7$).
\vspace{-0.5em}
\newline
\newline
{\textbf{Assign pseudo labels.} In each epoch, after $N_{max}$ mini-batch updating, we randomly select $n_t^l$ unlabeled target data and assign them pseudo labels using the Labeling function \mbox{\cite{KSaito_ICML17}}, $F_1$ and $F_2$ (lines $8$). Following \mbox{\cite{KSaito_ICML17}}, the Labeling function in Algorithm~{\ref{alg: ButterNET}} (line $8$) assigns pseudo labels to unlabeled target data, when predictions of $F_1$ and $F_2$ agree and at least one of them is confident about their predictions (probability above $0.9$ or $0.95$).} Using this function, we can obtain the pseudo-labeled target data $\tilde{D}^l_t$ for training Branch-II in the next epoch. Then, we merge $\tilde{D}^l_t$ and $\tilde{D}_s$ to be $\tilde{D}$ for training Branch-I in the next epoch (line $9$). 
\vspace{-0.5em}
\newline
\newline
\textbf{Update other parameters.} Finally, we update $n_t^l$, $R(T)$ and $R_t(T)$ in lines $10$-$11$.
{Note that $R(T)$ and $R_t(T)$ are actually piecewise-defined linear functions:}
\begin{equation*}
R(T)=\left\{
\begin{aligned}
1-\tau,&~~~~~~~ T\geq T_k, \\
1 - T/T_k \times \tau,&~~~~~~~T\le T_k,\\
\end{aligned}
\right.
\end{equation*}
\begin{equation*}
R_t(T)=\left\{
\begin{aligned}
1-\tau_t,&~~~~~~~ T\geq T_k, \\
1 - T/T_k \times \tau_t,&~~~~~~~T\le T_k.\\
\end{aligned}
\right.
\end{equation*}
In Algorithm~\ref{alg: ButterNET}, we use $\tau$ to represent the noise rate (i.e., the ratio of data with incorrect labels) in MD and use $\tau_t$ to represent the noise rate in TD. However, in WUDA, we cannot obtain the ground-truth $\tau$ and $\tau_t$. Thus, we regard $\tau$ and $\tau_t$ as hyper-parameters. 

\subsection{Can we realize DCP using other models?}
{Based on Theorem~{\ref{thm: epsilon_effects}}, if we check high-correctness source data and pseudo-labeled target data out, we can reduce the negative effects of noisy source data significantly. Thus, we propose the DCP to check correct data out, which is introduced in Section~{\ref{Sec:dualCheckingP}}. In Butterfly, we realize DCP using deep networks, since the memorization effects of deep networks ensures that we can check correct data out. For non-network models, if they also have memorization effects like deep networks, they can also be used into our approach. We also tried other models. Unfortunately, these models cannot fit the pattern first (like what deep networks did when fitting training data), meaning that, currently, we can only realize our approach using deep networks.}
\vspace{-1em}
\subsection{A Generalization Bound for WUDA}
In this subsection, we prove a generalization bound for WUDA problem using the loss function Eq.~\eqref{eq: loss_butterfly} and Theorem~\ref{thm: epsilon_effects}\footnote{Please note that this is a generalization bound for WUDA problem rather than Butterfly. In Butterfly, we essentially have four classifiers ($F_1,F_2,F_{t1},F_{t2}$), which is very difficult to analyze it. We will develop a generalization and estimation error bound for Butterfly in the future.}. Practitioner may safely skip it. First, we introduce the Rademacher complexity of a class of vector-valued functions \cite{BartlettM02_Rader,Mansour2009,maurer2016vector,JianLi_NeurIPS19_local_RC,JianLi_IJCAI19_local_RC,Long19_ICML_theory}, which measures the degree to which a class can fit random noise. 
Rademacher Complexity of $\mathcal{H}$ is defined as follows.


\begin{mydef}[Rademacher Complexity of $\mathcal{H}$] Given a sample $S=\{(x_i)\}_{i=1}^n$, the empirical Rademacher complexity of the set $\mathcal{H}$ is defined as follows.
\begin{align*}
\hat{\Re}_S(\mathcal{H})=\frac2n\underset{\sigma}{\mathbb{E}}\Big( \underset{h\in \mathcal{H}}{\sup}\sum_{i=1}^n\sum_{k=1}^K \sigma_{ik} h_k(x_{i})  \Big),
\end{align*}
where $h_k(\cdot)$ is the $k^{th}$ component of function $h\in\mathcal{H}$ and the $\sigma_{ik}$ are $n\times K$ matrix of independent Rademacher variables \cite{maurer2016vector}.
The Rademacher complexity of the set $\mathcal{H}$ is defined as the expectation of $\hat{\Re}_H(\mathcal{H})$ over all samples of size $n$:
\begin{align*}
{\Re}_n(\mathcal{H})=\underset{S}{\mathbb{E}}\Big(\hat{\Re}_S(\mathcal{H}) \Big| |S|=n \Big).
\end{align*}
\end{mydef}
Then, using the Rademacher complexity, we can prove an upper bound of $\tilde{R}^{\text{po}}_s(h,u_s)$ to show the relation between $\tilde{R}^{\text{po}}_s(h,u_s)$ and the loss function Eq.~\eqref{eq: loss_butterfly}. As a common practice \cite{mohri2018foundations,kiryo2017positive}, we assume that, 1) there are $C_h>0$ and $C_L>0$ such that $\sup_{h\in\mathcal{H}}\|h\|_{\infty}\leq C_h$ and $\sup_{\|t\|_{\infty}\leq C_h}\max_y \ell(t,y) \leq C_L$, and 2) $\ell(t,y)$ is Lipschitz continuous in $\|t\|_{\infty}\leq C_h$ with a Lipschitz constant $L_\ell$.
\begin{mylem}
\label{lem:R_s_po_est_err_bound}
Given a sample $S_s=\{(x_{si},y_{si},u_{si})\}_{i=1}^n$ drawn from the probability density $\tilde{p}_s^{\text{po}}(x_s,y_s,u_s)$, with the probability of at least $1-\delta$ over samples $S_s$ of size $n$ drawn from $\tilde{p}_s^{\text{po}}(x_s,y_s,u_s)$, the following inequality holds.
\begin{align}
\label{eq:Gbound_1st_risk}
\tilde{R}^{\text{po}}_s(h,\bm{u}_s) 
\leq &~\mathcal{L}(\theta,h;\bm{u}_s,D^{xy}_s) + \frac{\sqrt{2}L_\ell\hat{\Re}_{D^x_s}(\mathcal{H})}{1-\tau_s} \nonumber \\
&~+\frac{3C_L}{1-\tau_s}\sqrt{\frac{\ln\frac{\delta}{2}}{2n}},
\end{align}
where $\mathcal{L}$ is defined in Eq.~\eqref{eq: loss_butterfly}, $D^{xy}_s=\{x_{si},y_{si}\}_{i=1}^n$, $D^x_s=\{x_{si}\}_{i=1}^n$, $\bm{u}_s=[u_{s1},\dots,u_{sn}]^T$ and  $\tau_s=\rho_{u_s}=1 - \sum_{i=1}^n u_{si}/n$.
\end{mylem}
\begin{proofskt}
For simplicity, in this proof, we let $\mathcal{L}_{S_s}(\ell,  h)=\mathcal{L}(\theta,h;\bm{u}_s,D^{xy}_s)$, $\tilde{R}^{\text{po}}_s(\ell,  h) = \tilde{R}^{\text{po}}_s(h,u_s)$, and $\mathbb{E}_{S_s}[\cdot]=\mathbb{E}_{S_s\sim (\tilde{P}_s^{\text{po}})^n}[\cdot]$, where $\tilde{P}_s^{\text{po}}$ is the probability measure corresponding to the density $\tilde{p}_s^{\text{po}}$. We first prove that $\mathcal{L}_{S_s}(\ell,  h)$ is an unbiased estimator of $\tilde{R}^{\text{po}}_s(\ell,  h)$ based on the definition of $\tilde{R}^{\text{po}}_s(h,u_s)$ in Section~\ref{sec:butter_eliminate_noise}. 

Then, let $\Phi(S_s)=\sup_{\ell \in \mathbb{L}_{\mathcal{H}}}\big(\tilde{R}^{\text{po}}_s(\ell,h) - \mathcal{L}_{S_s}(\ell,h)\big)$. Changing a point of $S_s$ affects $\Phi(S_s)$ at most $C_L/(n(1-\tau_s))$. Thus, by McDiarmid’s inequality applied to $\Phi(S_s)$, for any $\delta>0$, with probability of at least $1-\delta/2$, the following inequality holds.
\begin{align*}
    \Phi(S_s)\leq \mathbb{E}_{S_s}[\Phi(S_s)] + \frac{C_L}{1-\tau_s}\sqrt{\frac{\ln(\delta/2)}{2n}}.
\end{align*}
Then, we have
\begin{align}
\label{eq:Lemma3_skt1}
    &\mathbb{E}_{S_s}[\Phi(S_s)]=\mathbb{E}_{S_s}\Big[\sup_{\ell \in \mathbb{L}_{\mathcal{H}}}\big(\tilde{R}^{\text{po}}_s(h,u_s) - \mathcal{L}_{S_s}(h)\big)\Big] \nonumber \\
    \leq~& \frac{2}{n(1-\tau_s)}\mathbb{E}_{\sigma,S_s}\Big[\sup_{\ell \in \mathbb{L}_{\mathcal{H}}}\sum_{i=1}^n{\sigma_iu_{si}\ell(h(x_{si}),y_{si})}\Big].
\end{align}
Because of the existence of $u_{si}$, Eq.~\eqref{eq:Lemma3_skt1} is not the Rademacher complexity of $\mathbb{L}_{\mathcal{H}}$ (i.e., $\Re(\mathbb{L}_{\mathcal{H}})$). However, we can prove that Eq.~\eqref{eq:Lemma3_skt1} can be bounded by $\Re(\mathbb{L}_{\mathcal{H}})/(1-\tau_s)$ using the property of $\sup$.

Since changing a point of $S_s$ affects $\Re_n(\mathbb{L}_\mathcal{H})$ at most $2C_L/n$, by McDiarmid’s inequality, for any $\delta>0$, with probability of at least $1-\delta/2$, the following inequality holds.
\begin{align*}
    \Re_n(\mathbb{L}_{\mathcal{H}})\leq\hat{\Re}_{S_s}(\mathbb{L}_{\mathcal{H}}) + 2C_L\sqrt{\frac{\ln(\delta/2)}{2n}}.
\end{align*}
Since $\ell$ is Lipschitz continuous, according to \cite{maurer2016vector}, we have
\begin{align*}
    \hat{\Re}_{S_s}(\mathbb{L}_{\mathcal{H}})\leq \sqrt{2}L_{\ell}\hat{\Re}_{D_s^x}(\mathcal{H}),
\end{align*}
which proves this lemma.
\end{proofskt}

Finally, we prove the generalization bound for WUDA problem as follows.

\begin{mythm}
\label{thm:generalization}
Given a sample $S_s=\{(x_{si},y_{si},u_{si})\}_{i=1}^{n_s}$ drawn from the probability density $\tilde{p}_s^{\text{po}}(x_s,y_s,u_s)$ and a sample $S_t=\{(x_{ti},u_{ti})\}_{i=1}^{n_t}$ drawn from the probability density $\tilde{p}_t^{\text{po}}(x_t,u_t)$, under the assumptions in Remark \ref{rem: tremondous noise assumption}, Remark \ref{rem: tremondous_noise_assumption_f} and Lemma~\ref{lem: selection discrepancy_t}, 
$\forall \epsilon \in (0,1)$, there are $\delta_s$ and $\delta_t$, if $\rho^s_{01}<\delta_s$ and $\rho^t_{01}<\delta_t$,
then, with the probability of at least $1-3\delta$, for any $h\in \mathcal{H}$, the following inequality holds.
\begin{align}
\label{eq:empicial_bound_NEW}
    R_t(h,f_t) \leq & 2\Big(\mathcal{L}(\theta,h;\bm{u}_s,D^{xy}_s) + {\mathcal{L}(\theta,h;\bm{u}_s,D^{xy}_{\tilde{s}})}\Big) \nonumber \\
    &+ \mathcal{L}(\theta,h;\bm{u}_t,D^{xy}_{\tilde{t}})+  \frac{4\sqrt{2}L_\ell\hat{\Re}_{D^x_s}(\mathcal{H})}{1-\tau_s} \nonumber \\
    &+ \frac{\sqrt{2}L_\ell\hat{\Re}_{D^x_t}(\mathcal{H})}{1-\tau_t} +\frac{12C_L}{1-\tau_s}\sqrt{\frac{\ln\frac{\delta}{2}}{2n_s}}\nonumber \\
    &+ \frac{3C_L}{1-\tau_t}\sqrt{\frac{\ln\frac{\delta}{2}}{2n_t}} + 6\epsilon,
\end{align}
where $\mathcal{L}$ is defined in Eq.~\eqref{eq: loss_butterfly}, 
$D^{xy}_s=\{x_{si},y_{si}\}_{i=1}^{n_s}$, $D^{xy}_{\tilde{s}}=\{x_{si},\tilde{f}_t(x_{si})\}_{i=1}^{n_s}$, $D^{xy}_{\tilde{t}}=\{x_{ti},\tilde{f}_t(x_{ti})\}_{i=1}^{n_t}$ $D^x_s=\{x_{si}\}_{i=1}^{n_s}$, $D^x_t=\{x_{ti}\}_{i=1}^{n_t}$, $\bm{u}_s=[u_{s1},\dots,u_{sn_s}]^T$,  $\tau_s=\rho_{u_s}=1 - \sum_{i=1}^{n_s} u_{si}/{n_s}$, $\bm{u}_t=[u_{t1},\dots,u_{tn_t}]^T$ and  $\tau_t=\rho_{u_t}=1 - \sum_{i=1}^{n_t} u_{ti}/{n_t}$.
\end{mythm}
\begin{proof}
We prove this theorem (i.e., Inequality \eqref{eq:empicial_bound_NEW}) according to Inequality \eqref{eq:risk_bound_NEW:basic}, where \eqref{eq:empicial_bound_NEW} has $7$ terms in the right side and \eqref{eq:risk_bound_NEW:basic} have $6$ terms in the right side.

1) For last $3$ terms in \eqref{eq:risk_bound_NEW:basic}, 
according to \eqref{eq:first2_epsilon}, \eqref{eq:second2_epsilon} and \eqref{eq:third2_epsilon}, we know the sum of last three terms of \eqref{eq:risk_bound_NEW:basic} is less than or equal to $4\epsilon$.

2) For first $3$ terms in \eqref{eq:risk_bound_NEW:basic}, we have shown that (in Section~\ref{sec:principle_butterfly}) the sum of the first $3$ terms in \eqref{eq:risk_bound_NEW:basic} is less than or equal to $(*)$:
\begin{align*}
    2\tilde{R}_s^{\text{po}}(h,u_s)+2\tilde{R}^{\text{po}}_s(h,\tilde{f}_t,\bm{u}_s)+\tilde{R}^{\text{po}}_t(h,\tilde{f}_t,\bm{u}_t)+2\epsilon.
\end{align*}
Then, we can prove that (similar with Lemma~\ref{lem:R_s_po_est_err_bound}), with probability of at least $1-\delta$, for any $h\in\mathcal{H}$,  
\begin{align}
\label{eq:Gbound_2nd_risk}
\tilde{R}^{\text{po}}_s(h,\tilde{f}_t,\bm{u}_s) 
\leq &~\mathcal{L}(\theta,h;\bm{u}_s,D^{xy}_{\tilde{s}}) + \frac{\sqrt{2}L_\ell\hat{\Re}_{D^x_s}(\mathcal{H})}{1-\tau_s} \nonumber \\
&~+\frac{3C_L}{1-\tau_s}\sqrt{\frac{\ln\frac{\delta}{2}}{2n_s}},
\end{align}
\begin{align}
\label{eq:Gbound_3rd_risk}
\tilde{R}^{\text{po}}_t(h,\tilde{f}_t,\bm{u}_t) 
\leq &~\mathcal{L}(\theta,h;\bm{u}_t,D^{xy}_{\tilde{t}}) + \frac{\sqrt{2}L_\ell\hat{\Re}_{D^x_t}(\mathcal{H})}{1-\tau_s} \nonumber \\
&~+\frac{3C_L}{1-\tau_t}\sqrt{\frac{\ln\frac{\delta}{2}}{2n_t}}.
\end{align}
Combining \eqref{eq:Gbound_1st_risk}, \eqref{eq:Gbound_2nd_risk}, \eqref{eq:Gbound_3rd_risk} with $(*)$, based on 1), we prove this theorem. Note that, $6\epsilon$ equals $4\epsilon$ (in 1)) $+$ $2\epsilon$ (in $(*)$).
\end{proof}

\begin{mycor}[{Generalization Bound for WUDA}]
\label{cor:generalization}
{Given a sample $S_s$ and a sample $S_t$ defined in Theorem~{\ref{thm:generalization}}, under the assumptions in Remark {\ref{rem: tremondous noise assumption}}, Remark {\ref{rem: tremondous_noise_assumption_f}} and Lemma~{\ref{lem: selection discrepancy_t}}, 
if $\rho^s_{01}<C_\rho^s/\sqrt{n_sT}$  and $\rho^t_{01}<C_\rho^t/\sqrt{n_tT}$,
then, with the probability of at least $1-3\delta$, for any $h\in \mathcal{H}$, the following inequality holds.}
\begin{align}
\label{eq:empicial_bound_NEW_cor}
    &R_t(h,f_t) \nonumber\\ 
    \leq~& 2\Big(\mathcal{L}(\theta,h;\bm{u}_s,D^{xy}_s) + {\mathcal{L}(\theta,h;\bm{u}_s,D^{xy}_{\tilde{s}})}\Big) \nonumber \\
    &+ \mathcal{L}(\theta,h;\bm{u}_t,D^{xy}_{\tilde{t}})+  \frac{4\sqrt{2}L_\ell\hat{\Re}_{D^x_s}(\mathcal{H})}{1-\tau_s} \nonumber \\
    &+ \frac{\sqrt{2}L_\ell\hat{\Re}_{D^x_t}(\mathcal{H})}{1-\tau_t} +\frac{12C_L}{1-\tau_s}\sqrt{\frac{\ln\frac{\delta}{2}}{2n_s}}+ \frac{3C_L}{1-\tau_t}\sqrt{\frac{\ln\frac{\delta}{2}}{2n_t}}\nonumber \\
    & + \frac{C_\rho^s(2M_s+M_t)}{\sqrt{n_sT}} + \frac{3C_\rho^tM_t}{\sqrt{n_tT}},
\end{align}
{where $\mathcal{L}$, $D^{xy}_s$, $D^{xy}_{\tilde{s}}$, $D^{xy}_{\tilde{t}}$, $D^x_t$, $\bm{u}_s$, $\tau_s$, $\bm{u}_t$, $\tau_t$ are defined in Theorem~{\ref{thm:generalization}},
$T$ is the number of training epochs, and $C_\rho^s$ and $C_\rho^t$ are two finite constants.}
\end{mycor}
\begin{myrem}\upshape
{In Corollary~{\ref{thm:generalization}}, we assume that $\rho_{01}^s$ and $\rho_{01}^t$ will go to zero with the convergence speed of $O(1/\sqrt{n_sT})$ and $O(1/\sqrt{n_tT})$, respectively. In Section~{\ref{sec:verify_rho}}, we verify this assumption through our experiments.} 
\end{myrem}

Corollary~\ref{thm:generalization} shows the empirical upper bound of the target risk (i.e., $R_t(h,f_t)$). Based on this bound, we can obtain the estimation error bound of $R_t(h,f_t)$ as follows. First, let 
\begin{align}
    \hat{R}_t^{\mathcal{L}}(h,S_s,S_t)=&2\Big(\mathcal{L}(\theta,h;\bm{u}_s,D^{xy}_s) + {\mathcal{L}(\theta,h;\bm{u}_s,D^{xy}_{\tilde{s}})}\Big) \nonumber \\
    &+ \mathcal{L}(\theta,h;\bm{u}_t,D^{xy}_{\tilde{t}}),
\end{align}
where $D^{xy}_s$, $D^{xy}_{\tilde{s}}$ and $D^{xy}_{\tilde{t}}$ are defined in Theorem~\ref{thm:generalization}, and  $\widetilde{h} = \arg\min_{h\in\mathcal{H}}\hat{R}_t^{\mathcal{L}}(h,S_s,S_t)$ means the empirical minimizer of $\hat{R}_t^{\mathcal{L}}(h,S_s,S_t)$, and ${h^*} = \arg\min_{h\in\mathcal{H}}R_t(h,f_t)$ means the true risk minimizer of $R_t(h,f_t)$, and $\mathcal{H}' = \{h|\hat{R}_t^{\mathcal{L}}(h,S_s,S_t) \leq \epsilon'\}$. Then, we have
\begin{align}
    &R_t(\widetilde{h},f_t) - R_t(h^*,f_t) \nonumber \\
    =~& R_t(\widetilde{h},f_t) - \hat{R}_t^{\mathcal{L}}(\widetilde{h},S_s,S_t) + \hat{R}_t^{\mathcal{L}}(\widetilde{h},S_s,S_t) -R_t(h^*,f_t) \nonumber \\
    ~~~~&+\hat{R}_t^{\mathcal{L}}(h^*,S_s,S_t)-\hat{R}_t^{\mathcal{L}}(h^*,S_s,S_t) \nonumber \\
    =~& R_t(\widetilde{h},f_t) - \hat{R}_t^{\mathcal{L}}(\widetilde{h},S_s,S_t) + \hat{R}_t^{\mathcal{L}}(h^*,S_s,S_t) -R_t(h^*,f_t)\nonumber \\
    ~~~~&+\hat{R}_t^{\mathcal{L}}(\widetilde{h},S_s,S_t) -\hat{R}_t^{\mathcal{L}}(h^*,S_s,S_t) \nonumber \\
    \leq~& \sup_{h\in\mathcal{H'}}(R_t(h,f_t) - \hat{R}_t^{\mathcal{L}}(h,S_s,S_t)) +  \epsilon' + 0,
\end{align}
where $\hat{R}_t^{\mathcal{L}}(\widetilde{h},S_s,S_t) \leq \hat{R}_t^{\mathcal{L}}(h^*,S_s,S_t)$ due to the definition of $\widetilde{h}$. If all conditions in Theorem~\ref{thm:generalization} are satisfied, with the probability of at least $1-3\delta$, for any $h\in\mathcal{H}$, we have
\begin{align}
\label{eq:est_err_bound}
    &R_t(\widetilde{h},f_t) - R_t(h^*,f_t)\nonumber \\
    \leq~& \frac{4\sqrt{2}L_\ell\hat{\Re}_{D^x_s}(\mathcal{H})}{1-\tau_s} + \frac{\sqrt{2}L_\ell\hat{\Re}_{D^x_t}(\mathcal{H})}{1-\tau_t} +\frac{12C_L}{1-\tau_s}\sqrt{\frac{\ln\frac{\delta}{2}}{2n_s}}\nonumber \\
    &+ \frac{3C_L}{1-\tau_t}\sqrt{\frac{\ln\frac{\delta}{2}}{2n_t}} + \frac{C_{\rho}^s(2M_s+M_t)}{\sqrt{n_sT}} + \frac{3C_{\rho}^tM_t}{\sqrt{n_tT}} +  \epsilon'.
\end{align}
Eq.~\eqref{eq:est_err_bound} ensures that learning with $\hat{R}_t^{\mathcal{L}}(\widetilde{h},S_s,S_t)$ is consistent: as $n_s,n_t\rightarrow \infty$ and $\epsilon' \rightarrow 0$, $R_t(\widetilde{h},f_t) \rightarrow R_t(h^*,f_t)$. For linear-in-parameter model with a bounded norm, $\hat{\Re}_{D^x_s}(\mathcal{H})=\mathcal{O}(1/\sqrt{n_s})$ and $\hat{\Re}_{D^x_t}(\mathcal{H})=\mathcal{O}(1/\sqrt{n_t})$ and thus $R_t(\widetilde{h},f_t) \rightarrow R_t(h^*,f_t)$ in $\mathcal{O}(1/\sqrt{n_s}+1/\sqrt{n_t})$.

\section{Comparison to related works}
In this section, we compare Butterfly with related works and show why related works cannot handle WUDA problem.
\vspace{-1em}
\newline
\newline
\textbf{Relations to co-teaching.} As Butterfly is related to co-teaching, we discuss their major differences here. Although co-teaching applies the small-loss trick and the cross-update technique to train deep networks against noisy data, it can only deal with one-domain problem instead cross-domain problem. Besides, we argue that Butterfly is not a simple mixtrue of co-teaching and ATDA for two reasons.
\vspace{-1em}
\newline
\newline
First, network structure of Butterfly is different with that of ATDA and co-teaching: Butterfly maintains four networks; while ATDA maintains three and co-teaching maintains two. We cannot simply combine ADTA and co-teaching to derive Butterfly. Second, we have justified that the sequential mixture of co-teaching and ATDA (i.e., two-step method) cannot eliminate noise effects caused by noisy source data (see Section~\ref{sec:two-step}). Specifically, two-step methods only take care of part of noise effects but Butterfly takes care of the whole noise effects. Thus, Butterfly is the first method to eliminate noise effects rather than alleviate it.
\vspace{-1em}
\newline
\textbf{Relations to TCL.} Recently, \emph{transferable curriculum learning} (TCL) is a robust UDA method to handle noise \cite{TCL_long}. TCL uses small-loss trick to train DANN \cite{DANN_JMLR}. However, TCL can only minimize $(i)+(ii)+(iv)$, while Butterfly can minimize all terms in the right side of Eq.~(\ref{eq:risk bound NEW}).

\begin{figure*}[tp]
	\begin{center}
		\subfigure[\emph{MNIST}]
		{\includegraphics[width=0.25\textwidth]{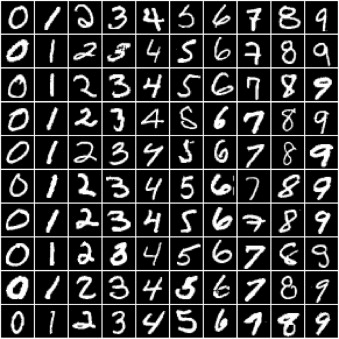}}
		\subfigure[\emph{SYND}]
		{\includegraphics[width=0.25\textwidth]{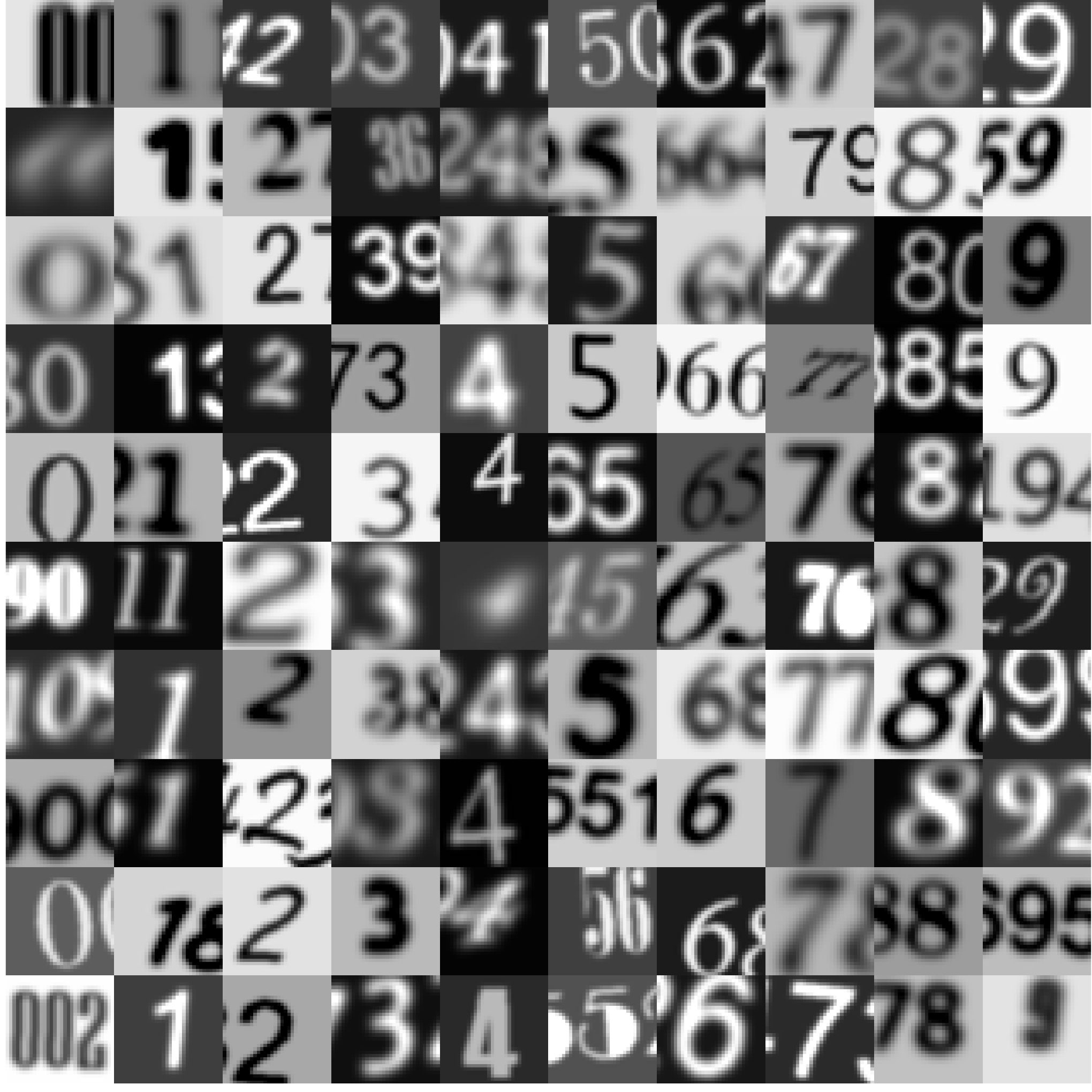}}
		\caption{Visualization of \emph{MNIST} and \emph{SYND}.}
		\label{fig:V_dig}
	\end{center}
    \vspace{-1em}
\end{figure*}
\begin{figure*}[h]
	\begin{center}
		\subfigure[\emph{Bing} provided by \cite{Bing_data}]
		{\includegraphics[width=0.24\textwidth]{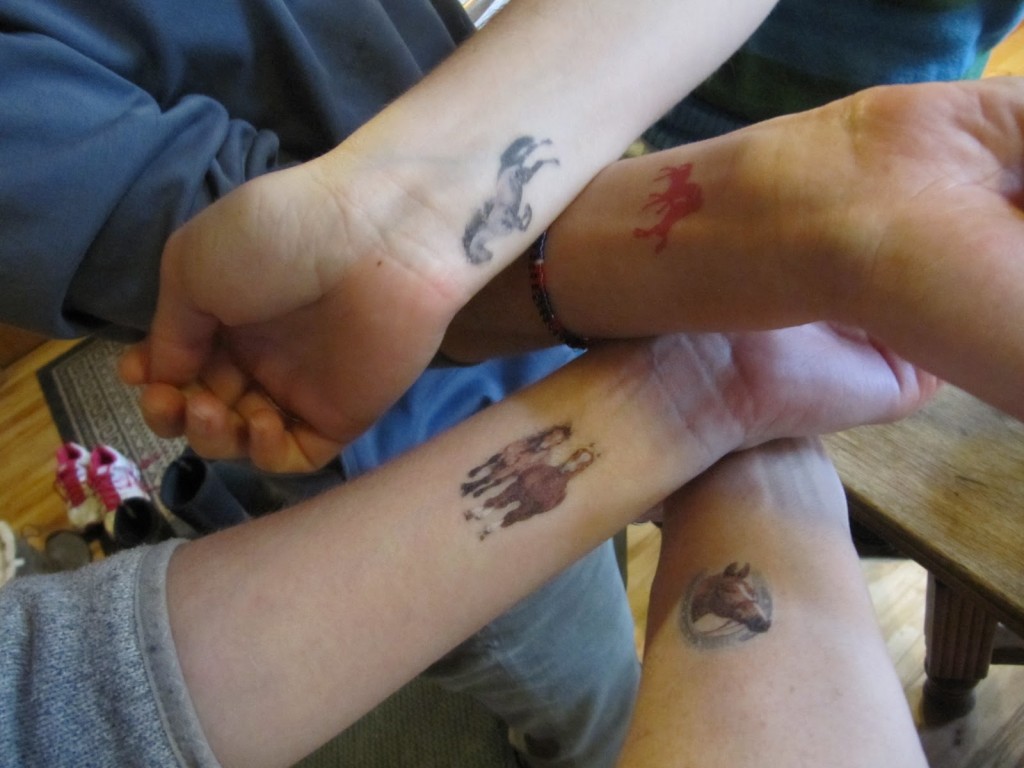}}
		\subfigure[\emph{Caltech256} provided by \cite{Caltech256}]
		{\includegraphics[width=0.24\textwidth]{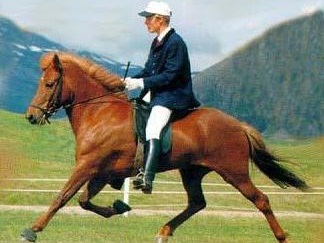}}
		\subfigure[\emph{ImageNet} provided by \cite{ImageNet}]
		{\includegraphics[width=0.24\textwidth]{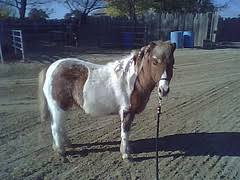}}
		\subfigure[\emph{SUN} provided by \cite{SUN_data}]
		{\includegraphics[width=0.24\textwidth]{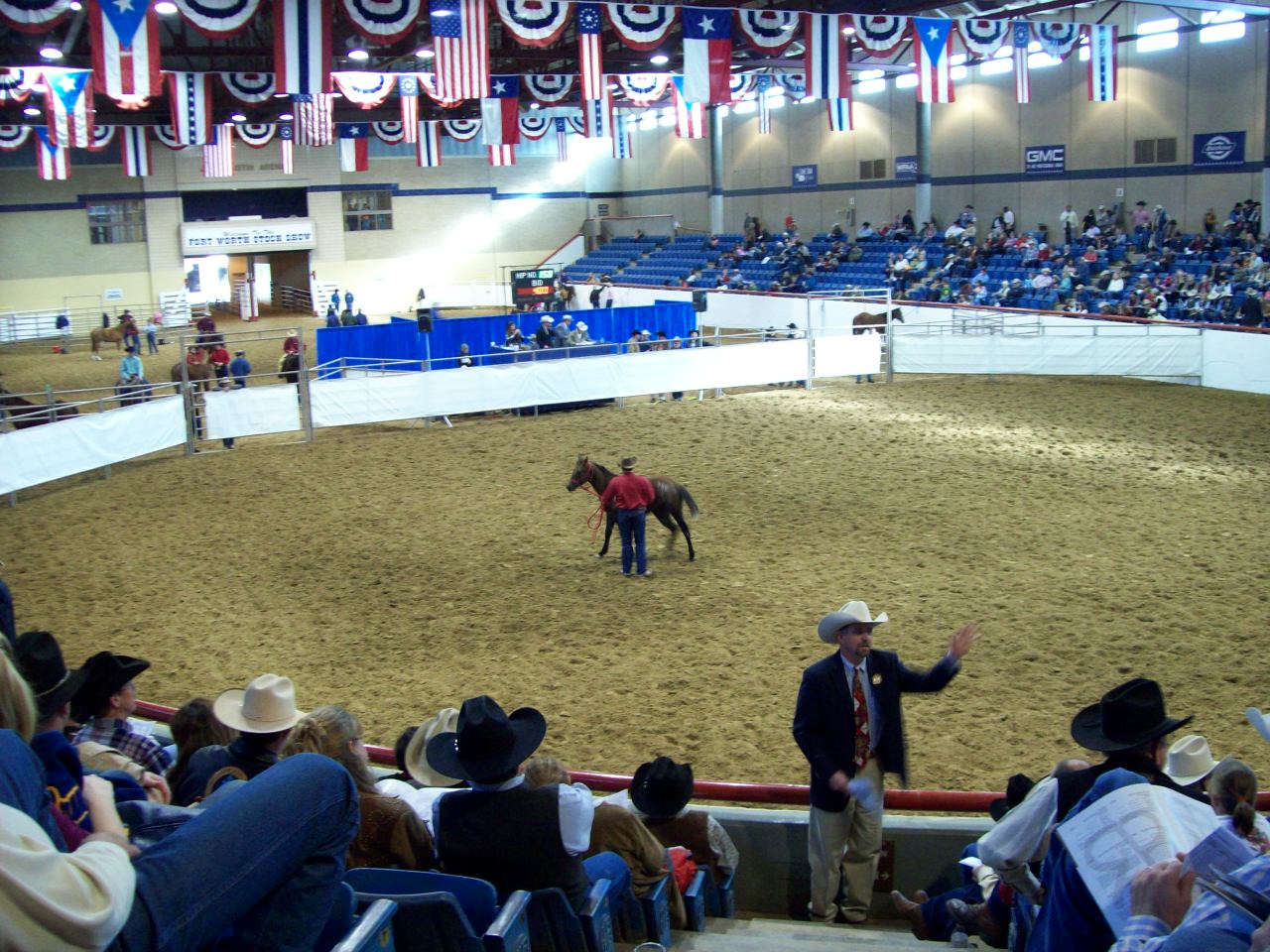} }
		\caption{Visualization of \emph{Bing}, \emph{Caltech256}, \emph{ImageNet} and \emph{SUN} (taking ``horse'' as the common class).}
		\label{fig:BCIS}
	\end{center}
	\vspace{-2em}
\end{figure*}

\section{Experiments}
\label{sec:exp}
We conduct experiments on $32$ simulated WUDA tasks and $3$ real-world WUDA tasks to verify the efficacy of Butterfly.
\subsection{Simulated WUDA tasks} We verify the effectiveness of our approach on three benchmark datasets (vision and text), including \textit{MNIST}, \textit{SYN-DIGITS (SYND)}\footnote{\emph{Digit} datasets (\emph{MNIST} and \emph{SYN Digit}) can be downloaded from official code of ATDA. The link is \url{https://github.com/ksaito-ut/atda}.} and \textit{human-sentiment} analysis (i.e., \textit{Amazon products reviews} on \emph{book}, \emph{dvd}, \emph{electronics} and \emph{kitchen}) \footnote{\emph{Sentiment} datasets (\emph{Amazon products reviews}) can be downloaded from the official code of marginalized Stacked Denoising Autoencoder. The link is \url{https://www.cse.wustl.edu/~mchen/code/mSDA.tar}.}. They are used to construct $14$ basic tasks: \textit{MNIST}$\rightarrow$\textit{SYND} (\emph{M}$\rightarrow$\emph{S}), \textit{SYND}$\rightarrow$\textit{MNIST} (\emph{S}$\rightarrow$\emph{M}), \emph{book}$\rightarrow$\emph{dvd} (\emph{B}$\rightarrow$\emph{D}), \emph{book}$\rightarrow$\emph{electronics} (\emph{B}$\rightarrow$\emph{E}), $\ldots$~, and \emph{kitchen} $\rightarrow$ \emph{electronics} (\emph{K}$\rightarrow$\emph{E}). These tasks are often used for evaluation of UDA methods \cite{DANN_JMLR,KSaito_ICML17,Saito:MCD}. Figure~\ref{fig:V_dig} shows datasets \emph{MNIST} and \emph{SYND}.  

Since all source datasets are clean, we corrupt source data using symmetry flipping \cite{Patrini_CVPR2017} and pair flipping \cite{Co-teaching} with noise rate $\rho$ chosen from $\{0.2,0.45\}$. {Note that, there are other ways to generate the noisy source data, such as asymmetry flipping. However, since the asymmetry flipping can be regarded as the combination of symmetry flipping and pair flipping, we only use symmetry flipping and pair flipping to generate \emph{simulated} WUDA tasks. In \emph{real-world} WUDA tasks, we have more complex noisy source data, where the noisy type in the source domain is unknown.} 

Therefore, for each basic task, we have four kinds of noisy source data: \emph{Pair-}$45\%$ (P$45$), \emph{Pair-}$20\%$ (P$20$), \emph{Symmetry-}$45\%$ (S$45$), \emph{Symmetry-}$20\%$ (S$20$). Following \cite{Co-teaching,jiang2017mentornet}, we can corrupt clean-label datasets manually using the noise transition matrix $Q_S$ and $Q_P$.
Namely, we evaluate the performance of each method using $32$ simulated WUDA tasks: $8$ digit tasks and $24$ human-sentiment tasks. Since the human-sentiment task is a binary classification problem, pair flipping is equal to symmetry flipping, meaning that we have $24$ human-sentiment tasks. 



\vspace{-0.5em}
\subsection{Real-world WUDA tasks}
We also verify the efficacy of our approach on ``cross-dataset benchmark'' including \emph{Bing}, \emph{Caltech256}, \emph{Imagenet} and \emph{SUN} \cite{test_bed_dataset} \footnote{\emph{Real-world} datasets (\emph{BCIS}) can be downloaded from the website of the project ``A Testbed for Cross-Dataset Analysis'': \url{https://sites.google.com/site/crossdataset/home/files} ("setup DENSE decaf7", 1.3GB, decaf7 features).}. In this benchmark, \emph{Bing}, \emph{Caltech256}, \emph{Imagenet} and \emph{SUN} contain common $40$ classes. {Since \emph{Bing} dataset was formed by collecting images retrieved by Bing image search, it contains rich noisy data, with presence of multiple objects in the same image and caricaturization \mbox{\cite{test_bed_dataset}}. We use \emph{Bing} as noisy source data}, and \emph{Caltech256}, \emph{Imagenet} and \emph{SUN} as unlabeled target data, which can form three real-world WUDA tasks. Figure~\ref{fig:BCIS} shows datasets \textit{Bing}, \textit{Caltech256}, \textit{Imagenet} and \textit{SUN} (taking ``horse'' as the common class).
\vspace{-0.5em}
\subsection{Baselines} We realize Butterfly using four networks (B-Net) and compare B-Net with following baselines: 1) ATDA: representative pseudo-labeling-based UDA method \cite{KSaito_ICML17}; 2) DAN: representative IPM-based UDA method \cite{Long_DAN}; 3) DANN: representative adversarial-training-based UDA method \cite{DANN_JMLR}; 4) {\emph{Manifold embedded distribution alignment} (MEDA): a representative non-deep UDA method \mbox{\cite{MADA_ACMMM18}}}; 5) TCL: an existing robust UDA method; 6) co-teaching+ATDA (Co+ATDA): a two-step method (see Section~\ref{sec:two-step}); and {7) co-teaching+TCL (Co+TCL).} Since MEDA cannot extract features from images, we only compare with MEDA on human-sentiment tasks, where features are already given.

\begin{figure*}[tp]
	\begin{center}
		{\includegraphics[width=0.75\textwidth]{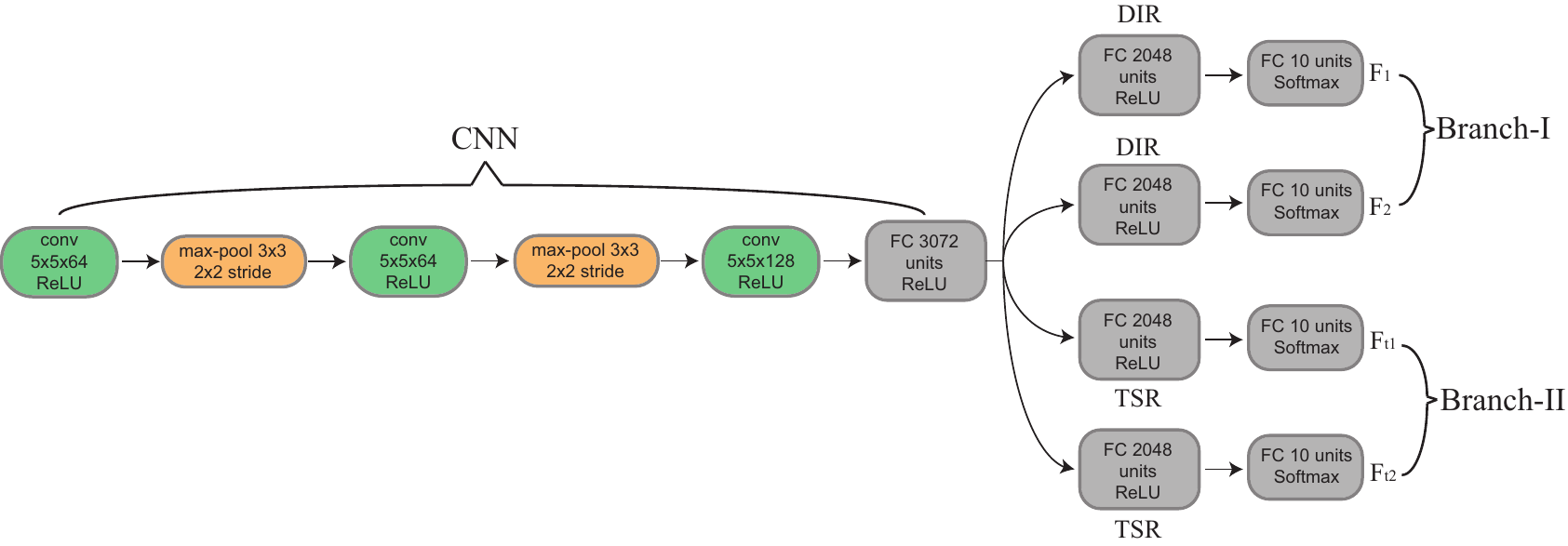}}
		\caption{The architecture of B-Net for digit WUDA tasks \emph{SYND} $\leftrightarrow$ \emph{MNIST}. We added BN layer in the last convolution layer in CNN and FC layers in $F_1$ and $F_2$. We also used dropout in the last convolution layer in CNN and FC layers in $F_1$, $F_2$, $F_{t1}$ and $F_{t2}$ (dropout probability is set to $0.5$).}
		\label{fig:dig_cnn}
	\end{center}
	\vspace{-2em}
\end{figure*}

\begin{figure*}[tp]
	\begin{center}
		\subfigure[Human-sentiment]
		{\includegraphics[width=0.4\textwidth]{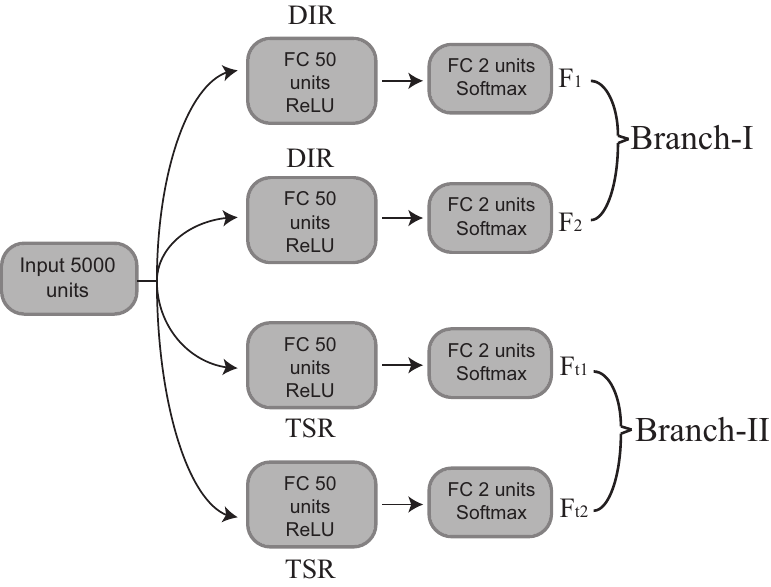}}
		\subfigure[Real-world]
		{\includegraphics[width=0.5\textwidth]{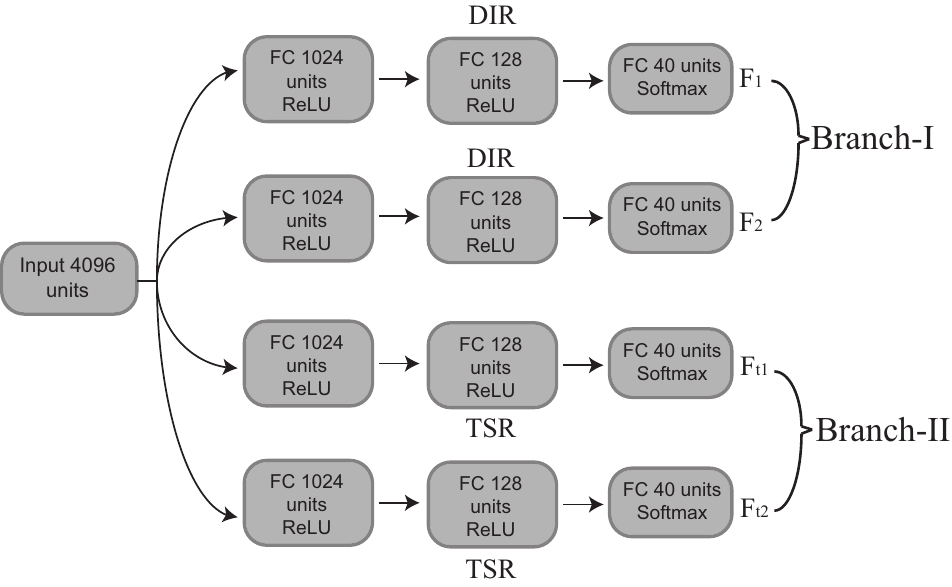}}
		\caption{The architecture of B-Net for (a) human-sentiment WUDA tasks and (b) real-world WUDA tasks. We added BN layer in the first FC layers in $F_1$ and $F_2$. We also used dropout in the first FC layers in $F_1$, $F_2$, $F_{t1}$ and $F_{t2}$ (dropout probability is set to $0.5$).}
		\label{fig: ama_cnn}
	\end{center}
	\vspace{-1em}
\end{figure*}
\vspace{-1em}

\subsection{Network structure and optimizer}
We implement all methods on Python 3.6 with a NIVIDIA P100 GPU. We use MomentumSGD for optimization in digit and real-world tasks, and set the momentum as $0.9$. We use Adagrad for optimization in human-sentiment tasks because of sparsity of review data \cite{KSaito_ICML17}. $F_{1}$, $F_{2}$, $F_{t1}$ and $F_{t2}$ are $6$-layer CNN ($3$ convolutional and $3$ fully-connected layers) for digit tasks; and are $3$-layer neural networks ($3$ fully-connected layers) for human-sentiment tasks; and are $4$-layer neural networks ($4$ fully-connected layers) for real-world tasks. The ReLU active function is used as activation function of these networks. Besides, dropout and batch normalization are also used. The network topology is shown in Figures~\ref{fig:dig_cnn} and \ref{fig: ama_cnn}. As deep networks are highly nonconvex, even with the same network and optimization method, different initializations can lead to different local optimal. Thus, following \cite{DeCoupling}, we take four networks with the same architecture but different initialization as four classifiers.

\subsection{Experimental setup}

{Since this paper deals with the challenging situation where no labeled data are available in the target domain, we follow the common protocol to set hyperparameters that the similar tasks have the same hyperparameters \mbox{\cite{Long_JAN}}. For example, we set the same hyperparameters for all WUDA tasks regarding digit datasets (there are $8$ WUDA tasks regarding digit datasets). The selected hyperparameters are robust to many tasks rather than a specific task. Details can be found below.}

\begin{figure*}[tp]
	\begin{center}
		\subfigure[S$20$]
		{\includegraphics[width=0.24\textwidth]{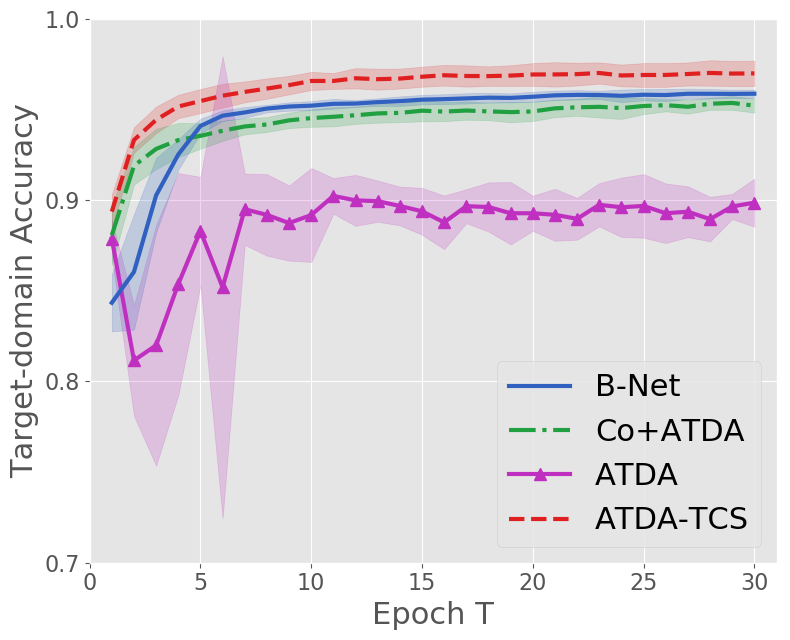}}
		\vspace{-1em}
		\subfigure[S$45$]
		{\includegraphics[width=0.24\textwidth]{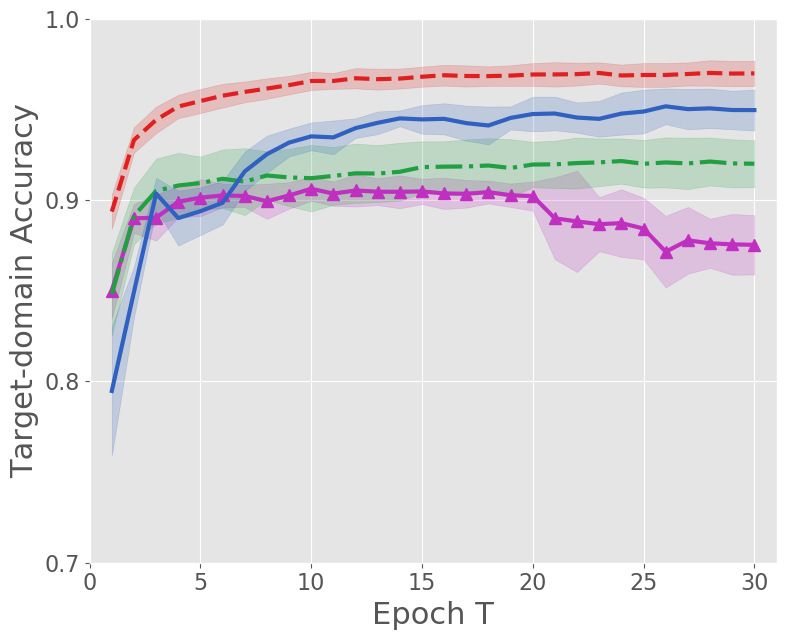}}
		\subfigure[P$20$]
		{\includegraphics[width=0.24\textwidth]{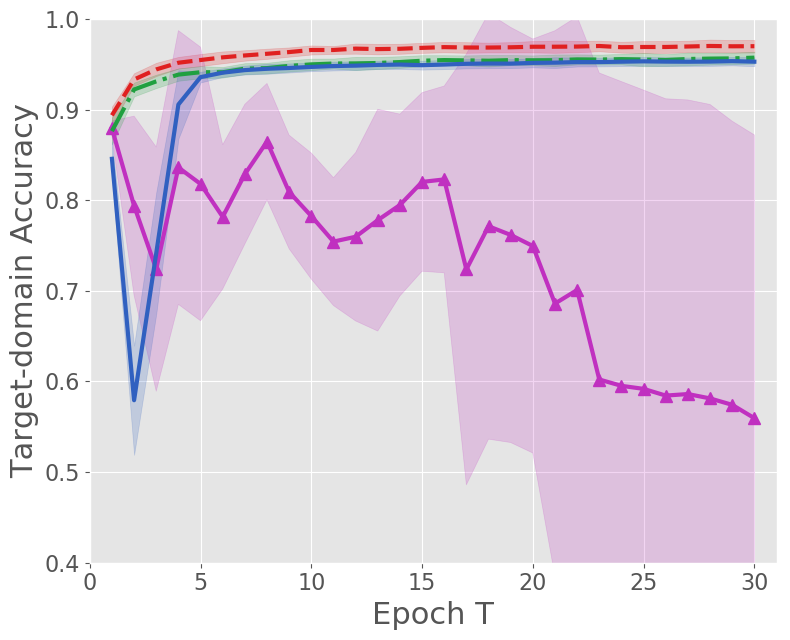}}
		\subfigure[P$45$]
		{\includegraphics[width=0.24\textwidth]{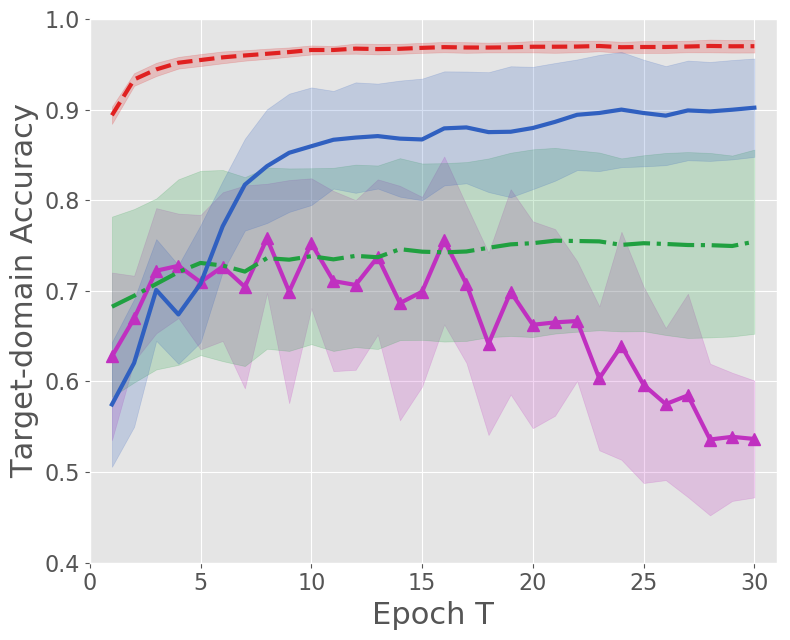}}
		\caption{{\color{mydarkblue}Target-domain accuracy vs. number of epochs on four \textit{SYND$\rightarrow$\textit{MNIST}} WUDA tasks.}}
		\label{fig: result-S2M-part}
	\end{center}
	\vspace{-2em}
\end{figure*}
\begin{figure*}[tp]
	\begin{center}
		\subfigure[S$20$]
		{\includegraphics[width=0.24\textwidth]{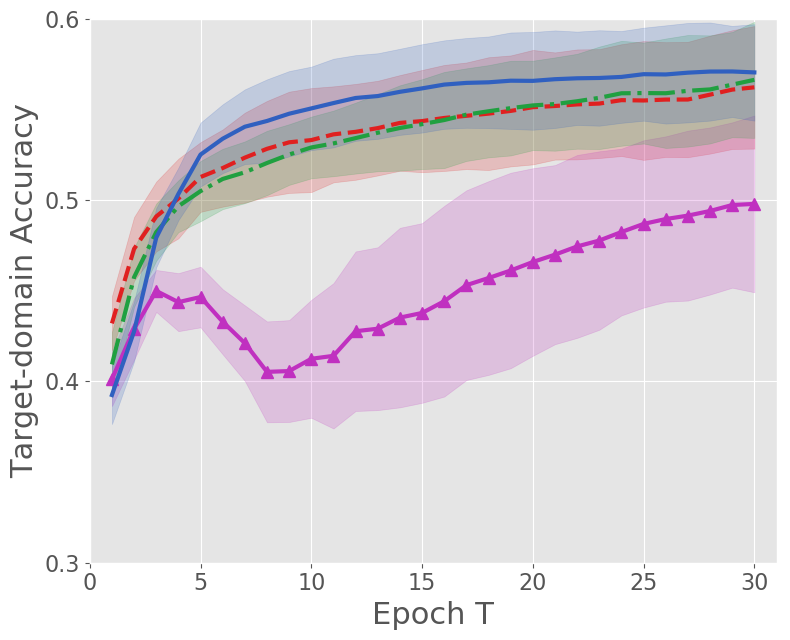}}
		\vspace{-1em}
		\subfigure[S$45$]{
		\includegraphics[width=0.24\textwidth]{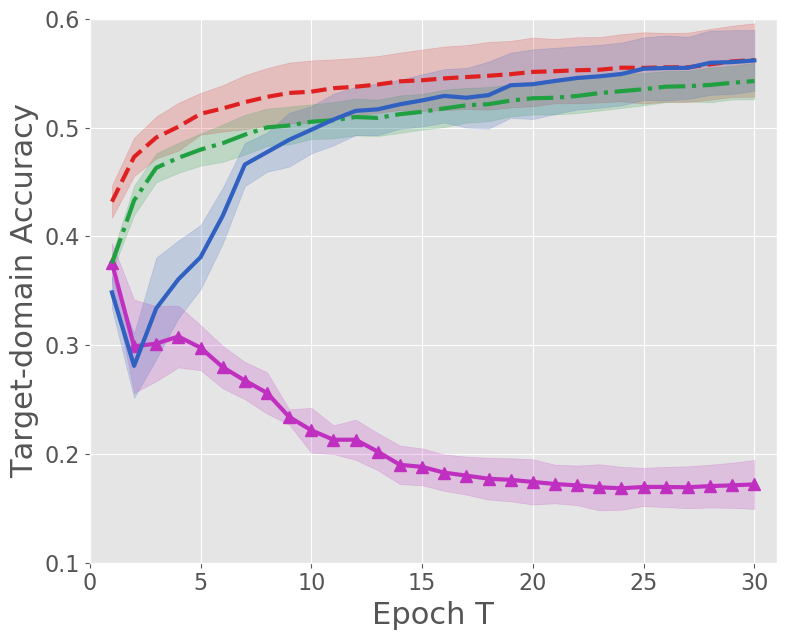}}
		\subfigure[P$20$]
		{\includegraphics[width=0.24\textwidth]{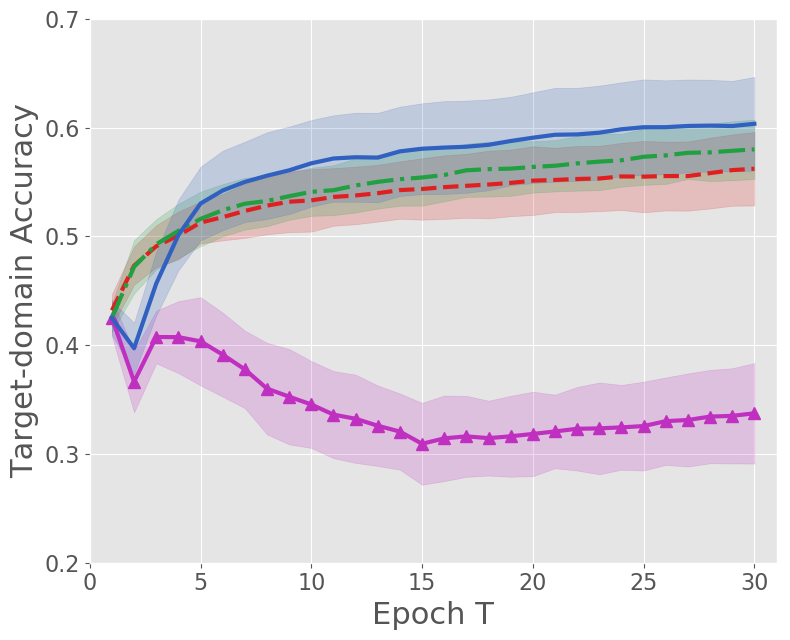}}
		\subfigure[P$45$]{
		\includegraphics[width=0.24\textwidth]{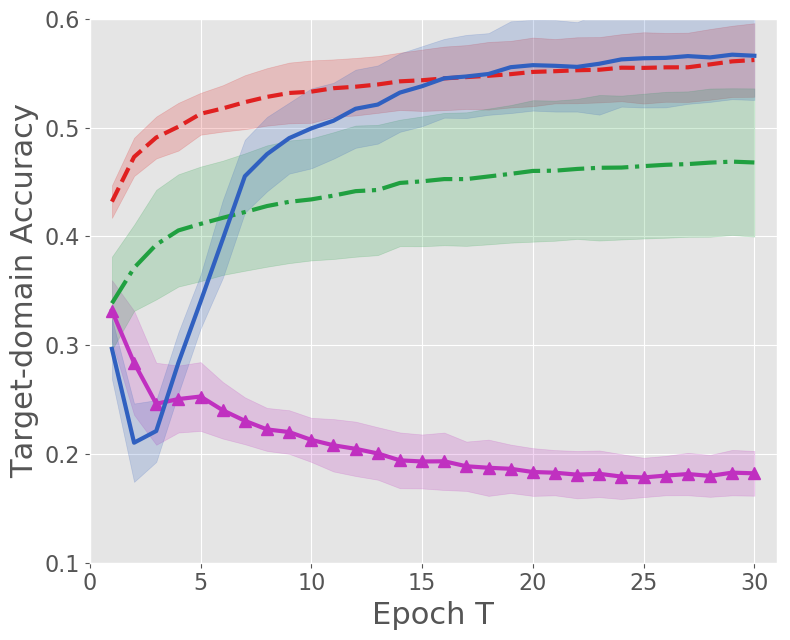}}
		\caption{{\color{mydarkblue}Target-domain accuracy vs. number of epochs on four \textit{MNIST$\rightarrow$\textit{SYND}} WUDA tasks.}}
		\label{fig: result-M2S-part}
	\end{center}
	\vspace{-1em}
\end{figure*}
\begin{table*}[!tp]
  \centering
  \footnotesize
  \caption{Target-domain accuracy on $8$ digit WUDA tasks (\textit{SYND}$\leftrightarrow$\textit{MNIST}). Bold value represents the highest accuracy in each row.}
  \vspace{-1em}
    \begin{tabular}{ccllllllll}
    \toprule
    \multicolumn{1}{l}{Tasks} & \multicolumn{1}{l}{Type} & DAN & DANN & \multicolumn{1}{p{3.335em}}{ATDA} & \multicolumn{1}{p{3.28em}}{TCL} & \multicolumn{1}{p{3.92em}}{Co+TCL} & \multicolumn{1}{p{3.92em}}{Co+ATDA} & \multicolumn{1}{p{3.555em}}{B-Net} \\
    \midrule
    \multicolumn{1}{l}{\multirow{4}[2]{*}{\textit{S}$\rightarrow$\textit{M}}} & \multicolumn{1}{l}{P$20$} & 90.17\% &79.06\% & 55.95\% & 80.81\% & {\color{mydarkblue}88.56\%} &\textbf{95.37\%} & 95.29\% \\
       & \multicolumn{1}{l}{P$45$} & 67.00\% & 55.34\% & 53.66\% & 55.97\% & {\color{mydarkblue}73.27\%} & 75.43\% &  \textbf{90.21\%} \\
       & \multicolumn{1}{l}{S$20$} & 90.74\% & 75.19\% & 89.87\% & 80.23\% & {\color{mydarkblue}85.88\%} & 95.22\% &  \textbf{95.88\%} \\
       & \multicolumn{1}{l}{S$45$} & 89.31\% & 65.87\% & 87.53\% & 68.54\% & {\color{mydarkblue}75.69\%} & 92.03\% &  \textbf{94.97\%} \\
    \midrule
    \multicolumn{1}{l}{\multirow{4}[2]{*}{\textit{M}$\rightarrow$\textit{S}}} & \multicolumn{1}{l}{P$20$} & 40.82\% & 58.78\% & 33.74\% & 58.88\% & {\color{mydarkblue}59.08\%} & 58.02\% &  \textbf{60.36\%} \\
       & \multicolumn{1}{l}{P$45$} & 28.41\% & 43.70\% & 19.50\% & 45.31\% & {\color{mydarkblue}47.15\%} & 46.80\% &  \textbf{56.62\%} \\
       & \multicolumn{1}{l}{S$20$} & 30.62\% & 53.52\% & 49.80\% & 56.74\% & {\color{mydarkblue}56.91\%} & 56.64\% &  \textbf{57.05\%} \\
       & \multicolumn{1}{l}{S$45$} & 28.21\% & 43.76\% & 17.20\% & 49.91\% & {\color{mydarkblue}51.22\%} & 54.29\% &  \textbf{56.18\%} \\
    \midrule
    \multicolumn{2}{c}{Average} & 58.16\% & 58.01\% & 50.91\% & 62.05\% & {\color{mydarkblue}67.22\%} & 71.73\% &  \textbf{75.82\%} \\
    \bottomrule
    \end{tabular}%
  \label{tab: digit_results}%
  \vspace{-1.2em}
\end{table*}%
For all $35$ WUDA tasks, $T_k$ is set to $5$, and $T_{max}$ is set to $30$, and $\ell(\cdot,\cdot)$ is the cross-entropy loss function. Learning rate is set to $0.01$ for simulated tasks and $0.05$ for real-world WUDA tasks, $\tau_t$ is set to $0.05$ for simulated tasks and $0.02$ for real-world WUDA tasks. Confidence level of labeling function in line $8$ of Algorithm \ref{alg: ButterNET} is set to $0.95$ for $8$ digit tasks, and $0.9$ for $24$ human-sentiment tasks and $0.8$ for real-world WUDA tasks. $\tau$ is set to $0.4$ for digit tasks, $0.1$ for human-sentiment tasks, $0.2$ for real-world WUDA tasks. $n_{t,max}^l$ is set to $15,000$ for digit tasks, $500$ for human-sentiment tasks and $4000$ for real-world WUDA tasks. $N_{max}$ is set to $1000$ for digit tasks, and $200$ for human-sentiment and real-world tasks. Batch size is set to $128$ for digit, real-world WUDA tasks, and $24$ for human-sentiment tasks. Penalty parameter is set to $0.01$ for digit, real-world WUDA tasks, and $0.001$ for human-sentiment tasks.

To fairly compare all methods, they have the same network structure. Namely, ATDA, DAN, DANN, TCL and B-Net adopt the same network structure for each dataset. Note that DANN and TCL use the same structure for their discriminate networks. All experiments are repeated $10$ times and we report the average accuracy values and \emph{standard deviation} (STD) of accuracy values of $10$ experiments.

\subsection{Results on simulated WUDA tasks} 
This subsection presents accuracy on unlabled target data (i.e., target-domain accuracy) in $32$ simulated WUDA tasks.
\subsubsection{Results on digits WUDA tasks}
Table~\ref{tab: digit_results} reports the target-domain accuracy in $8$ digit tasks. As can be seen, average target-domain accuracy of B-Net is higher than those of all baselines. On S$20$ case (the easiest case), most methods work well. ATDA has a satisfactory performance although it does not consider the noise effects explicitly. Then, when facing harder cases (i.e., P$20$ and P$45$), ATDA fails to transfer useful knowledge from noisy source data to unlabeled target data. When facing the hardest cases (i.e., \textit{M}$\rightarrow$\textit{S} with P$45$ and S$45$), DANN has higher accuracy than DAN and ATDA have. However, when facing the easiest cases (i.e., \textit{S}$\rightarrow$\textit{M} with P$20$ and S$20$), the performance of DANN is worse than that of DAN and ATDA.

{Although two-step method Co+ATDA (or Co+TCL) outperforms ATDA (or TCL) in all $8$ tasks, it cannot beat one-step method: B-Net in terms of average target-domain accuracy.} This result is an evidence for the claim in Section \ref{sec:two-step}. In the task \textit{S}$\rightarrow$\textit{M} with P$20$, Co+ATDA outperforms all methods (slightly higher than B-Net), since pseudo-labeled source data are almost correct.

Figures~\ref{fig: result-S2M-part} and \ref{fig: result-M2S-part} show the target-domain accuracy vs. number of epochs among ATDA, Co+ATDA and B-Net. Besides, we show the accuracy of ATDA trained with clean source data (ATDA-TCS) as a reference point. When accuracy of one method is close to that of ATDA-TCS (red dash line), this method successfully eliminates noise effects. From our observations, it is clear that B-Net is very close to ATDA-TCS in $7$ out of $8$ tasks (except for \textit{S$\rightarrow$M} task with P$45$, Figure~\ref{fig: result-S2M-part}-(d)), which is an evidence that Butterfly can eliminate noise effects. {Since P$45$ case is the hardest one and we only have finite samples, it is reasonable that B-Net cannot perfectly eliminate noise effects.} An interesting phenomenon is that, B-Net outperforms ATDA-TCS in $2$ \textit{M$\rightarrow$S} tasks (Figure~\ref{fig: result-M2S-part}-(a), (c)). This means that B-Net transfers more useful knowledge (from noisy source data to unlabeled target data) even than ATDA-TCS (from clean source data to unlabeled target data). 

\begin{table*}[h]
  \centering
  \footnotesize
  \caption{Target-domain accuracy on $12$ {human-sentiment} WUDA tasks with the $20\%$ noise rate. Bold values mean the highest values in each row.}
  \vspace{-1em}
    \begin{tabular}{lllllllll}
    \toprule
    Tasks & \multicolumn{1}{p{3.5em}}{DAN} & \multicolumn{1}{p{3.5em}}{DANN} & \multicolumn{1}{p{3.22em}}{ATDA} & \multicolumn{1}{p{3.22em}}{TCL} & \multicolumn{1}{p{4.28em}}{{MEDA}} & \multicolumn{1}{p{4.28em}}{Co+TCL} & \multicolumn{1}{p{4.28em}}{Co+ATDA} & \multicolumn{1}{p{3.5em}}{B-Net} \\
    \midrule
    \emph{B}$\rightarrow$\emph{D} & 68.28\% & 68.08\% & 70.31\% & 71.40\% & 74.81\% &{\color{mydarkblue}67.81\%} & 66.70\% &  \bf{71.84\%} \\
    \emph{B}$\rightarrow$\emph{E} & 63.78\% & 63.53\% & 72.79\% & 65.08\% & 65.18\% &{\color{mydarkblue}60.54\%} & 68.89\% &  \bf{75.92\%} \\
    \emph{B}$\rightarrow$\emph{K} & 65.48\% & 64.63\% & 71.79\% & 66.80\% & 68.65\% &{\color{mydarkblue}61.23\%} & 66.51\% &  \bf{76.32\%} \\
    \emph{D}$\rightarrow$\emph{B} & 64.63\% & 64.52\% & 70.25\% & 67.33\% & 67.63\% &{\color{mydarkblue}65.22\%} & 68.04\% &  \bf{70.56\%} \\
    \emph{D}$\rightarrow$\emph{E} & 65.33\% & 65.16\% & 69.99\% & 66.74\% & 69.51\% &{\color{mydarkblue}64.55\%} & 67.32\% &   \bf{73.73\%} \\
    \emph{D}$\rightarrow$\emph{K} & 65.68\% & 66.28\% & 74.53\% & 68.82\% & 72.24\% &{\color{mydarkblue}67.98\%} & 72.20\% &  \bf{77.97\%} \\
    \emph{E}$\rightarrow$\emph{B} & 60.41\% & 60.15\% & \bf{63.89\%} & 63.13\% & 63.36\% &{\color{mydarkblue}61.18\%} & 61.08\% &  62.22\% \\
    \emph{E}$\rightarrow$\emph{D} & 62.35\% & 61.67\% & 62.30\% & 62.93\% & 66.18\% &{\color{mydarkblue}60.81\%} & 59.77\% &  \bf{63.53\%} \\
    \emph{E}$\rightarrow$\emph{K} & 72.05\% & 71.51\% & 74.00\% & 75.36\% & 75.42\% &{\color{mydarkblue}72.65\%} & 70.85\% &  \bf{78.96\%} \\
    \emph{K}$\rightarrow$\emph{B} & 59.94\% & 59.40\% & \bf{63.53\%} & 62.77\% & 65.13\% &{\color{mydarkblue}60.71\%} & 61.22\% &  63.36\% \\
    \emph{K}$\rightarrow$\emph{D} & 61.46\% & 61.51\% & 64.66\% & 64.16\% & 66.87\% &{\color{mydarkblue}64.15\%} & 64.94\% &  \bf{66.98\%} \\
    \emph{K}$\rightarrow$\emph{E} & 70.60\% & 72.23\% & 74.75\% & 74.14\% & 75.99\% &{\color{mydarkblue}68.95\%} & 69.69\% &  \bf{76.96\%} \\
    \midrule
    Average & 65.00\% & 64.89\% & 69.40\% & 67.39\% & 69.25\% &{\color{mydarkblue}64.65\%} & 66.43\% & \textbf{71.53\%} \\
    \bottomrule
    \end{tabular}%
  \label{tab: Amazon20}%
  \vspace{-1em}
\end{table*}%

\begin{table*}[h]
  \centering
  \footnotesize
  \caption{Target-domain accuracy on $12$ {human-sentiment} WUDA tasks with the $45\%$ noise rate. Bold values mean the highest values in each row.}
  \vspace{-1em}
    \begin{tabular}{lllllllll}
    \toprule
    Tasks & \multicolumn{1}{p{3.5em}}{DAN} & \multicolumn{1}{p{3.5em}}{DANN} & \multicolumn{1}{p{3.22em}}{ATDA} & \multicolumn{1}{p{3.22em}}{TCL}  & \multicolumn{1}{p{4.28em}}{{MEDA}} & \multicolumn{1}{p{4.28em}}{Co+TCL} & \multicolumn{1}{p{4.28em}}{Co+ATDA} & \multicolumn{1}{p{3.5em}}{B-Net} \\
    \midrule
    \emph{B}$\rightarrow$\emph{D} & 52.43\% & 52.98\% & 53.56\% & 54.44\% & 54.50\% &{\color{mydarkblue}53.21\%} & 54.32\% & \bf{56.59\%} \\
    \emph{B}$\rightarrow$\emph{E} & 52.17\% & 53.50\% & 55.14\% & 54.14\% & 54.29\% &{\color{mydarkblue}53.98\%} & \bf{57.34\%} & 55.74\% \\
    \emph{B}$\rightarrow$\emph{K} & 52.89\% & 51.84\% & 51.14\% & 53.32\% & 53.68\% &{\color{mydarkblue}51.77\%} & 53.28\% &  \bf{57.00\%} \\
    \emph{D}$\rightarrow$\emph{B} & 53.11\% & 53.04\% & 54.48\% & 53.27\% & 53.66\% &{\color{mydarkblue}54.85\%} & \bf{55.95\%} & 55.15\% \\
    \emph{D}$\rightarrow$\emph{E} & 51.30\% & 53.04\% & 54.21\% & 53.77\% & 54.11\% &{\color{mydarkblue}55.63\%} & 56.08\% & \bf{58.91\%} \\
    \emph{D}$\rightarrow$\emph{K} & 52.15\% & 53.17\% & 57.99\% & 52.45\% & 52.45\% &{\color{mydarkblue}58.10\%} & 59.94\% &  \bf{66.20\%} \\
    \emph{E}$\rightarrow$\emph{B} & 51.38\% & 51.08\% & 52.54\% & 52.14\% & 52.56\% &{\color{mydarkblue}54.88\%} & 53.30\% &  \bf{54.93\%} \\
    \emph{E}$\rightarrow$\emph{D} & 52.83\% & 51.24\% & 49.02\% & 52.57\% & 53.03\% &{\color{mydarkblue}50.03\%} & 49.62\% &  \bf{52.88\%} \\
    \emph{E}$\rightarrow$\emph{K} & 54.21\% & 53.58\% & 51.66\% & 55.04\% & 55.42\% &{\color{mydarkblue}56.15\%} & 52.10\% &  \bf{56.12\%} \\
    \emph{K}$\rightarrow$\emph{B} & 50.44\% & 51.77\% & \bf{51.96\%} & 51.50\% & 51.52\% &{\color{mydarkblue}53.81\%} & 52.59\% &  51.39\% \\
    \emph{K}$\rightarrow$\emph{D} & 52.20\% & 51.45\% & 52.86\% & 53.19\% & 53.38\% &{\color{mydarkblue}55.69\%} & 54.52\% &  \bf{53.53\%} \\
    \emph{K}$\rightarrow$\emph{E} & \bf{54.72\%} & 53.33\% & 52.11\% & 53.46\% & 53.81\% &{\color{mydarkblue}51.26\%} & 52.62\% & 53.71\% \\
    \midrule
    Average & 52.49\% & 52.50\% & 53.65\% & 53.27\% & 53.54\% &{\color{mydarkblue}54.11\%} & 54.31\% & \textbf{56.01\%} \\
    \bottomrule
    \end{tabular}%
  \label{tab: Amazon45}%
  \vspace{-1em}
\end{table*}%
\vspace{-0.5em}
\begin{table*}[!tp]
  \caption{Target-domain accuracy on $3$ real-world WUDA tasks. The source domain is the \emph{Bing} dataset that contains noisy information from the Internet. Bold value represents the highest accuracy in each row.}
  \vspace{-1em}
  \label{tab: Real}%
  \footnotesize
  \begin{center}
    \begin{tabular}{llllllll}
    \toprule
    Target & \multicolumn{1}{p{3.445em}}{DAN} & DANN & \multicolumn{1}{p{3.335em}}{ATDA} & \multicolumn{1}{p{3.28em}}{TCL} & \multicolumn{1}{p{4.22em}}{Co+TCL}& \multicolumn{1}{p{4.22em}}{Co+ATDA} & \multicolumn{1}{p{3.555em}}{B-Net} \\
    \midrule
    \textit{Caltech256} & 77.83\% & 78.00\% & 80.84\% & 79.35\% &{\color{mydarkblue}79.27\%} & 79.89\% &  \bf{81.71\%} \\
    \textit{Imagenet} & 70.29\% & 72.16\% & 74.89\% & 72.53\% &{\color{mydarkblue}72.33\%} & 74.73\% &  \bf{75.00\%} \\
    \textit{SUN} & 24.56\% & 26.80\% & 26.26\% & 28.80\% &{\color{mydarkblue}29.15\%} & 26.31\% &  \bf{30.54\%} \\
    \midrule
    Average & 57.56\% & 58.99\% & 60.66\% & 60.23\% &{\color{mydarkblue}60.25\%} & 60.31\% &  \bf{62.42\%} \\
    \bottomrule
    \end{tabular}%
    \vspace{-1em}
  \end{center}

\end{table*}%



\vspace{-0.2em}
\subsubsection{Results on human sentiment WUDA tasks}

\label{Asec:Results_Amazon}
Tables \ref{tab: Amazon20} and \ref{tab: Amazon45} report the target-domain accuracy of each method in $24$ human-sentiment WUDA tasks. For these tasks, B-Net has the highest average target-domain accuracy. It should be noted that two-step method does not always perform better than existing UDA methods, such as for $20\%$-noise situation. The reason is that co-teaching performs poorly when pinpointing clean source data from noisy source data. Another observation is that noise effects is not eliminated like target-domain accuracy in $8$ digit WUDA tasks. The reason mainly includes that 1) these datasets only provide predefined features (i.e., we cannot extract better features from original contents in the training process), and 2) we only have finite samples and the number of samples in these datasets is smaller than those of digit datasets. 

\subsection{Results on real-world WUDA tasks} Table~\ref{tab: Real} reports the target-domain accuracy in $3$ tasks. B-Net enjoys the best performance on all tasks. It should be noted that, in \textit{Bing}$\rightarrow$\textit{Caltech256} and  \textit{Bing}$\rightarrow$\textit{ImageNet} tasks, ATDA is slightly worse than B-Net. However, in \textit{Bing}$\rightarrow$\textit{SUN} task, ATDA is much worse than B-Net. The reason is that the DIR between \textit{Bing} and \textit{SUN} are more affected by noisy source data. This is also observed when comparing DANN and TCL. Compared to Co+ATDA, ATDA is slightly better than Co+ATDA. This abnormal phenomenon can be explained using $\Delta$ (see Section~\ref{sec:two-step}), after using co-teaching to assign pseudo labels to noisy source data, the second term in $\Delta_s$ may increase, which results in that $\Delta$ increases, i.e., noise effects actually increase. This phenomenon is an evidence that a two-step method may not really reduce noise effects.

\begin{figure*}[tp]
	\begin{center}
		\subfigure[Values of $\rho_{01}^s$ and $\rho_{01}^t$]
		{\includegraphics[width=0.26\textwidth]{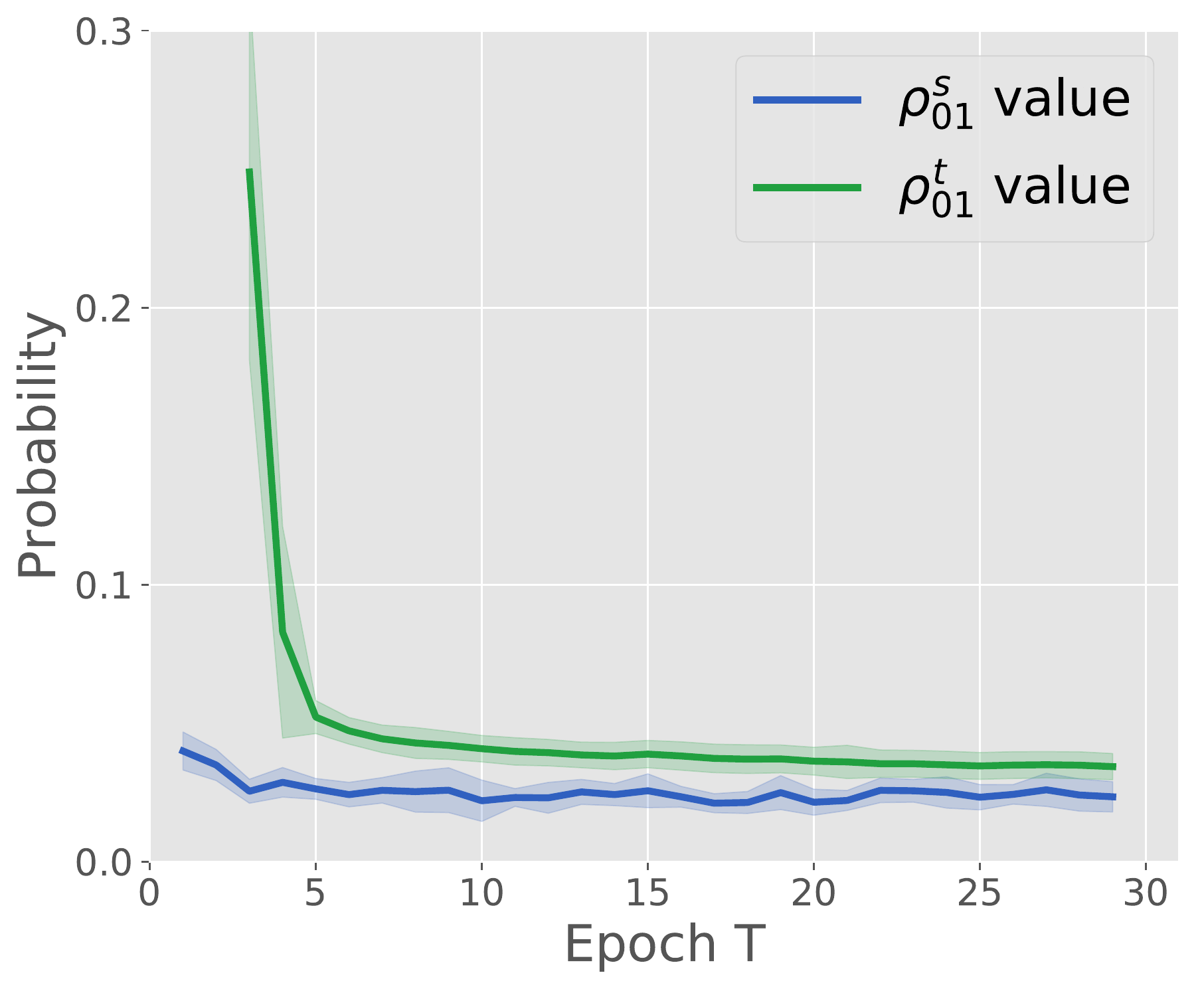}}
		\vspace{-1em}
		\subfigure[Convergence speed of $\rho_{01}^s$]{
		\includegraphics[width=0.26\textwidth]{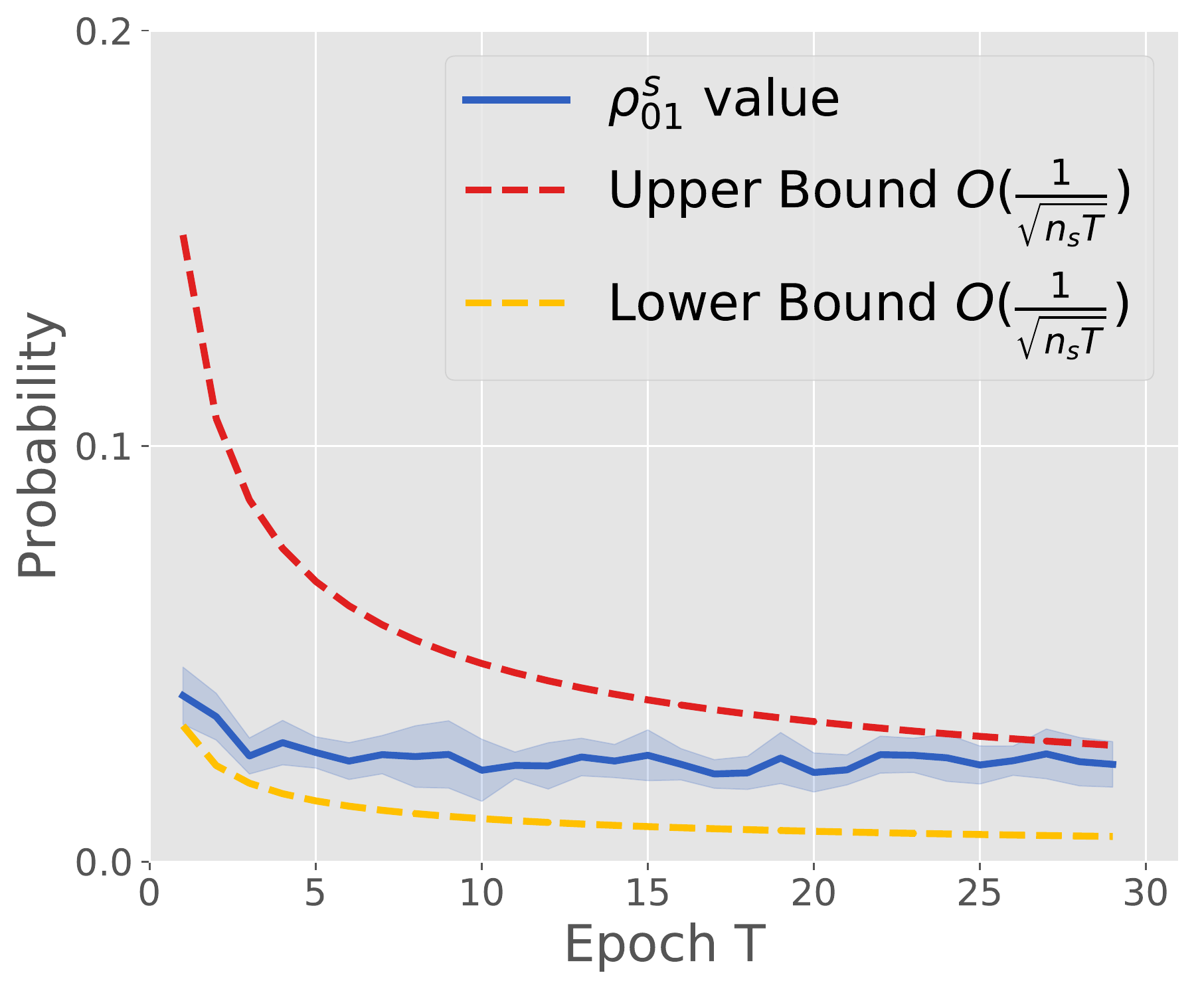}}
		\subfigure[Convergence speed of $\rho_{01}^t$]
		{\includegraphics[width=0.26\textwidth]{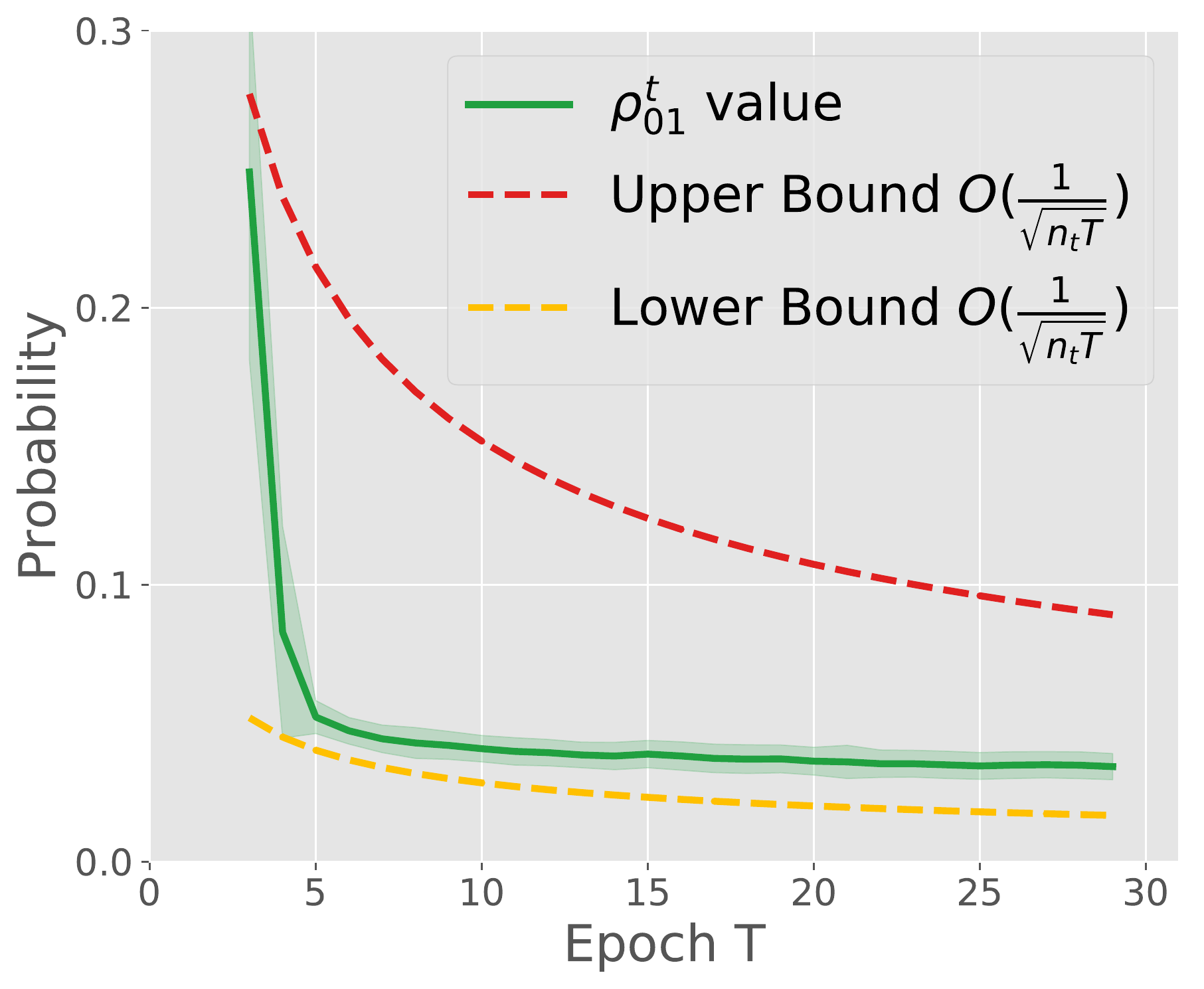}}
		\caption{{The values of $\rho_{01}^s$ and $\rho_{01}^t$ on the \textit{S}$\rightarrow$\textit{M} task (P20), where $n_s=494,000$ is much larger than $n_t=10,000$. Since we do not have pseudo-labeled target data in the first epoch, we illustrate $\rho_{01}^t$ from the second epoch.}}
		\label{fig: result-S2M-thm}
	\end{center}
	\vspace{-1em}
\end{figure*}

\begin{table*}[!t]
  \centering
  \caption{Results of ablation study. Average target-domain accuracy on $8$ simulated digit WUDA tasks (\emph{Digit}), $24$ simulated human-sentiment WUDA tasks (\emph{Sentiment}) and $3$ real-world WUDA tasks (\emph{Real-world}). Bold value represents the highest accuracy in each row.}
  \vspace{-1em}
  \label{tab: ablation}%
    \begin{tabular}{lllllllllll}
    \toprule
    Datasets & Tri-C-Net &B w/o C & DCP-D & DCP-M & B-Net-S & B-Net-T & B-Net-ST & B-Net-M & B-Net \\
    \midrule
    \emph{Digit} & 59.80\% & 74.52\% & 59.19\% & 70.85\% & 71.93\% & 52.00\% & 72.27\% & 73.89\% & \bf{75.82\%} \\
    \emph{Sentiment} & 61.25\% & 63.57\% & 61.37\% & 63.39\% & 61.49\% & 61.12\% & 61.73\% & 62.21\% & \bf{63.77\%} \\
    \emph{Real-world} & 61.50\% & 62.27\% & 59.82\% & 62.34\% & 61.91\% & 60.87\% & 62.24\% & 62.17\% & \bf{62.42\%} \\
    \bottomrule
    \end{tabular}%
    \vspace{-1em}
\end{table*}%

\vspace{-0.5em}

\subsection{Can we check correct data out?}
\label{sec:verify_rho}
{This subsection verifies that $\rho_{01}^s$ and $\rho_{01}^t$ will go to zero with the convergence speed of $O(1/\sqrt{n_sT})$ and $O(1/\sqrt{n_tT})$, respectively. Figure~{\ref{fig: result-S2M-thm}} shows the values of $\rho_{01}^s$ and $\rho_{01}^t$. It can be seen that $\rho_{01}^s$ and $\rho_{01}^t$ will go to zero when increasing the training epochs. $\rho_{01}^s$ is always lower than $\rho_{01}^t$ because that $n_s$ is much larger than $n_t$, indicating that we can check more correct data out when more samples are available. Figure~{\ref{fig: result-S2M-thm}}-(b) shows that we can always find two finite $C_\rho^s$ such that $\rho_{01}^s$ goes to the zero with the convergence speed of $O(1/\sqrt{n_sT})$. So do $C_\rho^t$ and $\rho_{01}^t$ in Figure~{\ref{fig: result-S2M-thm}}-(c).}

\subsection{Ablation study} 
\label{sec:abl_study}
Finally, we conduct thorough experiments to show the contribution of individual components in B-Net. We report average target-domain accuracy on $32$ simulated WUDA tasks ($8$ digit and $24$ human-sentiment WUDA tasks) and $3$ real-world WUDA tasks. We consider following baselines: 

\begin{itemize}
\item Tri-C-Net: \underline{tri}ply \underline{c}heck data in SD, MD and TD. Compared to B-Net, Tri-C-Net has another branch (denoted by Branch-III) to check data in SD. Namely, Tri-C-Net has three branches (i.e., six networks). Parameters of CNN of the Branch-III are the same with that of Branch-I and Branch-II.
\item B w/o C: train \underline{B}-Net by Algorithm~\ref{alg: ButterNET}, \underline{without} adding $|\theta_{f11}^T \theta_{f21}|$ into the loss function of B-Net.
\item DCP-D: realize \underline{DCP} via \underline{D}ecoupling \cite{DeCoupling} to check data in MD and TD. 
\item DCP-M: realize \underline{DCP} via \underline{M}entorNet \cite{jiang2017mentornet} to check data in MD and TD.
\item B-Net-S: train \underline{B-Net} where the check is turned on for \underline{S}ource data in MD. 
\item B-Net-T: train \underline{B-Net} where the check is turned on for \underline{T}arget data in TD.
\item B-Net-ST: train \underline{B-Net} where the checks are turned on for \underline{S}ource data in MD and \underline{T}arget data in TD. 
\item B-Net-M: train \underline{B-Net} where the check is turned on for all data in \underline{M}D. 
\end{itemize}
Note that in the full B-Net, the checks are turned on for all data in MD and TD.
Comparing B-Net with Tri-C-Net shows whether two branches (i.e., four networks) are the optimal design. Comparing B-Net with B w/o C reveals if the constraint $|\theta_{f11}^T \theta_{f21}|$ takes effects. Comparing B-Net with DCP-D and DCP-M shows whether realizing DCP via co-teaching is the optimal way. Comparing B-Net with B-Net-S, B-Net-T, B-Net-ST and B-Net-M reveals if DCP is necessary.

Table~\ref{tab: ablation} reports average target-domain accuracy of above baselines and B-Net. As can be seen, 1) maintaining $4$ networks (like B-Net) is better than maintaining $6$ networks (like Tri-C-Net) since B-Net outperforms Tri-C-Net in terms of average target-domain accuracy; 2) B-Net benefits from adding the constraint to the loss function $\mathcal{L}$; 3) realizing DCP by co-teaching is better than using Decoupling or MentorNet; and 4) DCP is necessary since accuracy of B-Net is higher than those of B-Net-S, B-Net-T, B-Net-ST and B-Net-M.

\vspace{-0.5em}
\section{Conclusions}

This paper opens a new problem called \emph{wildly unsupervised domain adaptation} (WUDA). However, existing UDA methods cannot handle WUDA well. To address this problem, we propose a robust one-step approach called \emph{Butterfly}. Butterfly maintains four deep networks simultaneously: Two take care of all adaptations; while the other two can focus on classification in target domain. We compare Butterfly with existing UDA methods on $32$ simulated and $3$ real-world WUDA tasks. Empirical results demonstrate that Butterfly can robustly transfer knowledge from noisy source data to unlabeled target data. In the  future, we will extend our Butterfly framework to address open-set WUDA, where label space of target domain is larger than that of source domain.

\ifCLASSOPTIONcompsoc
  \section*{Acknowledgments}
\else
  \section*{Acknowledgment}
\fi

FL, JL and GZ were supported by the Australian Research Council (ARC) under FL190100149. BH was supported by the RGC Early Career Scheme No. 22200720 and NSFC Young Scientists Fund No. 62006202, HKBU Tier-1 Start-up Grant, HKBU CSD Start-up Grant, HKBU CSD Departmental Incentive Grant, and a RIKEN BAIHO Award. GN and MS were supported by JST AIP Acceleration Research Grant Number JPMJCR20U3, Japan. MS was also supported by the Institute for AI and Beyond, UTokyo. 

\ifCLASSOPTIONcaptionsoff
  \newpage
\fi

\bibliographystyle{IEEEtran}
\small
\bibliography{bib}
\vspace{-4em}

\begin{IEEEbiography}[{\includegraphics[width=1in,height=1.25in,clip,keepaspectratio]{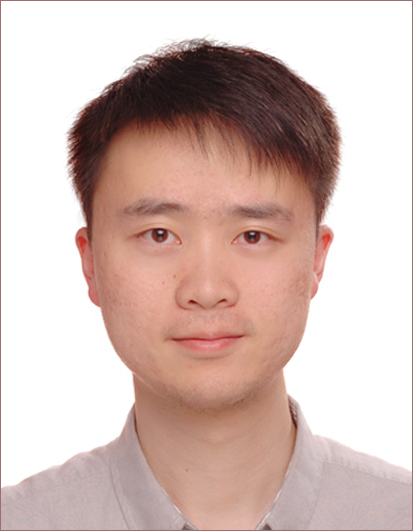}}]{Feng Liu}
is a Doctoral candidate in Centre for Artificial intelligence, Faculty of Engineering and Information Technology, University of Technology Sydney, Australia. He received an M.Sc. degree in probability and statistics and a B.Sc. degree in pure mathematics from the School of Mathematics and Statistics, Lanzhou University, China, in 2015 and 2013, respectively. His research interests include domain adaptation and two-sample test. He has served as a senior program committee member for ECAI and program committee members for NeurIPS, ICML, IJCAI, CIKM, ECAI, FUZZ-IEEE and ISKE. He also served as reviewers for TPAMI, TNNLS, TFS and TCYB. He has received the UTS-FEIT HDR Research Excellence Award (2019), Best Student Paper Award of FUZZ-IEEE (2019) and UTS Research Publication Award (2018).
\end{IEEEbiography}

\begin{IEEEbiography}[{\includegraphics[width=1in,height=1.25in,clip,keepaspectratio]{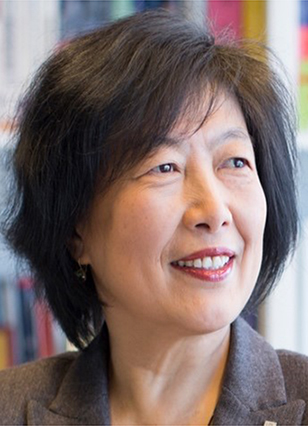}}]{Jie Lu}
(F'18) is a Distinguished Professor and the Director of the Centre for Artificial Intelligence at the University of Technology Sydney, Australia. She received the Ph.D. degree from Curtin University of Technology, Australia, in 2000.

Her main research expertise is in fuzzy transfer learning, decision support systems, concept drift, and recommender systems. She has published six research books and $400$ papers in Artificial Intelligence, IEEE transactions on Fuzzy Systems and other refereed journals and conference proceedings. She has won over $20$ Australian Research Council (ARC) discovery grants and other research grants for over \$$7$ million. She serves as Editor-In-Chief for Knowledge-Based Systems (Elsevier) and Editor-In-Chief for International Journal on Computational Intelligence Systems (Atlantis), has delivered $20$ keynote speeches at international conferences, and has chaired $10$ international conferences. She is a Fellow of IEEE and Fellow of IFSA.
\end{IEEEbiography}

\begin{IEEEbiography}[{\includegraphics[width=1.1in,height=1.375in,clip,keepaspectratio]{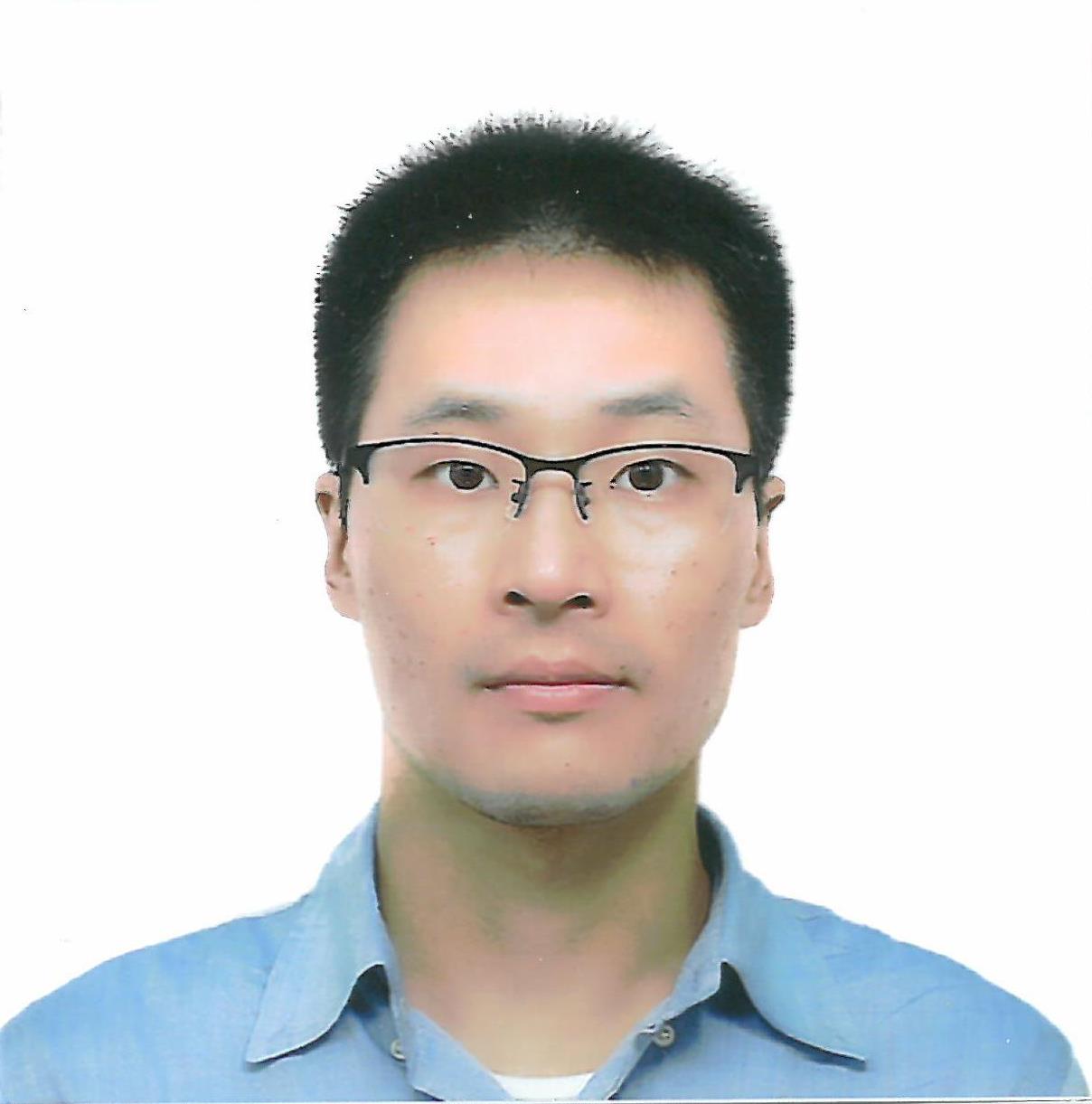}}]{Bo Han} is currently an Assistant Professor of Computer Science at Hong Kong Baptist University and a Visiting Scientist at RIKEN Center for Advanced Intelligence Project (RIKEN AIP), hosted by Masashi Sugiyama. He was a Postdoc Fellow at RIKEN AIP (2019-2020), advised by Masashi Sugiyama. He received his Ph.D. degree in Computer Science from University of Technology Sydney (2015-2019), advised by Ivor W. Tsang and Ling Chen. During 2018-2019, he was a Research Intern with the AI Residency Program at RIKEN AIP, working on robust deep learning projects with Masashi Sugiyama, Gang Niu and Mingyuan Zhou. His current research interests lie in machine learning, deep learning and artificial intelligence. His long-term goal is to develop trustworthy intelligent systems, which can learn from a massive volume of complex (e.g., weakly-supervised, adversarial, and private) data (e.g, single-/multi-label, ranking, domain, similarity, graph and demonstration) automatically. He has served as program committes of NeurIPS, ICML, ICLR, AISTATS, UAI, AAAI, IJCAI, ACML and ICDM. He received the National Scholarship (2013), UTS International Research Scholarship (2014) and UTS Research Publication Award (2017 and 2018).
\end{IEEEbiography}

\begin{IEEEbiography}[{\includegraphics[width=1.1in,height=1.375in,clip,keepaspectratio]{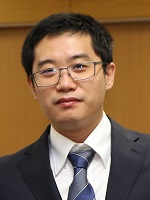}}]{Gang Niu} is a research scientist at RIKEN Center for Advanced Intelligence Project. He received the PhD degree in computer science from Tokyo Institute of Technology in 2013. His research interests include mainly weakly-supervised learning and its applications. He has published $10$ NeurIPS (including $1$ oral and $1$ spotlight) and $10$ ICML papers and also served as an area chair for ICML 2019, NeurIPS 2019 and ICML 2020.
\end{IEEEbiography}

\begin{IEEEbiography}[{\includegraphics[width=1in,height=1.25in,clip,keepaspectratio]{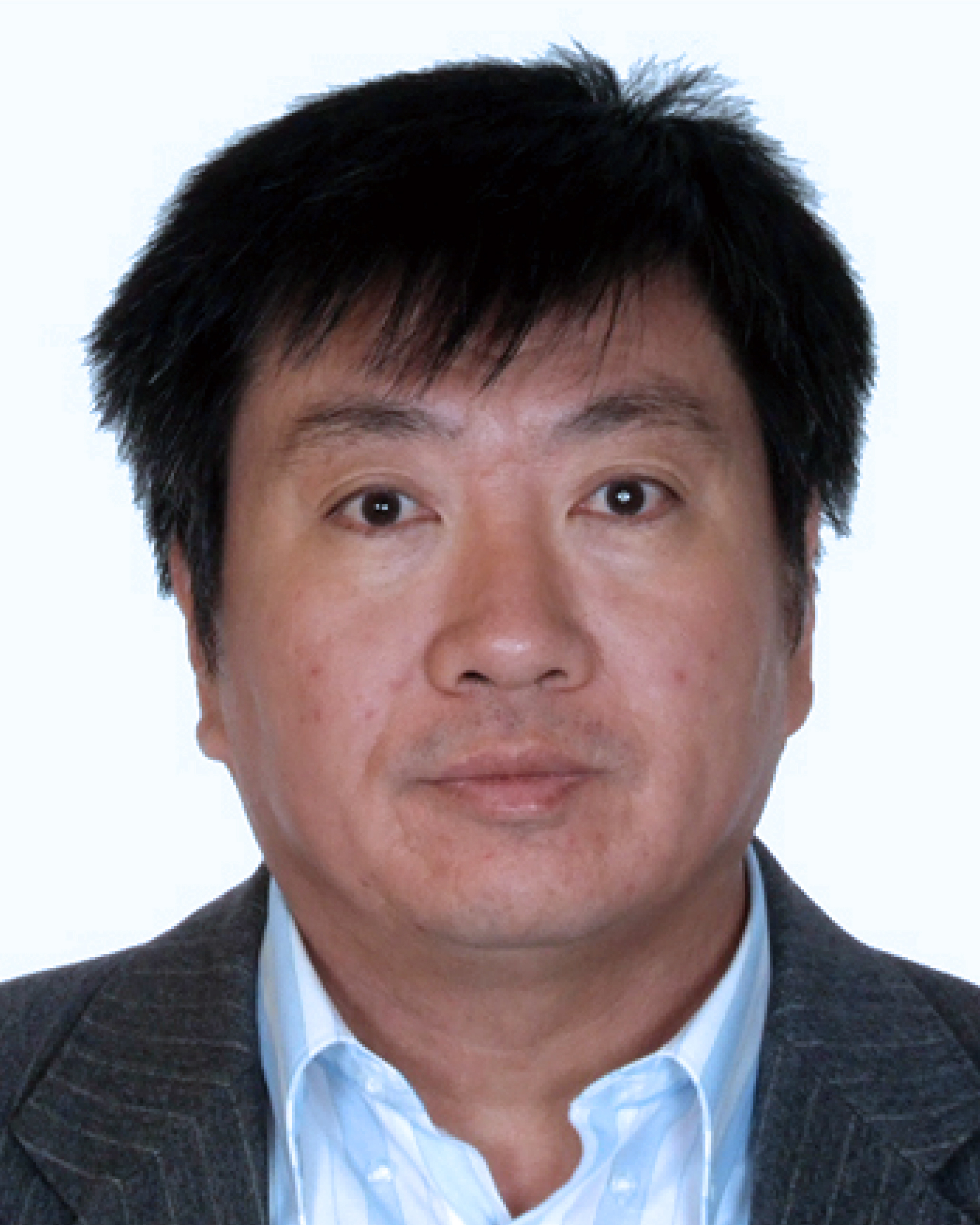}}]{Guangquan Zhang}
is an Associate Professor and Director of the Decision Systems and e-Service Intelligent (DeSI) Research Laboratory at the University of Technology Sydney, Australia. He received the Ph.D. degree in applied mathematics from Curtin University of Technology, Australia, in 2001.

His research interests include fuzzy machine learning, fuzzy optimization, and machine learning. He has authored five monographs, five textbooks, and  $460$ papers including $220$ refereed international journal papers. 
Dr. Zhang has won seven Australian Research Council (ARC) Discovery Projects grants and many other research grants. He was awarded an ARC QEII fellowship in 2005.  
He has served as a member of the editorial boards of several international journals, as a guest editor of eight special issues for IEEE transactions and other international journals, and co-chaired several international conferences and workshops in the area of fuzzy decision-making and knowledge engineering. 
\end{IEEEbiography}

\begin{IEEEbiography}[{\includegraphics[width=1.1in,height=1.375in,clip,keepaspectratio]{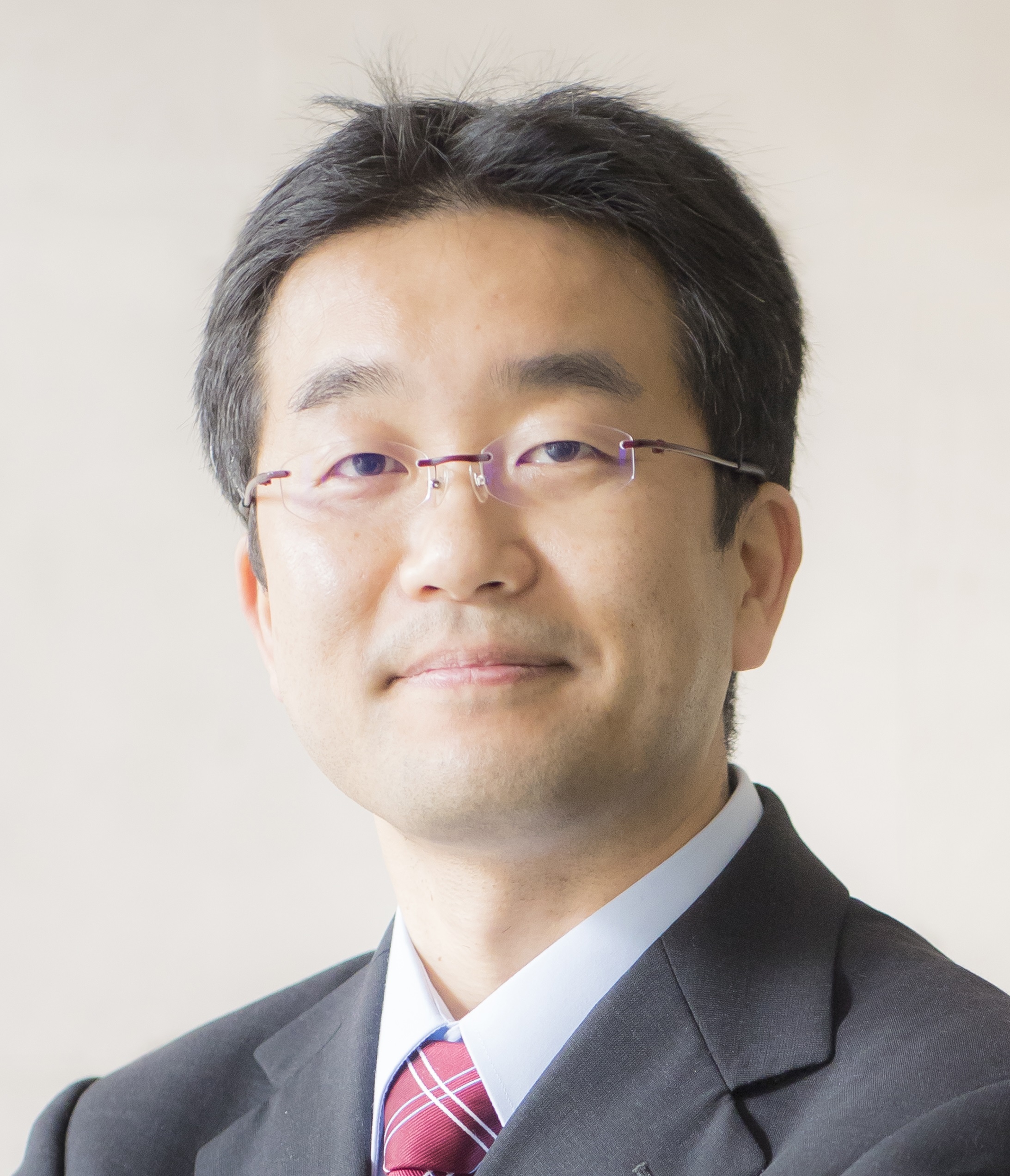}}]{Masashi Sugiyama} is Director of RIKEN Center for Advanced Intelligence Project and Professor at the University of Tokyo. He received the PhD degree in computer science from Tokyo Institute of Technology in 2001. His research interests include theories and algorithms of machine learning. He was awarded the Japan Society for the Promotion of Science Award and the Japan Academy Medal in 2017.
\end{IEEEbiography}
\newpage
\newpage
\appendices
\onecolumn

\section{Proofs}
\label{Asec:proofs}
This section presents the completed proofs for theoretical results obtained in this paper. Since we have provided completed proofs regarding Theorems~\ref{thm: epsilon_effects} and \ref{thm:generalization}, we do not repeat them here.
\subsection{Proof of Theorem \ref{thm: risks relation}}
\begin{proof}
We will fist prove Eq.~(\ref{eq: risks relation_Q_TM}) (Case $1$) and then prove Eq.~(\ref{eq: risks relation}) (Case $2$).

\textbf{Case $1$}. According to definition of $\tilde{R}_s(h)$, we have
\begin{align}
\label{eq:thm1:1}
    \tilde{R}_s(h) & =\mathbb{E}_{\tilde{p}_s(x_s,\tilde{y}_s)}[\ell(h(x_s),\tilde{y}_s)] \nonumber \\
    & = \int_{\mathcal{X}}\sum_{\tilde{y}_s=1}^K \ell(h(x_s),\tilde{y}_s)\tilde{p}_s(x_s,\tilde{y}_s)dx_s \nonumber \\
    & = \int_{\mathcal{X}}\sum_{\tilde{y}_s=1}^K \ell(h(x_s),\tilde{y}_s)\tilde{p}_{\tilde{Y}_s|X_s}(\tilde{y}_s|x_s)p_{x_s}(x_s)dx_s\nonumber \\
    & = \int_{\mathcal{X}} \bm{\tilde{\eta}}^T(x_s)\bm{\ell}(h(x_s))p_{x_s}(x_s)dx_s,
\end{align}
where $\bm{\ell}(h(x_s)) = [\ell(h(x_s),1),..., \ell(h(x_s),K)]^T$ and $\bm{\tilde{\eta}}(x_s)=[\tilde{p}_{\tilde{Y}_s|X_s}(1|x_s), \dots,$ $ \tilde{p}_{\tilde{Y}_s|X_s} (K|x_s)]^T$. According to definition of the transition matrix $Q$, we know that
\begin{align}
\label{eq:thm1:2}
    \bm{\tilde{\eta}}^T(x_s) = \bm{{\eta}}^T(x_s)Q,
\end{align}
where $\bm{\eta}(x_s)=[p_{Y_s|X_s}(1|x_s),...~,p_{Y_s|X_s}(K|x_s)]^T$. Substituting Eq.~(\ref{eq:thm1:2}) into Eq.~(\ref{eq:thm1:1}), we have
\begin{align*}
    \tilde{R}_s(h) &= \int_{\mathcal{X}} \bm{{\eta}}^T(x_s)Q\bm{\ell}(h(x_s))p_{x_s}(x_s)dx_s \\
    & = \int_{\mathcal{X}} \bm{{\eta}}^T(x_s)I\bm{\ell}(h(x_s))p_{x_s}(x_s)dx_s + \int_{\mathcal{X}} \bm{{\eta}}^T(x_s)(Q-I)\bm{\ell}(h(x_s))p_{x_s}(x_s)dx_s \\
    & = R_s(h) + \mathbb{E}_{p_{x_s}(x_s)}[\bm{\eta}^T(x_s)(Q-I)\bm{\ell}(h(x_s))].
\end{align*}
Hence, Case $1$ is proved.

\textbf{Case $2$}. According to definition of $\tilde{R}_s(h)$ and Eq.~(\ref{eq: noise_observ_simple}), we have
\begin{align}
\label{eq:thm1:3}
   \tilde{R}_s(h)  &=\mathbb{E}_{\tilde{p}_s(x_s,\tilde{y}_s)}[\ell(h(x_s),\tilde{y}_s)] \nonumber \\
    & = \int_{\mathcal{X}}\sum_{\tilde{y}_s=1}^K \ell(h(x_s),\tilde{y}_s)\tilde{p}_s(x_s,\tilde{y}_s)dx_s \nonumber \\
    & = \int_{\mathcal{X}}\sum_{y_s=1}^K \ell(h(x_s),y_s)\big((1-\rho)p_s(x_s,y_s)+\rho q_s(x_s,y_s)\big)dx_s\nonumber \\
    & = (1-\rho)\int_{\mathcal{X}}\sum_{y_s=1}^K \ell(h(x_s),y_s)p_s(x_s,y_s)dx_s + \rho \int_{\mathcal{X}}\sum_{y_s=1}^K \ell(h(x_s),y_s)q_s(x_s,y_s)dx_s \nonumber \\
    & = (1-\rho)R_s(h) + \rho \int_{\mathcal{X}}\sum_{y_s=1}^K \ell(h(x_s),y_s)q_{Y_s|X_s}(y_s|x_s)q_{x_s}(x_s)dx_s.
\end{align}
Let $\bm{\eta_q}(x_s)=[q_{Y_s|X_s}(1|x_s),..., q_{Y_s|X_s}(K|x_s)]^T$, we have
\begin{align*}
    \tilde{R}_s(h) = (1-\rho)R_s(h) + \rho\mathbb{E}_{q_{x_s}(x_s)}[\bm{\eta_q}^T(x_s)\bm{\ell}(h(x_s))].
\end{align*}
Hence, Case $2$ is proved.
\end{proof}

\subsection{Proof of Theorem \ref{thm:upper_bound_target}}
\begin{proof}
For any $h\in\mathcal{H}$, we have
\begin{align}
\label{eq:thm2:1}
    R_t(h,f_t) &= R_t(h,f_t) + \tilde{R}_s(h) - \tilde{R}_s(h) + R_s(h,f_t) - R_s(h,f_t) \nonumber \\
    &=\tilde{R}_s(h) + R_t(h,f_t) - \tilde{R}_s(h,f_t) + R_s(h,f_t) - R_s(h)  +R_s(h)- \tilde{R}_s(h)+ \tilde{R}_s(h,f_t) - R_s(h,f_t).
\end{align}
Since we do not know $f_t$, we substitute following equations into Eq.~(\ref{eq:thm2:1}),
\begin{align*}
    R_t(h,f_t) = R_t(h,\tilde{f}_t) + R_t(h,f_t) - R_t(h,\tilde{f}_t), \\
    \tilde{R}_s(h,f_t) = \tilde{R}_s(h,\tilde{f}_t) + \tilde{R}_s(h,f_t) - \tilde{R}_s(h,\tilde{f}_t), \\
    R_s(h,f_t) = R_s(h,\tilde{f}_t) + R_s(h,f_t) - R_s(h,\tilde{f}_t).
\end{align*}
Then, we have
\begin{align*}
    R_t(h,f_t) =~&{\tilde{R}_s(h)} +{R_t(h,\tilde{f}_t) - \tilde{R}_s(h,\tilde{f}_t)} + {R_s(h,\tilde{f}_t) - R_s(h)} \\
    &+~{R_s(h) - \tilde{R}_s(h)+\tilde{R}_s(h,\tilde{f}_t) - R_s(h,\tilde{f}_t)}+{R_t(h,f_t) -  R_t(h,\tilde{f}_t)} \nonumber \\
    \leq~&{\tilde{R}_s(h)} +{|R_t(h,\tilde{f}_t) - \tilde{R}_s(h,\tilde{f}_t)|} + {|R_s(h,\tilde{f}_t) - R_s(h)|} \\ &+~{|\tilde{R}_s(h)-R_s(h)|+|\tilde{R}_s(h,\tilde{f}_t) - R_s(h,\tilde{f}_t)|}+{|R_t(h,f_t) -  R_t(h,\tilde{f}_t)|}.
\end{align*}
Hence, this theorem is proved.
\end{proof}

\subsection{Proof of Lemma 1}
According to definition of $\tilde{R}_s^{\text{po}}(h,u_s)$ in Section~\ref{sec:butter_eliminate_noise}, we have
\begin{align*}
    \tilde{R}_s^{\text{po}}(h,u_s)
    & = \frac{\int_\mathcal{X}\sum_{u_s=0}^1\sum_{y_s=1}^Ku_s\ell(h(x_s),y_s)\tilde{p}_s^{\text{po}}(x_s,y_s,u_s)dx_s}{1-\rho_{u_s}} \nonumber \\
    & = \frac{\int_\mathcal{X}\sum_{y_s=1}^K\ell(h(x_s),y_s)\tilde{p}_{X_s,Y_s|U_s}^{\text{po}}(x_s,y_s|1)\tilde{p}_{U_s}^{\text{po}}(1)dx_s}{1-\rho_{u_s}} \nonumber \\
    & \stackrel{\mathclap{(a)}}{=} \frac{1-\rho_{u_s}}{1-\rho_{u_s}}\int_\mathcal{X}\sum_{y_s=1}^K\ell(h(x_s),y_s)\big( \rho_{01}^sq_s(x_s,y_s)+ \rho_{11}^sp_s(x_s,y_s)\big)dx_s \nonumber \\
    & = \rho_{01}^s\mathbb{E}_{q_s(x_s,y_s)}[\ell(h(x_s),y_s)] + \rho_{11}^sR_s(h),
\end{align*}
where $(a)$ is based on the definition of $\rho_{u_s}$ and Eq.~(\ref{eq: transit_u_s}). Thus, we have
\begin{align*}
    |\tilde{R}_s^{\text{po}}(h,u_s) - R_s(h)| 
    & = |\rho_{01}^s\mathbb{E}_{q_s(x_s,y_s)}[\ell(h(x_s),y_s)] - (1-\rho_{11}^s)R_s(h)|\nonumber \\
    & \leq \rho_{01}^s\max \{\mathbb{E}_{q_s(x_s,y_s)}[\ell(h(x_s),y_s)], R_s(h)\}.
\end{align*}
This lemma is proved.


\subsection{Proof of Lemma 2}
According to definition of $\tilde{R}_t^{\text{po}}(h,\tilde{f}_t,u_t)$ in Section~\ref{sec:butter_eliminate_noise}, we have
\begin{align}
\label{eq:lem:2:1}
     \tilde{R}_t^{\text{po}}(h,\tilde{f}_t,u_t)
     & = (1-\rho_{u_t})^{-1}\int_\mathcal{X}\sum_{u_t=0}^1u_t\ell(h(x_t),\tilde{f}_t(x_t))\tilde{p}_t^{\text{po}}(x_t,u_t)dx_t \nonumber \\
    & = (1-\rho_{u_t})^{-1}\int_\mathcal{X}\ell(h(x_t),\tilde{f}_t(x_t))\tilde{p}_{X_t|U_t}^{\text{po}}(x_t|1)\tilde{p}_{U_t}^{\text{po}}(1)dx_t \nonumber \\
    & \stackrel{\mathclap{(a)}}{=} \frac{1-\rho_{u_t}}{1-\rho_{u_t}}\int_\mathcal{X}\ell(h(x_s),\tilde{f}_t(x_t))\big( \rho_{01}^tq_{x_t}(x_t)+ \rho_{11}^t{p}_{X_t|V_t}^{\text{po}}(x_t|1)\big)dx_t \nonumber \\
    & = \rho_{01}^t\mathbb{E}_{q_{x_t}(x_t)}[\ell(h(x_t),\tilde{f}_t(x_t))] + \rho_{11}^t\int_\mathcal{X}\ell(h(x_t),\tilde{f}_t(x_t))p_{X_t|V_t}^{\text{po}}(x_t|V_t=1)dx_t \nonumber \\
    & \stackrel{\mathclap{(b)}}{=} \rho_{01}^t\mathbb{E}_{q_{x_t}(x_t)}[\ell(h(x_t),\tilde{f}_t(x_t))] + \rho_{11}^t\int_\mathcal{X}\ell(h(x_t),f_t(x_t))p_{X_t|V_t}^{\text{po}}(x_t|V_t=1)dx_t \nonumber \\
    & = \rho_{01}^t\mathbb{E}_{q_{x_t}(x_t)}[\ell(h(x_t),\tilde{f}_t(x_t))] + \rho_{11}^t\int_\mathcal{X}\ell(h(x_t),f_t(x_t))p_{x_t}^{\prime}(x_t)dx_t \nonumber \\
    & = \rho_{01}^t\mathbb{E}_{q_{x_t}(x_t)}[\ell(h(x_t),\tilde{f}_t(x_t))] + \rho_{11}^t\mathbb{E}_{p_{x_t}^{\prime}(x_t)}[\ell(h(x_t),f_t(x_t))],
\end{align}
where $(a)$ is based on the definition of $\rho_{u_s}$ and Eq.~(\ref{eq: transit_u_s}) and $(b)$ is based on the definition of $V_t$ ($f_t(x_t)=\tilde{f}_t(x_t)$ when $V_t=1$). Since $\mathbb{E}_{p_{x_t}^{\prime}(x_t)}[\ell(h(x_t),f_t(x_t))]\leq R_t(h,f_t)+\rho_{01}^sM_t$, we have
\begin{align}
    \tilde{R}_t^{\text{po}}(h,\tilde{f}_t,u_t)\leq \rho_{01}^t\mathbb{E}_{q_{x_t}(x_t)}[\ell(h(x_t),\tilde{f}_t(x_t))] + \rho_{11}^t(R_t(h,f_t)+\rho_{01}^sM_t).
\end{align}
Thus, we have
\begin{align*}
    |\tilde{R}_t^{\text{po}}(h,\tilde{f}_t,u_t)-R_t(h,{f}_t)| 
    & = |\rho_{01}^t\mathbb{E}_{q_{x_t}(x_t)}[\ell(h(x_t),\tilde{f}_t(x_t))] + \rho_{11}^t\mathbb{E}_{p_{x_t}^{\prime}(x_t)}[\ell(h(x_t),f_t(x_t))] - R_t(h,{f}_t)|\nonumber \\
    & \leq |\rho_{01}^t\mathbb{E}_{q_{x_t}(x_t)}[\ell(h(x_t),\tilde{f}_t(x_t))] + \rho_{11}^t(R_t(h,f_t)+\rho_{01}^sM_t) - R_t(h,{f}_t)|\nonumber \\
    & = |\rho_{01}^t(\mathbb{E}_{q_{x_t}(x_t)}[\ell(h(x_t),\tilde{f}_t(x_t))] -R_t(h,f_t))+\rho_{11}^t\rho_{01}^sM_t|\nonumber \\
    & \leq \rho^t_{01} \max\{\mathbb{E}_{q_{x_t}(x_t)}[\ell(h(x_t),\tilde{f}_t(x_t))],R_t(h,f_t)\}+\rho_{11}^t\rho_{01}^sM_t.
\end{align*}
This lemma is proved.

\subsection{Proof of Lemma~\ref{lem:R_s_po_est_err_bound}}
For simplicity, in this proof, we let $\mathcal{L}_{S_s}(\ell,  h)=\mathcal{L}(\theta,h;\bm{u}_s,D^{xy}_s)$, $\tilde{R}^{\text{po}}_s(\ell,  h) = \tilde{R}^{\text{po}}_s(h,u_s)$, and $\mathbb{E}_{S_s}[\cdot]=\mathbb{E}_{S_s\sim (\tilde{P}_s^{\text{po}})^n}[\cdot]$, where $\tilde{P}_s^{\text{po}}$ is the probability measure corresponding to the density $\tilde{p}_s^{\text{po}}$. We first show that $\mathcal{L}_{S_s}(\ell,  h)$ is an unbiased estimator of $\tilde{R}^{\text{po}}_s(\ell,  h)$ based on the definition of $\tilde{R}^{\text{po}}_s(h,u_s)$ in Section~\ref{sec:butter_eliminate_noise}. Since $S_s=\{(x_{si},y_{si},u_{si})\}_{i=1}^n$ are i.i.d samples from $\tilde{P}_s^{\text{po}}$, $\mathbb{E}_{S_s}[\mathcal{L}_{S_s}(\ell,  h)]$ can be expressed as follows.
\begin{align}
    &\mathbb{E}_{S_s}\left[\frac{1}{\sum_{i=1}^nu_{si}}\sum_{i=1}^n u_{si}\ell(h(x_{si}),y_{si})\right] \nonumber \\
    =~&\int_{\mathcal{X}}\frac{1}{\sum_{i=1}^nu_{si}}\sum_{i=1}^n\sum_{u_{si}=0}^1\sum_{y_{si}=1}^K u_{si}\ell(h(x_{si}),y_{si})d\tilde{P}_{s}^{\text{po}} \nonumber \\
    =~&\frac{1}{n}\int_{\mathcal{X}}\frac{n}{\sum_{i=1}^nU_i}\sum_{i=1}^n\sum_{u_{si}=0}^1\sum_{y_{si}=1}^K u_{si}\ell(h(x_{si}),y_{si})d\tilde{P}_{s}^{\text{po}} \nonumber \\
    =~&\frac{1}{n}\int_{\mathcal{X}}(1-\rho_{u_s})^{-1}\sum_{i=1}^n\sum_{u_{si}=0}^1\sum_{y_{si}=1}^K u_{si}\ell(h(x_{si}),y_{si})d\tilde{P}_{s}^{\text{po}} \nonumber \\
    =~&(1-\rho_{u_s})^{-1}\frac{1}{n}\sum_{i=1}^n\int_{\mathcal{X}}\sum_{u_{si}=0}^1\sum_{y_{si}=1}^K u_{si}\ell(h(x_{si}),y_{si})d\tilde{P}_{s}^{\text{po}} \nonumber \\
    =~&(1-\rho_{u_s})^{-1}\frac{1}{n}\sum_{i=1}^n\mathbb{E}_{\tilde{p}_s^{\text{po}}(x_s,y_s,u_s)}[u_s\ell(h(x_s),y_s)] \nonumber \\
    =~&\tilde{R}^{\text{po}}_s(h,u_s)=\tilde{R}^{\text{po}}_s(\ell,  h) \label{eq:unbiased_L},
\end{align}
which means that $\mathcal{L}_{S_s}(\ell,h)$ is an unbiased estimator of $\tilde{R}^{\text{po}}_s(\ell,h)$. Then, let $\Phi(S_s)=\sup_{\ell \in \mathbb{L}_{\mathcal{H}}}\big(\tilde{R}^{\text{po}}_s(\ell,h) - \mathcal{L}_{S_s}(\ell,h)\big)$. Changing a point of $S_s$ affects $\Phi(S_s)$ at most $C_L/(n(1-\tau_s))$. Thus, by McDiarmid’s inequality applied to $\Phi(S_s)$, for any $\delta>0$, with probability of at least $1-\delta/2$, the following inequality holds.
\begin{align}
\label{eq:th4:main}
    \Phi(S_s)\leq \mathbb{E}_{S_s}[\Phi(S_s)] + \frac{C_L}{1-\tau_s}\sqrt{\frac{\ln(\delta/2)}{2n}}.
\end{align}
Then, we have
\begin{align}
    &\mathbb{E}_{S_s}[\Phi(S_s)]=\mathbb{E}_{S_s}\Big[\sup_{\ell \in \mathbb{L}_{\mathcal{H}}}\big(\tilde{R}^{\text{po}}_s(h,u_s) - \mathcal{L}_{S_s}(h)\big)\Big] \nonumber \\
    =~&\mathbb{E}_{S_s}\Big[\sup_{\ell \in \mathbb{L}_{\mathcal{H}}}\big(\mathbb{E}_{S_s^{\prime}}[\mathcal{L}_{S_s^{\prime}}(\ell,h)] - \mathcal{L}_{S_s}(h)\big)\Big] \label{eq:part1}\\
    =~&\mathbb{E}_{S_s}\Big[\sup_{\ell \in \mathbb{L}_{\mathcal{H}}}\big(\mathbb{E}_{S_s^{\prime}}[\mathcal{L}_{S_s^{\prime}}(\ell,h) - \mathcal{L}_{S_s}(h)]\big)\Big] \nonumber \\
    \leq~& \mathbb{E}_{S_s,S_s^{\prime}}\Big[\sup_{\ell \in \mathbb{L}_{\mathcal{H}}}\big(\mathcal{L}_{S_s^{\prime}}(\ell,h) - \mathcal{L}_{S_s}(h)\big)\Big] \label{eq:part2} \\
    =~& \frac{1}{n}\mathbb{E}_{S_s,S_s^{\prime}}\Big[\sup_{\ell \in \mathbb{L}_{\mathcal{H}}}\sum_{i=1}^n\Big(\frac{u_{si}^{\prime}\ell(h(x_{si}^{\prime}),y_{si}^{\prime})-u_{si}\ell(h(x_{si}),y_{si}))}{1-\tau_s} \Big)\Big] \nonumber \\
    =~& \frac{1}{n}\mathbb{E}_{\sigma,S_s,S_s^{\prime}}\Big[\sup_{\ell \in \mathbb{L}_{\mathcal{H}}}\sum_{i=1}^n\sigma_i\Big(\frac{u_{si}^{\prime}\ell(h(x_{si}^{\prime}),y_{si}^{\prime})-u_{si}\ell(h(x_{si}),y_{si}))}{1-\tau_s} \Big)\Big] \nonumber \\
    \leq~&\frac{1}{n(1-\tau_s)}\mathbb{E}_{\sigma,S_s^{\prime}}\Big[\sup_{\ell \in \mathbb{L}_{\mathcal{H}}}\sum_{i=1}^n{\sigma_iu_{si}^{\prime}\ell(h(x_{si}^{\prime}),y_{si}^{\prime})}\Big] + \frac{1}{n(1-\tau_s)}\mathbb{E}_{\sigma,S_s}\Big[\sup_{\ell \in \mathbb{L}_{\mathcal{H}}}\sum_{i=1}^n{-\sigma_iu_{si}\ell(h(x_{si}),y_{si})}\Big] \label{eq:part3} \\
    =~& \frac{2}{n(1-\tau_s)}\mathbb{E}_{\sigma,S_s}\Big[\sup_{\ell \in \mathbb{L}_{\mathcal{H}}}\sum_{i=1}^n{\sigma_iu_{si}\ell(h(x_{si}),y_{si})}\Big] \label{eq:final_part},
\end{align}
where Eq.~\eqref{eq:part1} is based on Eq.~\eqref{eq:unbiased_L}, Inequalities \eqref{eq:part2} and \eqref{eq:part3} are based on Jensen's Inequality. Because of existence of $u_{si}$, Eq.~\eqref{eq:final_part} is not the Rademacher complexity of $\mathbb{L}_{\mathcal{H}}$ (i.e., $\Re(\mathbb{L}_{\mathcal{H}})$). However, in following, we prove that Eq.~\eqref{eq:final_part} can be bounded by $\Re(\mathbb{L}_{\mathcal{H}})/(1-\tau_s)$. 
\begin{align}
    &\mathbb{E}_{\sigma,S_s}\Big[\sup_{\ell \in \mathbb{L}_{\mathcal{H}}}\sum_{i=1}^n{\sigma_iu_{si}\ell(h(x_{si}),y_{si})}\Big] \nonumber \\
    =~&\mathbb{E}_{\sigma,S_s}\Big[\sup_{\ell \in \mathbb{L}_{\mathcal{H}}}\Big(\sigma_1u_{s1}\ell(h(x_{s1}),y_{s1})+\sum_{i=2}^n{\sigma_iu_{si}\ell(h(x_{si}),y_{si})}\Big)\Big] \nonumber \\
    =~& \frac12\mathbb{E}_{\sigma,S_s}\Big[\sup_{\ell,\ell^{\prime} \in \mathbb{L}_{\mathcal{H}}}\Big(u_{s1}\ell(h(x_{s1}),y_{s1})+\sum_{i=2}^n{\sigma_iu_{si}\ell(h(x_{si}),y_{si})} + (-u_{s1})\ell^{\prime}(h(x_{s1}),y_{s1})+\sum_{i=2}^n{\sigma_iu_{si}\ell^{\prime}(h(x_{si}),y_{si})}\Big)\Big] \nonumber
\end{align}
\begin{align}
    =~& \frac12\mathbb{E}_{\sigma,S_s}\Big[\sup_{\ell,\ell^{\prime} \in \mathbb{L}_{\mathcal{H}}}\Big(u_{s1}(\ell(h(x_{s1}),y_{s1})-\ell^{\prime}(h(x_{s1}),y_{s1})) + \sum_{i=2}^n{\sigma_iu_{si}\ell(h(x_{si}),y_{si})}+\sum_{i=2}^n{\sigma_iu_{si}\ell^{\prime}(h(x_{si}),y_{si})}\Big)\Big] \nonumber \\
    \leq~& \frac12\mathbb{E}_{\sigma,S_s}\Big[\sup_{\ell,\ell^{\prime} \in \mathbb{L}_{\mathcal{H}}}\Big(\ell(h(x_{s1}),y_{s1})-\ell^{\prime}(h(x_{s1}),y_{s1}) + \sum_{i=2}^n{\sigma_iu_{si}\ell(h(x_{si}),y_{si})}+\sum_{i=2}^n{\sigma_iu_{si}\ell^{\prime}(h(x_{si}),y_{si})}\Big)\Big] \label{eq:2nd_part1} \\
    =~&\mathbb{E}_{\sigma,S_s}\Big[\sup_{\ell \in \mathbb{L}_{\mathcal{H}}}\Big(\sigma_1\ell(h(x_{s1}),y_{s1})+\sum_{i=2}^n{\sigma_iu_{si}\ell(h(x_{si}),y_{si})}\Big)\Big], \nonumber 
\end{align}
where Inequality \eqref{eq:2nd_part1} is based on the fact that there are always $\ell,\ell^{\prime}\in\mathbb{L}_{\mathcal{H}}$ such that $\ell(h(x_{s1}),y_{s1})-\ell^{\prime}(h(x_{s1}),y_{s1})>0$. Repeat above procedures $n-1$ times, we have
\begin{align}
\label{eq:2nd_final}
    \frac{2}{n}\mathbb{E}_{\sigma,S_s}\Big[\sup_{\ell \in \mathbb{L}_{\mathcal{H}}}\sum_{i=1}^n{\sigma_iu_{si}\ell(h(x_{si}),y_{si})}\Big]
    \leq\frac{2}{n}\mathbb{E}_{\sigma,S_s}\Big[\sup_{\ell \in \mathbb{L}_{\mathcal{H}}}\sum_{i=1}^n{\sigma_i\ell(h(x_{si}),y_{si})}\Big]:=\Re_n(\mathbb{L}_{\mathcal{H}}).
\end{align}
Changing a point of $S_s$ affects $\Re_n(\mathbb{L}_\mathcal{H})$ at most $2C_L/n$. Thus, by McDiarmid’s inequality, for any $\delta>0$, with probability of at least $1-\delta/2$, the following inequality holds.
\begin{align}
\label{eq:th4:last}
    \Re_n(\mathbb{L}_{\mathcal{H}})\leq\hat{\Re}_{S_s}(\mathbb{L}_{\mathcal{H}}) + 2C_L\sqrt{\frac{\ln(\delta/2)}{2n}}.
\end{align}
Since $\ell$ is Lipschitz continuous, according to \cite{maurer2016vector}, we have
\begin{align}
\label{eq:th4:lastlast}
    \hat{\Re}_{S_s}(\mathbb{L}_{\mathcal{H}})\leq \sqrt{2}L_{\ell}\hat{\Re}_{D_s^x}(\mathcal{H}).
\end{align}
Combining \eqref{eq:th4:main}, \eqref{eq:final_part}, \eqref{eq:2nd_final}, \eqref{eq:th4:last} and \eqref{eq:th4:lastlast}, we prove this lemma.

\subsection{Proof of Corollary~\ref{cor:generalization}}
We prove this corollary (i.e., Inequality \eqref{eq:empicial_bound_NEW_cor}) according to Inequality \eqref{eq:risk_bound_NEW:basic}, where \eqref{eq:empicial_bound_NEW_cor} has $8$ terms in the right side and \eqref{eq:risk_bound_NEW:basic} have $6$ terms in the right side.

1) For last $3$ terms in \eqref{eq:risk_bound_NEW:basic}, since $\rho^s_{01}<C_\rho^s/\sqrt{n_sT}$  and $\rho^t_{01}<C_\rho^t/\sqrt{n_tT}$, according to \eqref{eq:first2_epsilon}, \eqref{eq:second2_epsilon} and \eqref{eq:third2_epsilon}, we know the sum of last three terms of \eqref{eq:risk_bound_NEW:basic} is less than or equal to $C_\rho^s(M_s+M_t)/{\sqrt{n_sT}} + {2C_\rho^tM_t}/{\sqrt{n_tT}}$ (i.e., the last $2$ terms in \eqref{eq:empicial_bound_NEW_cor}).

2) For first $3$ terms in \eqref{eq:risk_bound_NEW:basic}, we have shown that (in Section~\ref{sec:principle_butterfly}) the sum of the first $3$ terms in \eqref{eq:risk_bound_NEW:basic} is less than or equal to $(*)$:
\begin{align*}
    2\tilde{R}_s^{\text{po}}(h,u_s)+2\tilde{R}^{\text{po}}_s(h,\tilde{f}_t,\bm{u}_s)+\tilde{R}^{\text{po}}_t(h,\tilde{f}_t,\bm{u}_t)+\frac{C_\rho^sM_s}{\sqrt{n_sT}}+\frac{C_\rho^tM_t}{\sqrt{n_tT}}.
\end{align*}
Then, we can prove that (similar with Lemma~\ref{lem:R_s_po_est_err_bound}), with probability of at least $1-\delta$, for any $h\in\mathcal{H}$,  
\begin{align}
\label{eq:Gbound_2nd_risk_c1}
\tilde{R}^{\text{po}}_s(h,\tilde{f}_t,\bm{u}_s) 
\leq &~\mathcal{L}(\theta,h;\bm{u}_s,D^{xy}_{\tilde{s}}) + \frac{\sqrt{2}L_\ell\hat{\Re}_{D^x_s}(\mathcal{H})}{1-\tau_s} +\frac{3C_L}{1-\tau_s}\sqrt{\frac{\ln\frac{\delta}{2}}{2n_s}},
\end{align}
\begin{align}
\label{eq:Gbound_3rd_risk_c1}
\tilde{R}^{\text{po}}_t(h,\tilde{f}_t,\bm{u}_t) 
\leq &~\mathcal{L}(\theta,h;\bm{u}_t,D^{xy}_{\tilde{t}}) + \frac{\sqrt{2}L_\ell\hat{\Re}_{D^x_t}(\mathcal{H})}{1-\tau_s} +\frac{3C_L}{1-\tau_t}\sqrt{\frac{\ln\frac{\delta}{2}}{2n_t}}.
\end{align}
Combining \eqref{eq:Gbound_1st_risk}, \eqref{eq:Gbound_2nd_risk_c1}, \eqref{eq:Gbound_3rd_risk_c1} with $(*)$, we get the first $6$ terms in \eqref{eq:empicial_bound_NEW}. Hence we obtain all $6$ terms in \eqref{eq:empicial_bound_NEW} and prove this corollary.

\section{Additional Experimental Results}

In this section, we present the standard deviation of target-domain accuracy of all methods on WUDA tasks.

\begin{table*}[h]
  \centering
  \footnotesize
  \caption{The standard deviation of target-domain accuracy on $8$ digit WUDA tasks (\textit{SYND}$\leftrightarrow$\textit{MNIST}). Bold value represents the highest accuracy in each row.}
  \vspace{-1em}
    \begin{tabular}{ccllllllll}
    \toprule
    \multicolumn{1}{l}{Tasks} & \multicolumn{1}{l}{Type} & DAN & DANN & \multicolumn{1}{p{3.335em}}{ATDA} & \multicolumn{1}{p{3.28em}}{TCL} & \multicolumn{1}{p{3.92em}}{Co+TCL} & \multicolumn{1}{p{3.92em}}{Co+ATDA} & \multicolumn{1}{p{3.555em}}{B-Net} \\
    \midrule
    \multirow{4}[2]{*}{\textit{S}$\rightarrow$\textit{M}} & \multicolumn{1}{l}{P20} & 0.23\% & 1.12\% & 31.26\% & 3.88\% & 3.26\% & 0.66\% & 0.50\% \\
          & \multicolumn{1}{l}{P45} & 6.43\% & 6.88\% & 6.45\% & 7.08\% & 6.45\% & 4.02\% & 5.43\% \\
          & \multicolumn{1}{l}{S20} & 1.18\% & 1.29\% & 1.32\% & 1.17\% & 1.32\% & 0.38\% & 0.23\% \\
          & \multicolumn{1}{l}{S45} & 1.38\% & 1.59\% & 1.64\% & 1.62\% & 1.64\% & 1.29\% & 1.13\% \\
    \midrule
    \multirow{4}[2]{*}{\textit{M}$\rightarrow$\textit{S}} & \multicolumn{1}{l}{P20} & 4.30\% & 4.59\% & 4.62\% & 4.54\% & 4.62\% & 2.73\% & 4.31\% \\
          & \multicolumn{1}{l}{P45} & 2.01\% & 2.05\% & 2.06\% & 1.87\% & 2.06\% & 6.81\% & 4.06\% \\
          & \multicolumn{1}{l}{S20} & 4.82\% & 4.84\% & 4.88\% & 4.70\% & 4.88\% & 3.20\% & 2.66\% \\
          & \multicolumn{1}{l}{S45} & 2.02\% & 2.25\% & 2.25\% & 2.22\% & 2.25\% & 1.68\% & 2.79\% \\
    \midrule
    \multicolumn{2}{c}{Average} & 2.80\% & 3.08\% & 6.81\% & 3.39\% & 3.31\% & 2.60\% & 2.64\% \\
    \bottomrule
    \end{tabular}%
  \label{tab: digit_results_std}%
  \vspace{-1.2em}
\end{table*}%

\begin{table}[t]
  \centering
  \footnotesize
  \caption{The standard deviation of target-domain accuracy on $12$ {human-sentiment} WUDA tasks with the $20\%$ noise rate. Bold values mean the highest values in each row.}
  \vspace{-1em}
    \begin{tabular}{lllllllll}
    \toprule
    Tasks & \multicolumn{1}{p{3.5em}}{DAN} & \multicolumn{1}{p{3.5em}}{DANN} & \multicolumn{1}{p{3.22em}}{ATDA} & \multicolumn{1}{p{3.22em}}{TCL} & \multicolumn{1}{p{4.28em}}{MEDA} & \multicolumn{1}{p{4.28em}}{Co+TCL} & \multicolumn{1}{p{4.28em}}{Co+ATDA} & \multicolumn{1}{p{3.5em}}{B-Net} \\
    \midrule
    \emph{B}$\rightarrow$\emph{D}   & 1.48\% & 1.37\% & 1.45\% & 1.41\% & 1.40\% & 1.38\% & 1.47\% & 1.47\% \\
    \emph{B}$\rightarrow$\emph{E}   & 1.82\% & 1.67\% & 1.81\% & 1.77\% & 1.76\% & 1.77\% & 1.68\% & 1.81\% \\
    \emph{B}$\rightarrow$\emph{K}   & 1.34\% & 1.33\% & 1.33\% & 0.97\% & 1.31\% & 1.21\% & 1.24\% & 1.33\% \\
    \emph{D}$\rightarrow$\emph{B}   & 1.84\% & 1.50\% & 1.83\% & 1.78\% & 1.83\% & 1.68\% & 1.79\% & 1.63\% \\
    \emph{D}$\rightarrow$\emph{E}   & 1.78\% & 1.72\% & 1.75\% & 1.77\% & 1.74\% & 1.66\% & 1.72\% & 1.77\% \\
    \emph{D}$\rightarrow$\emph{K}   & 2.02\% & 1.98\% & 2.00\% & 1.81\% & 1.97\% & 1.96\% & 1.88\% & 1.90\% \\
    \emph{E}$\rightarrow$\emph{B}   & 1.54\% & 1.42\% & 1.52\% & 1.22\% & 1.53\% & 1.45\% & 1.51\% & 1.53\% \\
    \emph{E}$\rightarrow$\emph{D}   & 1.72\% & 1.65\% & 1.71\% & 1.48\% & 1.67\% & 1.79\% & 1.53\% & 1.70\% \\
    \emph{E}$\rightarrow$\emph{K}   & 1.29\% & 1.12\% & 1.27\% & 1.22\% & 1.27\% & 1.15\% & 1.14\% & 1.28\% \\
    \emph{K}$\rightarrow$\emph{B}   & 1.86\% & 1.74\% & 1.84\% & 1.72\% & 1.82\% & 1.72\% & 1.63\% & 1.85\% \\
    \emph{K}$\rightarrow$\emph{D}   & 0.44\% & 0.11\% & 0.42\% & 0.27\% & 0.39\% & 0.31\% & 0.21\% & 0.43\% \\
    \emph{K}$\rightarrow$\emph{E}   & 1.00\% & 0.68\% & 0.98\% & 0.64\% & 0.98\% & 0.96\% & 0.79\% & 0.99\% \\
    \midrule
    Average & 1.51\% & 1.36\% & 1.49\% & 1.34\% & 1.47\% & 1.42\% & 1.38\% & 1.48\% \\
    \bottomrule
    \end{tabular}%
  \label{tab: Amazon20_std}%
  \vspace{-1em}
\end{table}%

\begin{table}[h]
  \centering
  \footnotesize
  \caption{The standard deviation of target-domain accuracy on $12$ {human-sentiment} WUDA tasks with the $45\%$ noise rate. Bold values mean the highest values in each row.}
  \vspace{-1em}
    \begin{tabular}{lllllllll}
    \toprule
    Tasks & \multicolumn{1}{p{3.5em}}{DAN} & \multicolumn{1}{p{3.5em}}{DANN} & \multicolumn{1}{p{3.22em}}{ATDA} & \multicolumn{1}{p{3.22em}}{TCL}  & \multicolumn{1}{p{4.28em}}{MEDA} & \multicolumn{1}{p{4.28em}}{Co+TCL} & \multicolumn{1}{p{4.28em}}{Co+ATDA} & \multicolumn{1}{p{3.5em}}{B-Net} \\
    \midrule
    \emph{B}$\rightarrow$\emph{D}   & 1.11\% & 0.83\% & 0.92\% & 1.11\% & 1.11\% & 1.07\% & 0.97\% & 0.88\% \\
    \emph{B}$\rightarrow$\emph{E}   & 2.92\% & 2.86\% & 2.37\% & 2.57\% & 2.90\% & 2.90\% & 2.85\% & 2.69\% \\
    \emph{B}$\rightarrow$\emph{K}   & 2.12\% & 1.95\% & 1.89\% & 1.90\% & 2.11\% & 2.03\% & 1.76\% & 1.91\% \\
    \emph{D}$\rightarrow$\emph{B}   & 1.81\% & 1.71\% & 1.26\% & 1.28\% & 1.81\% & 1.70\% & 1.54\% & 1.52\% \\
    \emph{D}$\rightarrow$\emph{E}   & 1.71\% & 1.52\% & 1.14\% & 1.71\% & 1.70\% & 1.62\% & 1.43\% & 1.55\% \\
    \emph{D}$\rightarrow$\emph{K}   & 1.91\% & 1.62\% & 1.86\% & 1.65\% & 1.90\% & 1.85\% & 1.51\% & 1.74\% \\
    \emph{E}$\rightarrow$\emph{B}   & 1.37\% & 1.02\% & 1.16\% & 1.12\% & 1.36\% & 1.26\% & 0.90\% & 1.24\% \\
    \emph{E}$\rightarrow$\emph{D}   & 1.53\% & 1.23\% & 1.35\% & 1.27\% & 1.51\% & 1.43\% & 1.32\% & 1.23\% \\
    \emph{E}$\rightarrow$\emph{K}   & 1.29\% & 0.71\% & 0.75\% & 0.85\% & 1.28\% & 1.18\% & 0.89\% & 1.05\% \\
    \emph{K}$\rightarrow$\emph{B}   & 2.26\% & 1.92\% & 1.58\% & 2.08\% & 2.24\% & 2.16\% & 2.01\% & 2.06\% \\
    \emph{K}$\rightarrow$\emph{D}   & 2.86\% & 2.23\% & 2.58\% & 2.41\% & 2.85\% & 2.70\% & 2.35\% & 2.62\% \\
    \emph{K}$\rightarrow$\emph{E}   & 1.89\% & 1.46\% & 1.25\% & 1.62\% & 1.86\% & 1.85\% & 1.38\% & 1.67\% \\
    \midrule
    Average & 1.90\% & 1.59\% & 1.51\% & 1.63\% & 1.88\% & 1.81\% & 1.58\% & 1.68\% \\

    \bottomrule
    \end{tabular}%
  \label{tab: Amazon45_std}%
  \vspace{-1em}
\end{table}%

\begin{table}[h]
  \caption{The standard deviation of target-domain accuracy on $3$ real-world WUDA tasks. The source domain is the \emph{Bing} dataset that contains noisy information from the Internet. Bold value represents the highest accuracy in each row.}
  \vspace{-1em}
  \footnotesize
  \begin{center}
    \begin{tabular}{llllllll}
    \toprule
    Target & \multicolumn{1}{p{3.445em}}{DAN} & DANN & \multicolumn{1}{p{3.335em}}{ATDA} & \multicolumn{1}{p{3.28em}}{TCL} & \multicolumn{1}{p{4.22em}}{Co+TCL}& \multicolumn{1}{p{4.22em}}{Co+ATDA} & \multicolumn{1}{p{3.555em}}{B-Net} \\
    \midrule
    \textit{Caltech256} & 0.65\% & 0.52\% & 0.60\% & 0.48\% & 0.61\% & 0.34\% & 0.36\% \\

    \textit{Imagenet} & 0.32\% & 0.24\% & 0.29\% & 0.21\% & 0.26\% & 0.34\% & 0.51\% \\

    \textit{SUN} & 1.61\% & 1.51\% & 1.61\% & 1.55\% & 1.59\% & 1.87\% & 1.46\% \\

    \midrule
    Average & 0.86\% & 0.75\% & 0.83\% & 0.75\% & 0.82\% & 0.85\% & 0.78\% \\
    \bottomrule
    \end{tabular}%
    \vspace{-1em}
  \end{center}

\end{table}%

\end{document}